%% file: neurips_2026.tex
\documentclass{article}

\PassOptionsToPackage{numbers}{natbib}

\usepackage[main, final]{neurips_2026}

\usepackage[utf8]{inputenc} 
\usepackage[T1]{fontenc}    
\usepackage{textcomp}
\usepackage{hyperref}       
\usepackage{url}            
\usepackage{booktabs}       
\usepackage{amsfonts}       
\usepackage{nicefrac}       
\usepackage{xcolor}         
\usepackage{amsmath}
\usepackage{xcolor}
\usepackage{siunitx}
\usepackage{subcaption}
\usepackage{multirow}

\usepackage{graphicx}
\usepackage{wrapfig}
\usepackage{paralist}

\usepackage[table]{xcolor}
\usepackage{tikz}

\usepackage[acronyms, shortcuts]{glossaries}
\glsdisablehyper
\input{acronyms}

\usepackage[capitalise,noabbrev]{cleveref}

\let\originalleft\left
\let\originalright\right
\renewcommand{\left}{\mathopen{}\mathclose\bgroup\originalleft}
\renewcommand{\right}{\aftergroup\egroup\originalright}

\title{Njord: A Probabilistic Graph Neural Network for Ensemble Ocean Forecasting}

\author{%
  Daniel Holmberg\\
  University of Helsinki \\
  \texttt{daniel.holmberg@helsinki.fi} \\
  \And
  Joel Oskarsson \\
  ETH AI Center \\
  \texttt{joel.oskarsson@outlook.com} \\
  \AND
  Erik Wikingsson \\
  Linköping University \\
  \texttt{erik.wikingsson@gmail.com} \\
  \Ands
  Fredrik Lindsten \\
  Linköping University \\
  \texttt{fredrik.lindsten@liu.se} \\
  \Ands
  Teemu Roos \\
  University of Helsinki \\
  \texttt{teemu.roos@helsinki.fi} \\
}

\begin{document}

\maketitle

\begin{abstract}
Ocean dynamics are inherently chaotic, yet existing machine learning ocean models produce only deterministic forecasts. We introduce \emph{Njord}, a probabilistic data-driven model for ocean forecasting, applicable to both global and regional domains. Njord combines a deep latent variable framework with a graph neural network architecture, enabling sampling each forecast step in a single forward pass. We apply Njord globally at \SI{0.25}{\degree} resolution and regionally to the Baltic Sea at \SI{2}{\kilo\meter} resolution. To scale to these large ocean grids we introduce K-means cluster meshes that adapt to irregular sea surface geometry. Experiments demonstrate strong performance on both domains compared to deterministic machine learning baselines, while also providing uncertainty estimates from the sampled ensemble forecasts. On the global OceanBench benchmark, Njord achieves the lowest errors on average across upper-ocean variables when evaluated against real-world observations, with the largest improvements in surface temperature prediction.
\end{abstract}

\section{Introduction}
Accurate ocean forecasting is essential for a wide range of applications, from maritime navigation and fisheries management to coastal hazard mitigation and environmental monitoring~\cite{le2019observation}. While numerical ocean models have long served as the backbone of operational forecasting, they are computationally expensive and require substantial infrastructure to run at the resolutions needed in operational global~\cite{lellouche2023evolution} and regional applications~\cite{karna2021nemo}, taking on the order of hours on CPU clusters~\cite{aouni2025glonet, holmberg2025accurate}.

Recent advances have demonstrated that machine learning ocean models can match or even surpass the accuracy of physics-based systems at a fraction of the computational cost~\cite{aouni2025glonet,cui2025forecasting,wang2024xihe,huang2025fuxi} 
at global short-range (1--10 days) timescales. These models are however, deterministic: they produce a single trajectory and are typically trained with mean squared error, which encourages predictions toward the conditional mean of the future state rather than capturing the full predictive distribution. Consequently, they tend to smooth over fine-scale variance and offer limited insight into the probability of extreme events. For operational decision-making, where risk mitigation relies on understanding forecast confidence and the full spectrum of possible scenarios, a single trajectory is often insufficient. Probabilistic forecasting adds critical value here by modeling distributions over future states, allowing for the generation of ensembles that capture dynamic variability and quantify uncertainty. In the atmospheric domain, conditional generative models~\cite{oskarsson2024probabilistic, price2025probabilistic} have emerged as a promising framework for generating skillful and spatially coherent ensemble forecasts, but no comparable probabilistic data-driven model exists for the ocean.

In this work we propose, to our knowledge, the first probabilistic data-driven model for short-range high-resolution global and regional ocean forecasting. The model uses a latent variable framework built on hierarchical \glspl{GNN}~\citep{oskarsson2024probabilistic}, and produces calibrated ensemble forecasts of the depth-resolved ocean state. In a regional setting the model also conditions on boundary data from an independent global ocean model, where previous such emulators either lack boundary forcing~\cite{chattopadhyay2024oceannet}, or depend on data from the very system they aim to replace at the boundary~\cite{holmberg2025accurate}.

\paragraph{Our contributions are:}
\begin{enumerate}
    \item We introduce \emph{Njord}, the first generative ensemble forecasting model for global ocean physics, operating at \SI{0.25}{\degree} resolution.
    \item Njord employs a \gls{GNN} architecture using a new clustering-based graph layout, which better conforms to the irregular geometry of the ocean surface.
    \item In addition to existing variables included in high-resolution machine learning ocean models, we incorporate sea ice. 
    Sea ice is an integral component of ocean physics simulations, but requires additional constraints to ensure physically realistic fields.
    \item We follow the OceanBench \citep{el2025oceanbench} evaluation for global ocean emulators, and show that Njord achieves competitive errors compared with state of the art models, while adding information about uncertainties through the ensemble approach. For key surface variables Njord achieves the lowest error both compared to analysis data and direct observations.
    \item We further demonstrate that the same framework can be applied to regional ocean modeling by constructing Njord-Baltic for the Baltic Sea at \SI{2}{\kilo\meter} resolution. Njord-Baltic achieves errors comparable to a deterministic baseline while also providing probabilistic forecasts.
\end{enumerate}

\section{Related work}
\label{sec:related}

\paragraph{Ocean emulators.} At medium-range timescales, data-driven global ocean models~\cite{wang2024xihe,aouni2025glonet,cui2025forecasting,huang2025fuxi} have demonstrated good performance for deterministic forecasting. Regional approaches have also shown promising results: OceanNet~\cite{chattopadhyay2024oceannet} is able to outperform a dynamical ocean model in predicting \gls{SSH}, and SeaCast~\cite{holmberg2025accurate} is more skillful at forecasting the Mediterranean Sea at \SI{4}{\kilo\meter} resolution than the operational numerical model. Beyond medium-range forecasting, deep learning has been applied to seasonal ocean forecasting~\cite{andersson2021seasonal,wang2024coupled}, and climate prediction~\cite{dheeshjith2025samudra}. 
Despite these advancements, ensemble ocean forecasting with data-driven models is largely unexplored.
Recently, FuXi-ONS~\cite{huang2026data} extended global ocean emulation to the ensemble setting at a \SI{1}{\degree} grid spacing and 5 day intervals by adding learned perturbations on top of a deterministic core. 
This model is not included in our comparison as it focuses on different timescales and resolutions than our setting, and is also not trained to match the distribution of future states.

\paragraph{Probabilistic weather forecasting.} 
Our approach is heavily inspired by advancements in data-driven ensemble weather forecasting.
GenCast~\cite{price2025probabilistic} introduced diffusion-based ensemble forecasting for medium-range weather.
Diffusion models are effective at capturing high-resolution details~\cite{stormcast,larsson2025diffusionlam}, but are computationally demanding, as generating a single forecast requires many sequential forward passes. More recently, training generative forecasting models using the \gls{CRPS} as training objective has been a popular approach both at global~\cite{lang2026aifs,pacchiardi2024probabilistic,bonev2025fourcastnet,alet2025skillful} and regional scales~\cite{larsson2025crps,bris_crps}. 
These approaches differ primarily in how stochasticity is introduced and in how the \gls{CRPS} is estimated.
The \gls{CRPS}-based models are at least an order of magnitude more efficient during inference than diffusion-based approaches. 
Our work builds on the Graph-EFM~\cite{oskarsson2024probabilistic} latent variable model, which uses a combination of variational training and the \gls{CRPS}-based objective.
Graph-EFM is purely a weather model, and we extend the approach to the ocean domain through architectural modifications that enable efficient modeling on substantially larger and more irregular ocean grids.

\begin{figure}[ht]
  \centering
  \includegraphics[width=\textwidth]{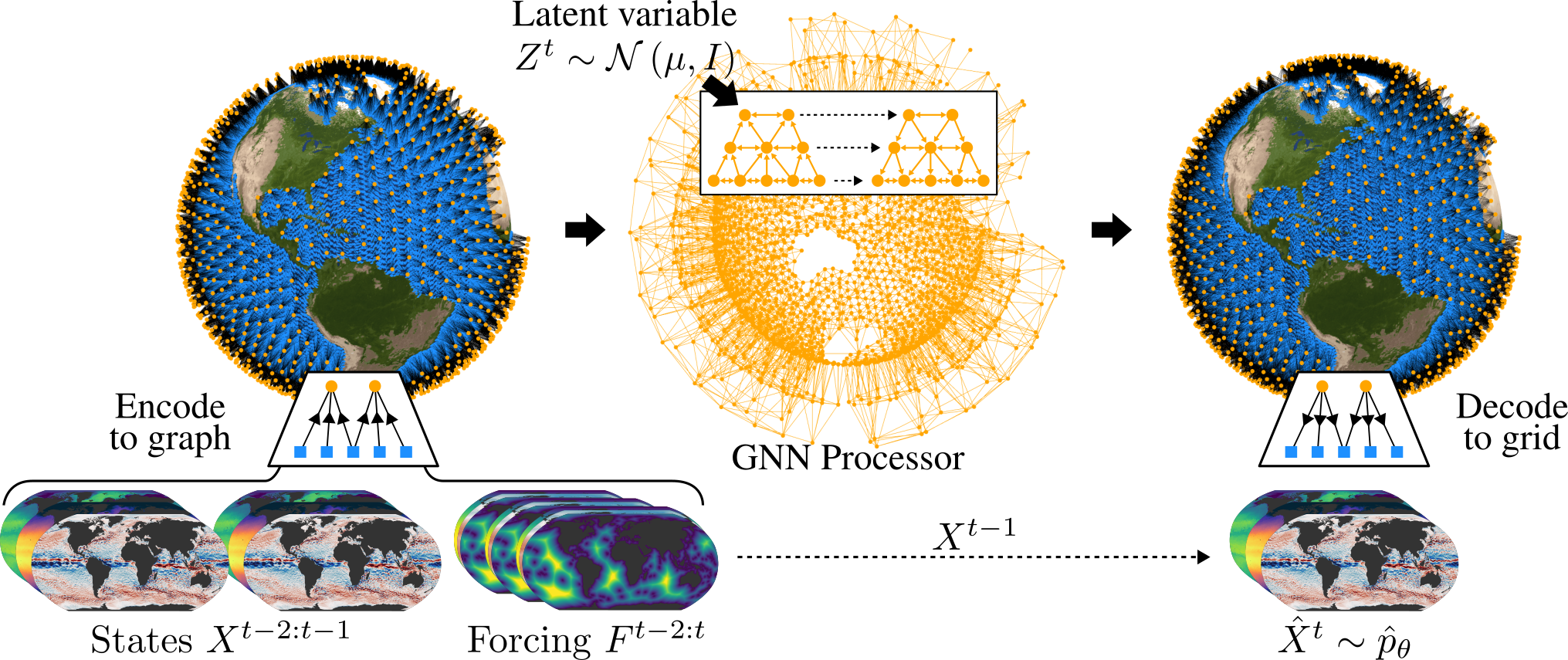}
  \caption{
  One-step prediction in the Njord model. 
  Residuals are predicted at time $t$, which are then added to the previous state $X^{t-1}$ in order to produce the sample $\hat{X}^{t}$.
  The corresponding overview of Njord-Baltic is shown in \cref{fig:baltic_model_diagram} in the appendix.
  }
  \label{fig:global_model_diagram}
\end{figure}

\section{Background}

\subsection{Problem formulation}
We are tackling the forecasting task of mapping a sequence of initial ocean states $X^{-p+1:0} = (X^{-p+1}, \ldots, X^{0})$, where $p$ is the number of past steps, to a sequence of future states $X^{1:T} = (X^{1}, \ldots, X^{T})$. 
Each state $X^t \in \mathbb{R}^{N \times d_x}$ contains $d_x$ ocean variables at $N$ different grid locations.
The state variables include quantities modeled at multiple vertical depth levels and surface-level variables. 
In addition to the initial states, the model receives additional forcing inputs on the same $N$ grid points.
This forcing $F^{-p+1:T}$ includes: 
\begin{inparaenum}[1)]
    \item known dynamic factors, such as the day of year,
    \item static features, e.g. depth, and
    \item atmospheric forcing at the surface, given by a weather model.
\end{inparaenum}
The atmospheric forcing includes relevant fields related to wind, pressure and temperature, that drive ocean surface dynamics.
In \cref{app:data-details} we provide complete lists of all state and forcing variables.

For ensemble forecasting we are specifically interested in a probabilistic mapping from initial states to forecasts.
We thus aim to model a distribution $p(X^{1:T} \mid X^{-p+1:0}, F^{-p+1:T})$ describing the possible future ocean states.
Under the Markov assumption, the ocean state at time $t$ depends only on the previous $p$ states. We therefore model $p(X^{t} \mid X^{t-p:t-1}, F^{-p+1:T})$ where the forcing sequence $F^{-p+1:T}$ is assumed to be available throughout the forecast horizon. This is reasonable in our setting because the forcing consists of pre-computed static and dynamic features, as well as weather forecasts.
Ensemble members, samples of $X^{1:T}$, can then be produced by sequentially sampling $X^t$ from this distribution, and conditioning on the sampled value for the next time step.

\subsection{Ensemble forecasting with latent variable models}
\label{sec:latent-variable}
The ensemble forecasting problem outlined above can be approached by training a deep generative model to approximately produce samples from $p\left(X^{t} \mid X^{t-p:t-1}, F^{-p+1:T}\right)$.
One family of such models are deep latent variable models, similar to conditional variational auto-encoders \citep{cvae}, which learn to map from conditioning inputs and a latent random variable $Z^t$ to samples from a distribution.
Typically the set of conditioning inputs is restricted to only a subset of forcing and previous states.
In our setting such a model is comprised of a neural network $f_\theta$ realizing the mapping
\begin{equation}
   \hat{X}^{t} = f_\theta\left(X^{t-2:t-1}, F^{t-2:t}, Z^t\right),
   \qquad
    Z^t \sim \mathcal{N}\left(\mu^t, I\right)
   \label{eq:dlvm}
\end{equation}
where $Z^t$ is chosen to be an isotropic Gaussian.
This sampling implicitly specifies the model distribution $\hat{p}_\theta\left(X^{t} \mid X^{t-2:t-1}, F^{t-2:t}\right)$.
As sampling in these models only requires a single forward-pass with $f_\theta$, it is far more computationally efficient than the iterative sampling of diffusion models~\citep{price2025probabilistic,hatanpaa2025aeris,finn2024diffusion,larsson2025diffusionlam}.
Recently there has been an increasing interest for using deep latent-variable models in earth system modeling \citep{swin_vrnn,oskarsson2024probabilistic,lang2026aifs,alet2025skillful,bonev2025fourcastnet}.

In this work we build on the Graph-EFM~\cite{oskarsson2024probabilistic} latent variable model, previously used for weather forecasting.
Graph-EFM uses a constructed mesh graph and \gls{GNN} layers \citep{battaglia2016interaction} in order to capture spatial relationships over the forecasted area.
The neural network $f_\theta$ is implemented as a \gls{GNN} following the encode-process-decode architecture.
Gridded inputs are first mapped to latent representations on the mesh graph.
The processor part of the model then consists of a hierarchy of \gls{GNN} layers operating over different spatial resolutions, with $Z^t$ being integrated at the coarsest level by adding it to the latent representation.
As \gls{GNN}s map from this coarse representation down through the hierarchy, and finally decodes back to the original grid points, the stochasticity from $Z^t$ can affect all outputs of the model.
These outputs are then added to the previous state $X^{t-1}$ through a residual connection, finally producing the random sample $\hat{X}^t$.
In Graph-EFM $Z^t$ is additionally sampled from a prior distribution with a learned mean, as $\mu^t = g_\theta\left(X^{t-2:t-1}, F^{t-2:t}\right)$.
This prior mapping is realized as another hierarchical \gls{GNN}.
The model is trained by optimizing the \gls{ELBO}, which also requires training a separate encoder network for $Z^t$. This encoder has a similar structure to the prior, but is only required as an auxiliary module for training, and not for forecasting.
For further details about Graph-EFM see \cref{app:model-details} and \citet{oskarsson2024probabilistic}.

At inference time, each ensemble member is sampled at time step $t$ by 
\begin{inparaenum}[1)]
    \item drawing a sample $Z^t \sim \mathcal{N}\left(\mu^t, I\right)$, and
    \item propagating this sample through $f_\theta$ together with conditioning variables,
\end{inparaenum}
in order to produce $\hat{X}^t$.
This is repeated auto-regressively across $T$ time steps, conditioning each step on previously sampled states, to obtain full state trajectories.
A complete ensemble is generated by repeating this with independent samples, which is fully parallelizable across members.

\section{Njord: A graph-based probabilistic ocean model}
To tackle the problem of probabilistic ocean forecasting, we follow the latent variable approach, providing efficient ensemble sampling and model training that directly targets the distribution of future ocean states.
We introduce Njord, a graph-based forecasting model for the global Ocean, and Njord-Baltic, a regional adaptation for the Baltic Sea (\cref{sec:regional_adapt}).
Njord extends the hierarchical GNN architecture of Graph-EFM \citep{oskarsson2024probabilistic} with several key modifications tailored to high-resolution ocean modeling.
Graph-based modeling is appealing for ocean forecasting due to the inherent ability to work with irregular grids.
As opposed to methods that work with regular latitude-longitude grids, Njord only operates over the actual ocean grid points.
This avoids wasting memory and compute on updating latent representations located over land, where there are no ocean dynamics to model.

Njord instantiates the latent variable framework from \cref{sec:latent-variable}, and we make the specific choice of including two previous states $X^{t-2:t-1}$ and both past and future forcing $F^{t-2:t}$ as gridded inputs to the networks $f_\theta$ and $g_\theta$.
Including multiple past states helps the model capture higher-order dynamics, and forcing information across time is generally useful for fields with clear relationships to the atmosphere (see \cref{sec:inputs_ablation} for an ablation). 
These gridded outputs are mapped to our hierarchical ocean graph, where $Z^t$ is integrated in the latent state, providing the stochasticity for our probabilistic ocean forecast.
An overview of the method is shown in \cref{fig:global_model_diagram}.
\begin{wrapfigure}{r}{0.4\textwidth}
    \vspace{-.7\intextsep}
    \begin{subfigure}{.2\textwidth}
      \includegraphics[width=\textwidth]{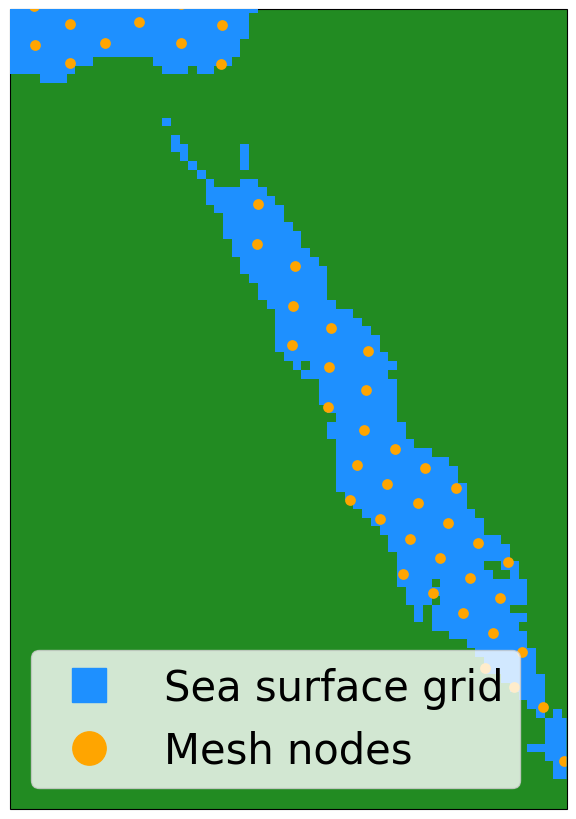}
      \caption{Icosahedral}
    \end{subfigure}%
    \begin{subfigure}{.2\textwidth}
      \includegraphics[width=\textwidth]{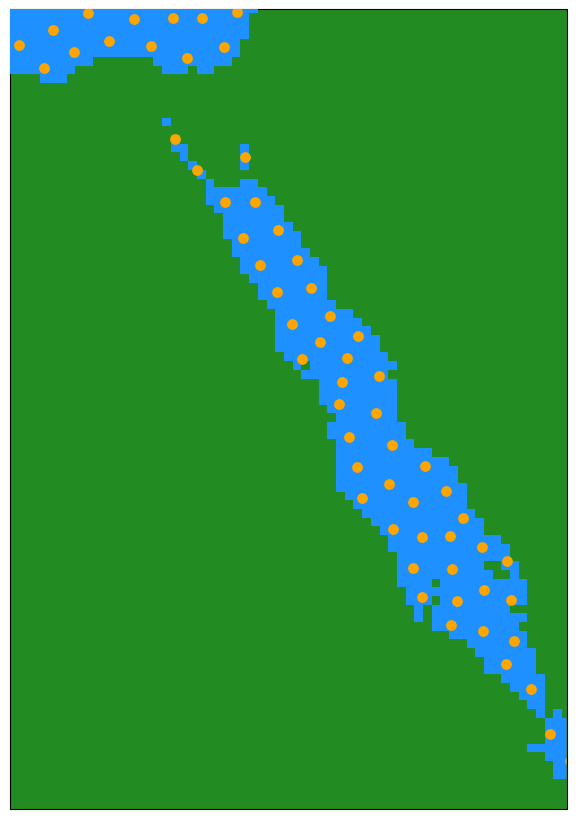}
      \caption{Cluster}
    \end{subfigure}
     \caption{
        Example of graph node placement in the Red Sea.
    }
    \label{fig:main_graph_example}
    \vspace{-3\intextsep}
\end{wrapfigure}
\subsection{A graph adapted to ocean geometry}
\label{sec:adapting_graph}
Graph-based global weather forecasting models use icosahedral meshes~\cite{lam2023learning,oskarsson2024probabilistic,lang2024aifs} for constructing the spatial graph that the model operates over. 
These meshes are constructed by iteratively subdividing an icosahedron, with each subdivision quadrupling the number of nodes and edges \citep{lam2023learning}.
As the size of the graph heavily impacts memory requirements, choosing the number of nodes and edges is a crucial choice in practice.
However, icosahedral refinement only allows discrete subdivision levels, limiting the available graph sizes.
Additionally, while the icosahedral shape is well-motivated for the atmosphere surrounding Earth, it is less suitable for modeling the ocean surface. 
Masking can be applied to only include nodes over the ocean, but without any adaptation to node placement this risks breaking apart large and important parts of the icosahedral graph connectivity.

To construct a graph better adapted to the geometry of the global ocean we instead place the graph nodes based on the density of ocean grid points.
We apply spherical K-means clustering of the ocean grid point 3D Cartesian coordinates, with latitude-based area weights to ensure equitable spatial coverage. 
As shown in \cref{fig:main_graph_example}, this leads to more evenly spaced nodes and avoids missing important bays, straits and canals (see \cref{sec:app_graph_construction} for more examples and \cref{sec:graph_type_ablation} for an empirical model comparison).
The clustering based approach can be used to place an arbitrary number of nodes, equal to the number of clusters.
Edges are then constructed via spherical Delaunay triangulation, followed by filtering of edges crossing land masses.
The procedure is repeated multiple times with a decreasing number of clusters, resulting in ocean graphs of increasing coarseness.
These then make up the hierarchy used by the \gls{GNN} layers in the model.
As the graph and grid points in Njord do not cover land, we instead add the minimum distance to the coastline as a static input feature in each grid point, informing the model where the ocean surface ends.
Further details on how the grid points are connected to the graph, land-crossing edge filtering, and static graph features are given in \cref{app:graph-details}.

\subsection{Training objective}
\label{sec:training}
Similar to Graph-EFM \citep{oskarsson2024probabilistic}, Njord is trained using a combination of the \gls{ELBO} and \gls{CRPS}.
We apply masking to adapt to the ocean's bathymetric structure and weight the loss for each prediction based on the area of the grid cell, the depth-level and the inverse variance of daily change in each variable~\citep{holmberg2025accurate}.
To use the \gls{ELBO}, we train also an additional encoder network as a variational approximation $q_\phi$. This is parametrized as another Gaussian distribution, similar to the prior.
The full loss function for predictions at time $t$ is
\begin{equation}
\begin{split}
    \mathcal{L} = &-\mathbb{E}_{q_\phi}\!\left[\log p_\theta(X^t \mid X^{t-2:t-1}, F^{t-2:t}, Z^t)\right]
    \\
    &+ \lambda_{\text{KL}}\, D_{\mathrm{KL}}\!\left(q_\phi(Z^t \mid X^{t-2:t}, F^{t-2:t}) \,\|\, p_\theta(Z^t \mid X^{t-2:t-1}, F^{t-2:t})\right) 
    + \lambda_{\mathrm{\gls{CRPS}}} \mathcal{L}_{\mathrm{\gls{CRPS}}}.
    \label{eq:total-loss}
\end{split}
\end{equation}
The likelihood $p_\theta$ is Gaussian with mean $\hat{X}^t$ and variance based on the earlier mentioned weightings.
We compute the \gls{CRPS} term $\mathcal{L}_{\mathrm{\gls{CRPS}}}$ using the almost-fair \gls{CRPS} estimator \citep{lang2026aifs} from two independently sampled forecasts.
Training follows a curriculum in multiple phases, gradually activating the KL term ($\lambda_{\text{KL}}$), introducing multi-step rollout training (summing \cref{eq:total-loss} for $t=1,2,\dots$), and finally enabling the \gls{CRPS} loss ($\lambda_{\mathrm{\gls{CRPS}}}$).
For more details about the training objective, including the exact training curriculum, see \cref{sec:training_details}.

\subsection{Sea ice modeling with physical constraints}
\label{sec:physical-constraints}

Unlike other machine learning models for high-resolution short term ocean forecasting~\citep{aouni2025glonet, cui2025forecasting, wang2024xihe}, Njord also models sea ice. These ice-related variables have specific physical constraints: \gls{SIC} is bounded to $[0, 1]$ and \gls{SIT} must be positive.
A large fraction of values also sit exactly at these limits.
Without explicitly enforcing these constraints, autoregressive forecast rollout can produce unphysical predictions outside of these bounds. As a probabilistic model, this is especially a concern for Njord, as we want the full probability distribution to conform to the constraints, not just the conditional mean.  

Instead of enforcing constraints only through post-processing, we aim to also prevent the model from encountering negative sea ice inputs during training. In \cref{sec:sea_ice_comparison}, we compare different strategies at global \SI{1}{\degree} resolution and find that a combination of soft clamping and a \emph{density channel}~\cite{gordon2019convolutional,bodnar2025foundation} performs best. A binary density channel $d \in \{0,1\}$ is constructed from \gls{SIC} as $d=\mathbb{I}[\mathrm{SIC}>0]$ and appended to the model state. The model predicts this channel jointly with all other variables. The predicted density logit is passed through a sigmoid and thresholded at $0.5$. Where the predicted density falls below the threshold, the density channel and all ice variables are set to zero in the next state; otherwise, the predicted ice values are retained. This ensures that locations predicted to be ice-free receive clean zero-ice inputs, rather than small residual values that may accumulate over rollouts. The raw, pre-threshold predictions are still used for the loss computation.

\subsection{Scaling to high-resolution global grids}
\label{sec:scaling}
Scaling Njord to global ocean grids at \SI{0.25}{\degree} presents several modeling and training challenges.
Similar to \citet{alet2025skillful}, we note that much of the memory footprint in graph-based models can be attributed to the many edges used for encoding the gridded data to the graph.
We modify the Interaction \cite{battaglia2016interaction} and Propagation Network \glspl{GNN} \cite{oskarsson2024probabilistic} in the model to use separate dimensionalities for the original grid-level embeddings, edge representations and graph node representations.
This allows us to substantially reduce the memory used for edge representations, and instead scale up the hidden dimensionality in the core \gls{GNN} processor of the model.
Exact dimensionalities are given in \cref{sec:app_scaling}.
We additionally apply gradient checkpointing \citep{optimal_grad_checkpointing, chen2016training} between time steps to allow for training on auto-regressive rollouts.


We further reduce compute and training time by following a two-stage training schedule, where we pretrain on \SI{1}{\degree} resolution data before finetuning at \SI{0.25}{\degree}.
The same graph is used for both resolutions, only swapping the encoding and decoding edges that connect the mesh to the high-resolution grid.
The full training curriculum is outlined in \cref{app:training-details}.

\subsection{Extension to regional modeling: Njord-Baltic}
\label{sec:regional_adapt}
The framework underpinning the global Njord model is general, and can also be applied to train regional ocean models.
We exemplify this by building Njord-Baltic, a probabilistic ocean forecasting model for the Baltic Sea.
Quadrilateral graphs have previously been used for regional ocean models \citep{holmberg2024regional}, but these suffer from some of the same limitations as icosahedral graphs in the global setting.
We instead use our clustering-based graph construction also for Njord-Baltic.

For regional models, boundary conditions can become an additional consideration.
We handle this by introducing an additional boundary forcing input $B^{t}$, and include this from times $t-2,t-1$ and $t$ for predicting $\hat{X}^t$.
This boundary forcing contains information about the surrounding ocean state, and can come from a global forecasting model or reanalysis data.
We use a separate encoder for the boundary inputs \citep{adamov2025building},
allowing these to stay at the original coarse resolution of the global model.
An overview of the regional setup for Njord-Baltic is shown in \cref{fig:baltic_model_diagram} in the appendix.

\section{Experiments}

Both global and regional models are trained on ocean reanalysis and atmospheric reanalysis data spanning 1993--2021, after which the models are finetuned on analysis data from 2023. The finetuning is helpful to achieve a better calibrated ensemble, as the operational initial conditions are produced by the same model. In the global data, there is also a slight mismatch in the bathymetry used by the reanalysis and analysis, and because we connect the GNN mesh to the surface grid, we prefer this to be the exact same for input and output. During evaluation, 52 forecasts are initialized weekly (every Tuesday) throughout 2024, following the OceanBench~\cite{el2025oceanbench} benchmark, and evaluated over a 10-day forecast horizon. The models are implemented~\footnote{Our code is available at \url{https://github.com/deinal/njord}.} in PyTorch.

\subsection{Global ocean forecasting}

\paragraph{Global ocean data.} The global ocean state for training is taken from the GLORYS12 global ocean reanalysis~\cite{jean2021copernicus}, a $1/12^\circ$ resolution NEMO~\cite{madec2015nemo} based reanalysis assimilating satellite altimetry, \acrfull{SST}, and in-situ profiles. GLORYS12 provides daily-mean fields of \acrfull{SSH}, \acrfull{SIC}, \acrfull{SIT}, \acrfull{T}, \acrfull{S}, \acrfull{U}, and \acrfull{V} at multiple depth levels. For the global Njord configuration, we use these variables at six representative depths. We additionally finetune on operational GLO12~\cite{lellouche2023evolution} analysis data. Surface atmospheric forcing is obtained from the ERA5 global reanalysis~\citep{hersbach2020era5}, produced by ECMWF at \SI{0.25}{\degree} resolution. We use eight surface atmospheric variables, bilinearly interpolated to the ocean model grid. During evaluation we switch to operational 10-day IFS \citep{ifs} atmospheric forecasts for the forcing. Further details are listed in~\cref{tab:variables_global}.
\glsunset{SST}\glsunset{SSH}\glsunset{SIC}\glsunset{SIT}\glsunset{T}\glsunset{S}\glsunset{U}\glsunset{V}

\paragraph{Global baselines.} We use the machine learning models from OceanBench~\cite{el2025oceanbench} as global baselines, namely GLONET~\cite{aouni2025glonet}, WenHai~\cite{cui2025forecasting}, and XiHe~\cite{wang2024xihe}. To also provide a deterministic counterpart with a comparable training strategy and data splits to Njord, we extend SeaCast~\cite{holmberg2025accurate} to the global setting by adopting a global icosahedral mesh instead of the regional quadrilateral mesh it normally uses.
As a physics-based operational baseline, we use GLO12~\cite{lellouche2023evolution}. 
We also include a persistence baseline, which repeats the last initial state over the whole forecast horizon. 
Njord, SeaCast, and GLONET operate at \SI{0.25}{\degree}, whereas WenHai, XiHe, and GLO12 use the native \SI{0.083}{\degree} simulation grid.

\paragraph{Global results.}

\begin{figure}[t]
    \centering
    \includegraphics[width=1.0\textwidth]{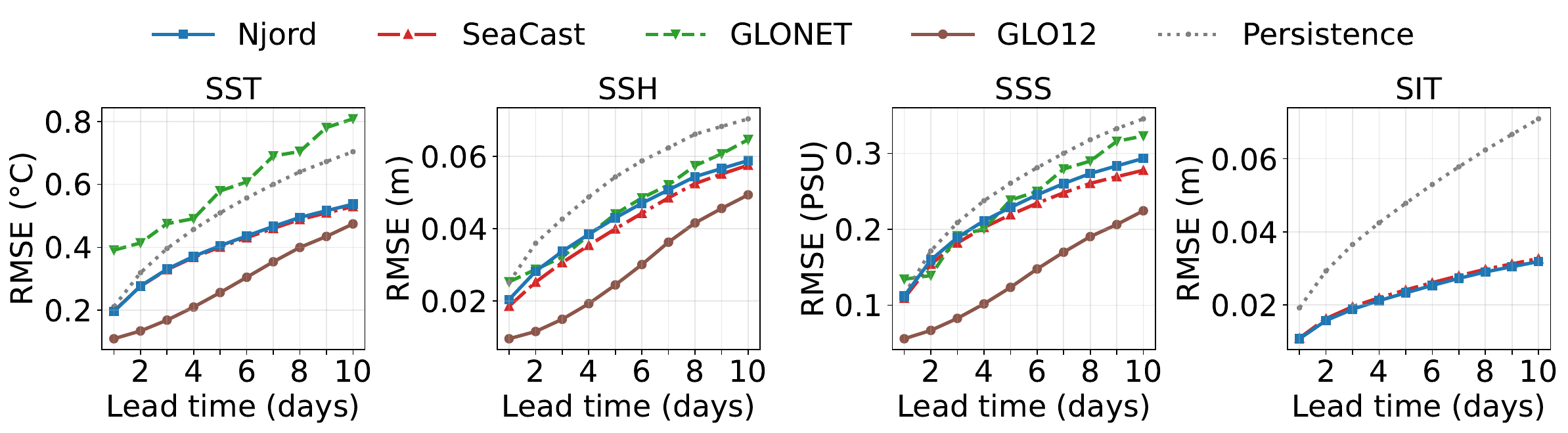}
    \caption{\gls{RMSE} for \acrfull{SST}, \acrfull{SSH}, \acrfull{SSS}, and \acrfull{SIT} at each lead time relative to GLO12 analysis.}
    \label{fig:global_metrics}
\end{figure}

Njord samples a single next-step ensemble member in \SI{3}{\second} on one AMD MI250X GPU. In these experiments, we generate a 20-member global ensemble. Lead-time performance is evaluated in \cref{fig:global_metrics} for selected variables. \gls{RMSE} is computed for the ensemble mean and compared against the GLO12 analysis at \SI{0.25}{\degree}. Full results from the global experiments are presented in \cref{sec:additional_results}, including scores for all models on OceanBench. Across all variables, Njord demonstrates competitive performance relative to machine learning baselines, with stable error growth over the 10-day forecast horizon. SeaCast exhibits similar behavior, as it is pretrained on reanalysis and fine-tuned on analysis data, consistent with Njord, whereas GLONET represents a state-of-the-art model trained solely on reanalysis. Results for the two other reanalysis-based models XiHe and WenHai are shown in \crefrange{tab:glorys12_track}{tab:observation_track}. The GLO12 forecast constitutes a particularly strong baseline, as it is generated by the same system used to produce the analysis targets. Due to the sparsity of ocean observations, outperforming the physical simulator on the analysis benchmark is inherently challenging. On the other hand, machine learning models trained exclusively on reanalysis data may be biased toward that dataset. They indeed outperform GLO12 on a number of reanalysis fields as seen in \cref{tab:glorys12_track}. A more independent assessment of performance is therefore comparing against observations. Nonetheless, analysis evaluation remains important, as initial conditions are derived from the same system, enabling comparison of consistent trajectories.

\begin{figure}[t]
    \centering
    \includegraphics[width=\textwidth]{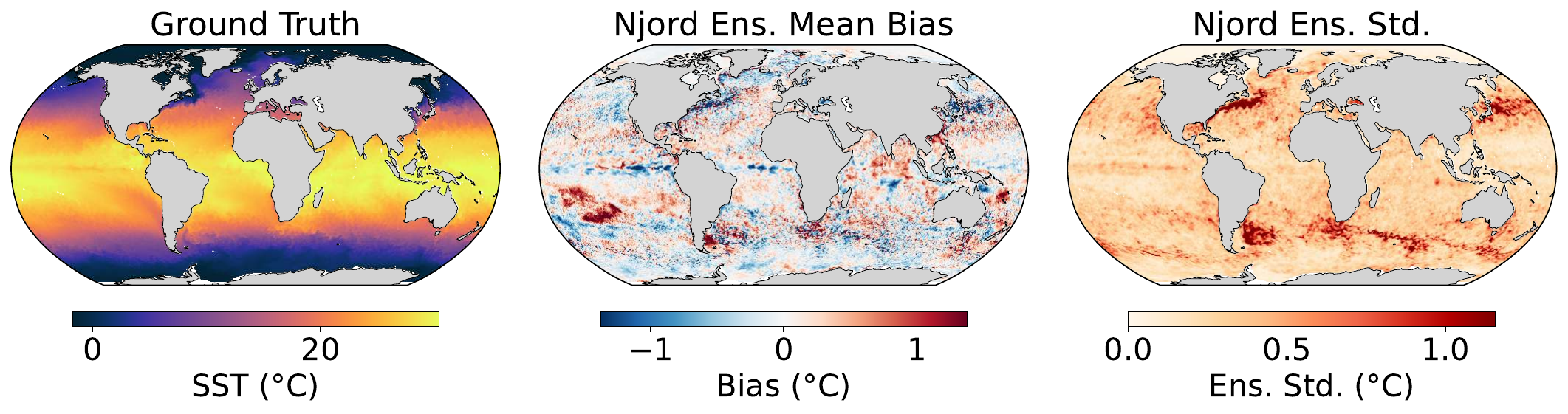}
    \caption{
    Global \gls{SST} at a \SI{10}{d} lead, initialized on 2024-01-30. 
    Ground truth is GLO12 analysis.
    }
    \label{fig:global_sst}
\end{figure}

\begin{wrapfigure}{r}{0.3\textwidth}
    \vspace{-1.05\intextsep}
    \centering
    \includegraphics[width=\linewidth]{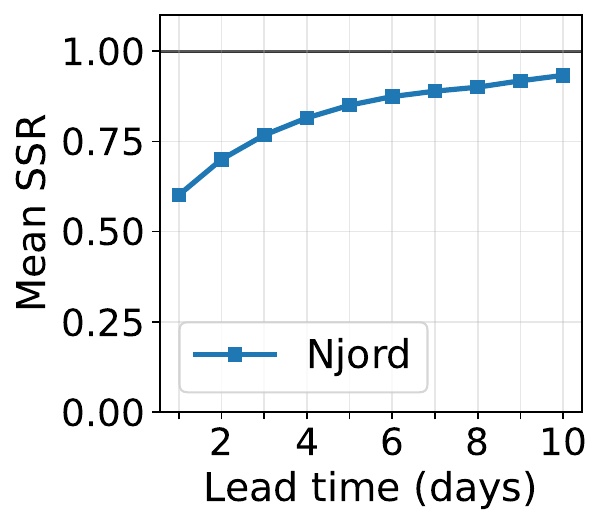}
    \caption{\acrshort{SSR} averaged over all global ocean variables.}
    \label{fig:global_mean_ssr}
    \vspace{-2\intextsep}
\end{wrapfigure}

The \gls{SSR} in \cref{fig:global_mean_ssr} measures the ratio between ensemble spread and forecast error, providing an assessment of probabilistic calibration. The 20-member Njord ensemble exhibits a slight underdispersion at short lead times, but the \gls{SSR} increases steadily toward 1, indicating good calibration and a reliable representation of forecast uncertainty. This behavior is consistent across the ocean variables. Additional probabilistic evaluation, including CRPS and variable-specific SSR across all lead times, is provided in~\crefrange{fig:global_metrics_surface}{fig:global_metrics_uo} in the appendix.

\begin{figure}[t]
    \centering
    \includegraphics[width=\textwidth]{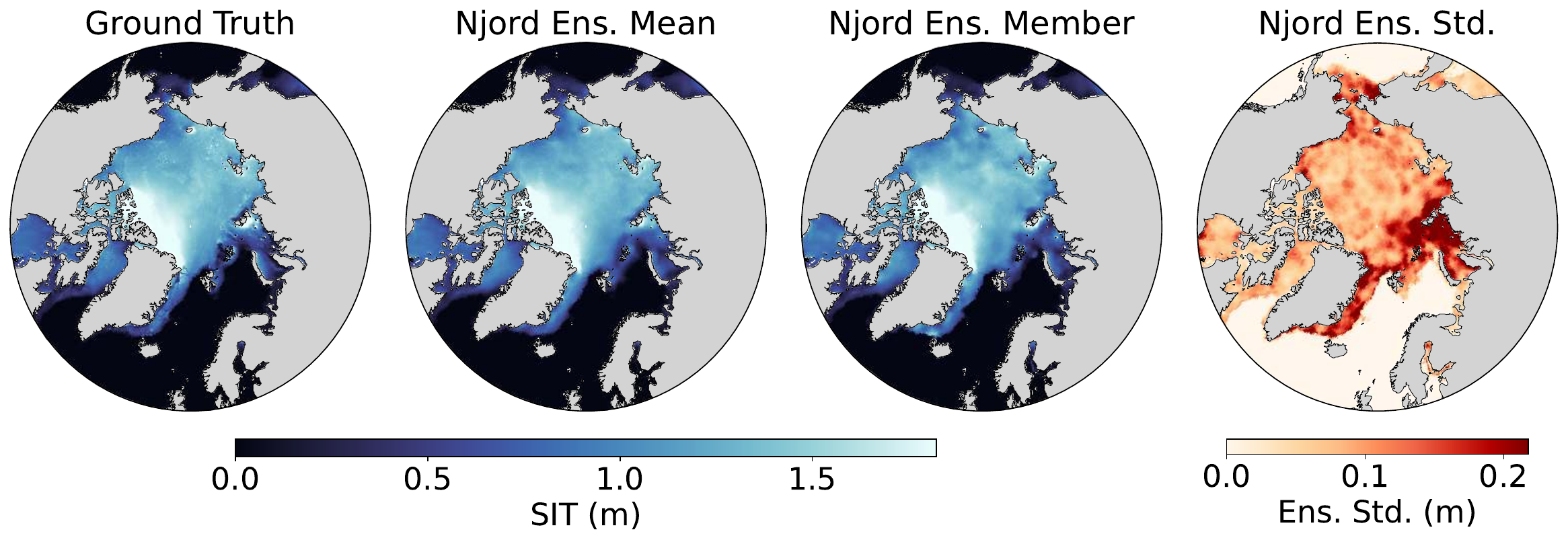}
    \caption{Arctic \acrshort{SIT} at \SI{10}{d} lead time, initialized 2024-01-30. 
    Ground truth is GLO12 analysis.
    }
    \label{fig:global_sit}
\end{figure}

Spatial performance for \gls{SST} at a 10-day lead time is shown in Figure~\ref{fig:global_sst}. The Njord ensemble mean accurately captures the global thermal structure of the GLO12 analysis, including sharp gradients associated with major current systems. The ensemble spread highlights regions of increased uncertainty, aligning with dynamically active areas such as the Gulf Stream, the Brazil--Malvinas Confluence, the Agulhas Retroflection, and the Kuroshio Extension. 
Forecasts of Arctic sea ice distribution is shown in Figure~\ref{fig:global_sit}. Njord reproduces large-scale ice thickness gradients and the marginal ice zone. Ensemble spread peaks in the Kara Sea and other highly dynamic marginal zones, reflecting uncertainty in sea ice transport over the 10-day forecast horizon.

\paragraph{Observation results.}

The OceanBench~\cite{el2025oceanbench} observation track enables independent verification using the IV-TT CLASS-4 dataset, which comprises temperature and salinity measured by a global array of autonomous floats, \gls{SST} from surface drifting buoys, \gls{SLA} from along-track satellite measurements, and surface currents at 15~m depth from drifter buoys. As shown in \cref{tab:observation_track}, Njord achieves the lowest \gls{RMSE} for 0--5~m temperature across all lead times, outperforming both the operational GLO12 baseline and other machine learning emulators. Njord tends to perform on par with the best model for each observable in the upper ocean, less than \SI{100}{m} depth, on the observation track.

\begin{wrapfigure}{r}{0.43\textwidth}
    \vspace{-1\intextsep}
    \centering
    \includegraphics[width=\linewidth]{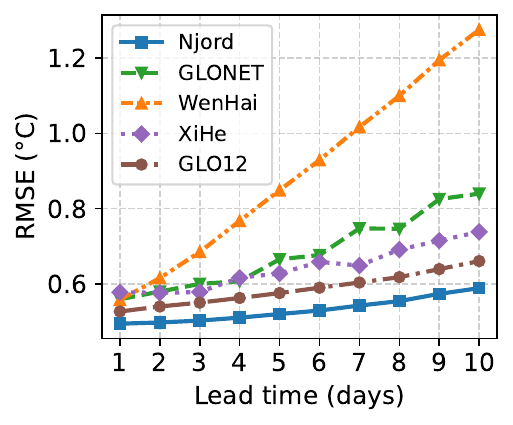}
    \caption{Global \gls{SST} predictions evaluated on satellite measurements.}
    \label{fig:sst_rmse_lineplot}
    \vspace{-1.5\intextsep}
\end{wrapfigure}
To further evaluate \gls{SST} forecasts outside of OceanBench, we compare the predicted potential temperature of the uppermost ocean layer against a global ocean bias-adjusted \gls{SST} product~\cite{cmems2026sst}, based on multi-sensor satellite observations. \Cref{fig:sst_rmse_lineplot} shows globally averaged \gls{SST} \gls{RMSE} over a 10-day forecast horizon, where all models are interpolated to the \SI{0.1}{\degree} \gls{SST} grid. Njord maintains the lowest error across all lead times compared to all other models.

Although Njord is not trained directly on observations, both the reanalysis and operational analysis targets assimilate observations, meaning that observational information is implicitly present in the training targets. We hypothesize that the combination of a comprehensive atmospheric forcing, the coastline-adapted hierarchical graph architecture, and ensemble averaging all contribute to the strong performance observed on independent measurements.

\subsection{Regional ocean forecasting}

\paragraph{Regional ocean data.}
The Baltic Sea state is obtained from the Baltic Sea Physics Reanalysis, produced with the NEMO ocean model~\cite{madec2015nemo} at \SI{2}{\kilo\meter} horizontal resolution. We use daily-mean fields of \gls{SLA}, \gls{SIC}, \gls{SIT}, and the three-dimensional \gls{T}, \gls{S}, \gls{U}, and \gls{V}, subsampled to five representative depth levels listed in~\cref{tab:variables_baltic}. For the regional configuration, GLORYS12 provides lateral boundary forcing during training, supplying the three-dimensional ocean state along the open boundaries of the Baltic domain. During evaluation, GLO12 forecasts are used as boundary forcing. Atmospheric forcing follows the global setup: ERA5 is used during training and operational IFS forecasts during evaluation, both interpolated to the Baltic Sea grid. 

\paragraph{Regional baselines.}
We compare Njord-Baltic to SeaCast~\cite{holmberg2024regional}, which serves as the primary baseline. We also include GLO12 forecasts from OceanBench for reference, although they are not expected to be competitive at kilometer-scale resolution due to their coarser \SI{0.083}{\degree} grid. Also here we include a persistence baseline.

\begin{wrapfigure}{r}{0.3\textwidth}
    \vspace{-1\intextsep}
    \centering
    \includegraphics[width=\linewidth]{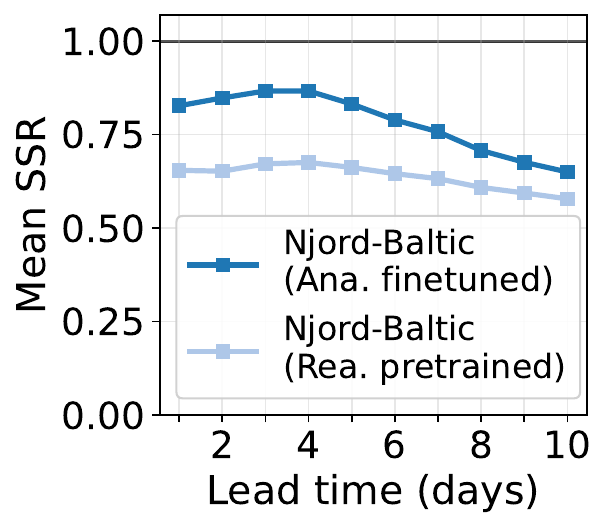}
    \caption{\gls{SSR} averaged over Baltic Sea variables.}
    \label{fig:baltic_mean_ssr}
    \vspace{-1\intextsep}
\end{wrapfigure}

\begin{figure}[tbp]
    \centering
    \includegraphics[width=\textwidth]{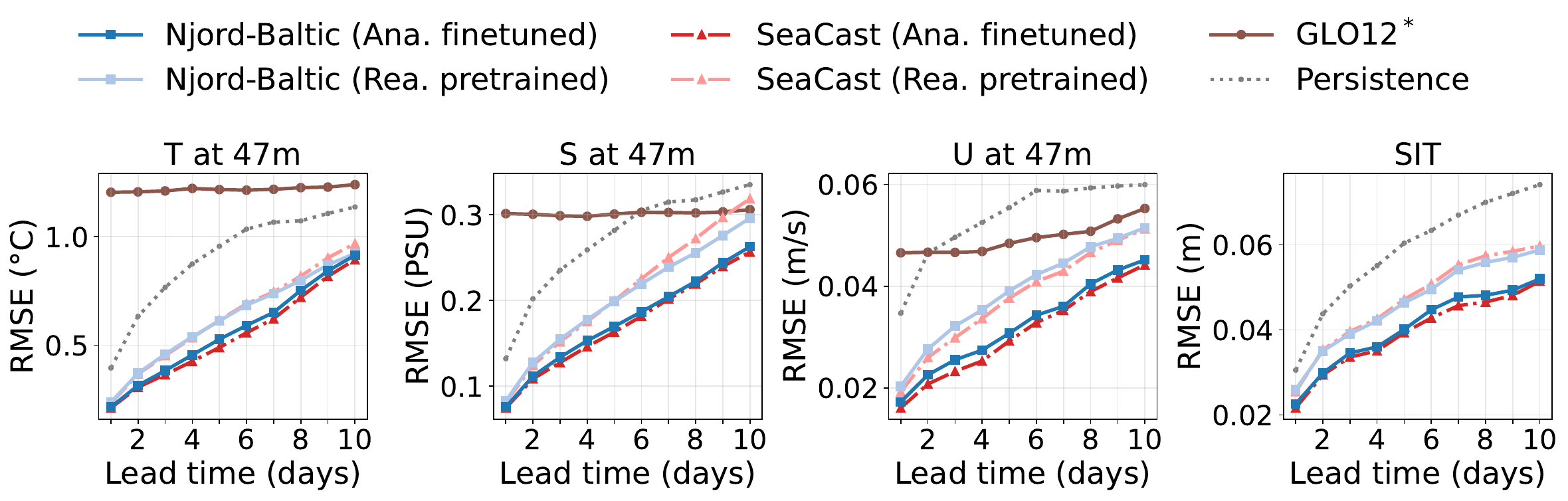}
    \caption{RMSE for \acrfull{T}, \acrfull{S}, \acrfull{U} at \SI{47}{m} depth, as well as \acrfull{SIT}, all relative to NEMO analysis. 
    }
    \label{fig:baltic_metrics}
\end{figure}

\paragraph{Regional results.}
Njord-Baltic samples a single next-step ensemble member in \SI{1}{\second} on one AMD MI250X GPU. In these experiments, we generate a 5-member ensemble. Both Njord-Baltic and SeaCast are evaluated under two training regimes: pretraining on reanalysis data and subsequent fine-tuning on operational analysis data. Fine-tuning leads to a consistent improvement in probabilistic calibration, as reflected by higher \gls{SSR} values in \cref{fig:baltic_mean_ssr}. It also reduces the error as seen in \cref{fig:baltic_metrics} containing model performance for \gls{T}, \gls{S}, \gls{U} at \SI{47}{m} depth, as well as \gls{SIT}.

Across variables, Njord-Baltic achieves RMSE values comparable to SeaCast while providing probabilistic forecasts. In this regional setting, GLO12 exhibits a relatively flat error curve, similar to a climatological baseline. Both Njord-Baltic and SeaCast clearly outperform persistence. 
Njord-Baltic matches SeaCast in deterministic accuracy while additionally providing ensemble forecasts, though it exhibits underdispersion at longer lead times. Per-variable CRPS and SSR scores are presented in~\crefrange{fig:baltic_metrics_surface}{fig:baltic_metrics_uo} in the appendix.
\begin{figure}[tbp]
    \centering
    \includegraphics[width=\textwidth]{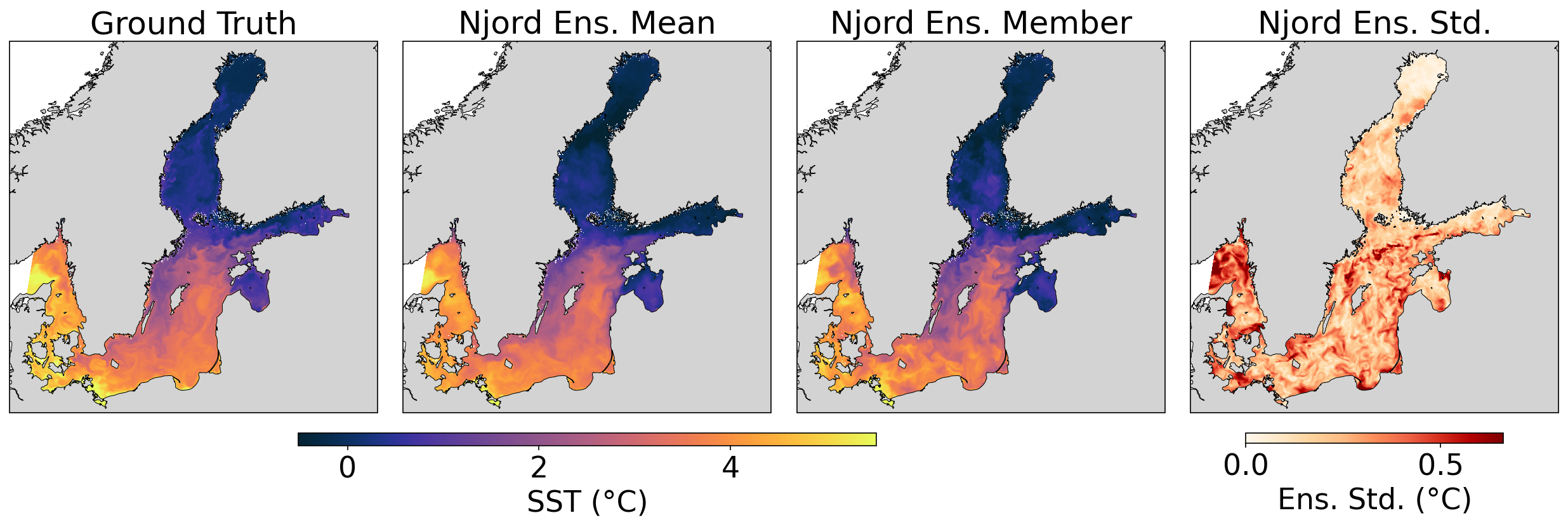}
    \caption{Baltic Sea \gls{SST} at \SI{10}{d} lead time, initialized 2024-03-05. Ground truth is NEMO analysis.
    }
    \label{fig:baltic_sst}
\end{figure}
Spatial \gls{SST} fields at a 10-day lead time are shown in \cref{fig:baltic_sst}. The ensemble mean captures the large-scale thermal structure but appears smoother than the ground truth. Individual ensemble members recover sharper gradients and localized features. The ensemble spread highlights regions of increased uncertainty, particularly in the Danish Straits (Kattegat and Skagerrak), where exchange with the North Sea introduces strong variability. The standard deviation is zero in ice-covered regions such as the very north of the Baltic Sea during winter.
All additional results and ablations can be found in \cref{sec:additional_results}.

\section{Discussion}
\label{sec:discussion}
\paragraph{Limitations.}
While the current number of predicted depth levels (5 for the Baltic Sea and 6 for the global ocean) is sufficient to benchmark the model, future probabilistic forecasting models should extend to a larger portion of the vertical column. The horizontal resolution can also be increased to the native \SI{0.083}{\degree} in the global case, to be able to resolve even finer ocean dynamics. We note that Njord-Baltic exhibits some visual artifacts in \gls{SLA} standard deviation fields, potentially due to higher noise-to-signal ratio in the satellite-derived ground-truth \gls{SLA} training targets. While we attempted to make Njord produce as well-calibrated ensembles as possible, there remains room for improvement, particularly by increasing the ensemble spread at early lead times for the global model and at longer lead times for the regional model. As an ensemble forecasting model, computing the minimum-error ensemble mean requires sampling multiple ensemble members, resulting in a computational cost proportionally higher than that of deterministic models. Given the fast and parallelizable sampling, this is a small price for reliable uncertainty estimates.

\paragraph{Conclusion.}
We have introduced \emph{Njord}, the first probabilistic generative machine learning model for global and regional ocean forecasting. By combining hierarchical \glspl{GNN} with a flexible K-means cluster mesh construction, we successfully scale generative ensemble modeling to high-resolution ocean grids with irregular coastal geometries. 
The results show that Njord not only matches or exceeds the accuracy of deterministic machine learning models and traditional numerical systems, but also provides uncertainty estimates for the predictions. This is important for operational use, where understanding forecast uncertainties in the ocean currents or advancing sea ice can help improve decision making.
Avenues for future work include scaling the approach to more vertical levels, additional variables, shorter timesteps and longer timescales.
Especially interesting is the prospect of combined oceanic and atmospheric modeling into a coupled ensemble system.

\begin{ack}
    The authors would like to thank Simon Adamov (ETHZ/MeteoSwiss) for assistance with the atmospheric forcing data and Lars Axell and Erik Mulder (SMHI) for useful discussions. This research was supported by the ETH AI Center through an ETH AI Center postdoctoral fellowship to Joel Oskarsson, the Research Council of Finland (grant no: 361902), the Swedish Research Council (grant no: 2020-04122, 2024-05011), the Wallenberg AI, Autonomous Systems and Software Program (WASP) funded by the Knut and Alice Wallenberg Foundation, and the Excellence Center at Linköping--Lund in Information Technology (ELLIIT). Daniel Holmberg acknowledges support from the Fulbright-KAUTE Foundation Award for conducting research at UC Santa Barbara. Our computations were enabled by the LUMI supercomputer, owned by the EuroHPC Joint Undertaking and hosted by CSC–IT Center for Science, and the Berzelius resource at the National Supercomputer Centre, provided by the Knut and Alice Wallenberg Foundation. This work was supported as part of the Swiss AI Initiative by a grant from the Swiss National Supercomputing Centre (CSCS) under project ID a122 on Alps.
\end{ack}

\bibliographystyle{unsrtnat} 
\bibliography{references}


\appendix
\crefalias{section}{appendix}
\crefalias{subsection}{appendix}

\section{Model Details}
\label{app:model-details}

\subsection{Graph-EFM details}
We adopt the probabilistic framework of Graph-EFM~\cite{oskarsson2024probabilistic}, a latent variable model in which stochasticity is introduced through latent variables $Z$ defined on the mesh graph. The generative model factorizes as:
\begin{equation}
    \hat{p}_\theta(X^{1:T} \mid X^{-1:0}, F^{-p+1:T}) = \prod_{t=1}^{T} \int p_\theta(X^t \mid Z^t, X^{t-2:t-1}, F^{t-2:t})\, p_\theta(Z^t \mid X^{t-2:t-1}, F^{t-2:t})\, \mathrm{d}Z^t
    \label{eq:generative}
\end{equation}
where $X^{t-2:t-1}$ denotes the previous states and $F^{t-2:t}$ all forcing inputs (boundary, atmosphere) available at time $t$. The prior $p_\theta(Z^t \mid X^{t-2:t-1}, F^{t-2:t})$ is a diagonal Gaussian parameterized by a \gls{GNN} that encodes the previous state onto the graph. The likelihood $p_\theta(X^t \mid Z^t, X^{t-2:t-1}, F^{t-2:t})$ is parameterized by the decoder, which combines the latent sample $Z^t$ with the encoded previous state to produce a predictive Gaussian over the next state. 
Since the marginal likelihood is intractable, the \gls{ELBO} is optimized:
\begin{equation}
    \begin{split}
    \log \hat{p}_\theta(X^t \mid X^{t-2:t-1}, F^{t-2:t}) \geq &\underbrace{\mathbb{E}_{q_\phi(Z^t \mid X^{t-2:t}, F^{t-2:t})} \!\left[\log p_\theta(X^t \mid Z^t, X^{t-2:t-1}, F^{t-2:t})\right]}_{\text{reconstruction / likelihood}} \\
    &- \underbrace{\lambda_{\text{KL}}\, D_{\mathrm{KL}}\!\left(q_\phi(Z^t \mid X^{t-2:t}, F^{t-2:t}) \,\|\, p_\theta(Z^t \mid X^{t-2:t-1}, F^{t-2:t})\right)}_{\text{KL divergence}}
    \end{split}
    \label{eq:elbo}
\end{equation}
where $q_\phi(Z^t \mid X^{t-2:t}, F^{t-2:t})$ is the approximate posterior (encoder), conditioned on both the previous and current state, and $\lambda_{\text{KL}}$ is a weighting factor for the KL term.
\paragraph{Encoder.}
The encoder takes as input the grid embedding conditioned on both the previous state and the target state $X^t$, maps it to the mesh via the grid-to-mesh \gls{GNN}, processes it through the mesh hierarchy, and outputs the parameters $(\mu_q, \sigma_q)$ of a diagonal Gaussian over the latent variables at the top mesh level:
\begin{equation}
    q_\phi(Z^t \mid X^{t-2:t}, F^{t-2:t}) = \mathcal{N}\!\left(\mu_q(X^{t-2:t}, F^{t-2:t}),\, \mathrm{diag}\!\left(\sigma_q^2(X^{t-2:t}, F^{t-2:t})\right)\right)
\end{equation}
\paragraph{Prior.}
The prior is parameterized by a separate network with the same architecture as the encoder but conditioned only on the previous state:
\begin{equation}
    p_\theta(Z^t \mid X^{t-2:t-1}, F^{t-2:t}) = \mathcal{N}\!\left(g_\theta(X^{t-2:t-1}, F^{t-2:t}),\, I\right)
\end{equation}
\paragraph{Decoder.}
Given a latent sample $Z^t \sim q_\phi$ (for the \gls{ELBO}) or $Z^t \sim p_\theta$ (during inference, and for the \gls{CRPS} loss term), the decoder injects $Z^t$ into the mesh representation at the top level, processes it through the full mesh hierarchy (upward and downward sweeps), decodes back to the grid, and outputs the predicted residual
$\hat{r}^t$. 
The forecast is computed as a residual: $ \hat{X}^{t} = f_\theta\left(X^{t-2:t-1}, F^{t-2:t}, Z^t\right) = X^{t-1} + \hat{r}^t$.

\paragraph{Ensemble generation.}
At inference, ensemble members are generated by drawing independent samples $Z^t_1, \ldots, Z^t_M \sim p_\theta(Z^t \mid X^{t-2:t-1}, F^{t-2:t})$ from the prior. Each sample is decoded independently through the same decoder, producing an ensemble of forecasts $\{\hat{X}^{t}_{m}\}_{m=1}^M$. The latent variables are defined at the coarsest mesh graph level, which has relatively few nodes, so the cost of generating additional ensemble members is dominated by the decoder pass.

\subsection[GNN layer formulations]{\gls{GNN} layer formulations}
\label{sec:flex-gnn}

Graph-EFM uses two different kinds of \gls{GNN} layers: Interaction Networks and Propagation networks.
To increase scalability, we generalize these in Njord to have more flexible input dimensionalities.
All different kinds of \glspl{GNN} are described in this section.

\paragraph{Interaction Networks.}
The \gls{GNN} layers in the encode--process--decode architecture are based on Interaction Networks~\cite{battaglia2016interaction}. For a graph $\mathcal{G}=(\mathcal{V},\mathcal{E})$ with sender node representations~$\mathbf{H}^S$, receiver node representations~$\mathbf{H}^R$, and edge representations~$\mathbf{E}$, all sharing dimensionality~$d_z$, the update $\mathbf{H}^R, \mathbf{E} \leftarrow \mathrm{\gls{GNN}}(\mathcal{G},\mathbf{H}^S,\mathbf{E},\mathbf{H}^R)$ is
\begin{align}
    \tilde{e}_{\alpha\to\beta}
        &\leftarrow \mathrm{MLP}_e\!\bigl(
            e_{\alpha\to\beta},\,
            \mathbf{H}^S_\alpha,\,
            \mathbf{H}^R_\beta
        \bigr)
    \label{eq:in-msg}\\
    e_{\alpha\to\beta}
        &\leftarrow e_{\alpha\to\beta} + \tilde{e}_{\alpha\to\beta}
    \label{eq:in-edge}\\
    \mathbf{H}^R_\beta
        &\leftarrow \mathbf{H}^R_\beta
        + \mathrm{MLP}_a\!\!\left(
            \mathbf{H}^R_\beta,\;
            \sum_{\alpha\in\mathcal{N}_e(\beta)}
                \tilde{e}_{\alpha\to\beta}
        \right)
    \label{eq:in-node}
\end{align}
where $\mathcal{N}_e(\beta)=\{\alpha:(\alpha,\beta)\in\mathcal{E}\}$ are the incoming neighbors of node~$\beta$. The edge residual in Eq.~\eqref{eq:in-edge} allows edge representations to accumulate information across successive \gls{GNN} layers. These layers are used for same-level mesh processing throughout the hierarchy.

\paragraph{Propagation Networks.}
Interaction Networks are biased towards retaining existing receiver node representations~\cite{oskarsson2024probabilistic}: when MLPs are initialized with outputs near zero, Eqs.~\eqref{eq:in-edge}--\eqref{eq:in-node} produce no change to~$e_{\alpha\to\beta}$ or~$\mathbf{H}^R_\beta$. To encourage information flow from senders to receivers, Propagation Networks~\cite{oskarsson2024probabilistic} modify Eqs.~\eqref{eq:in-msg}--\eqref{eq:in-node} to
\begin{align}
    \tilde{e}_{\alpha\to\beta}
        &\leftarrow
        \mathbf{H}^S_\alpha
        + \mathrm{MLP}_e\!\bigl(
            e_{\alpha\to\beta},\,
            \mathbf{H}^S_\alpha,\,
            \mathbf{H}^R_\beta
        \bigr)
    \label{eq:flex-pn-msg}\\[4pt]
    \tilde{\mathbf{H}}^R_\beta
        &\leftarrow
        \frac{1}{|\mathcal{N}_e(\beta)|}
        \sum_{\alpha\in\mathcal{N}_e(\beta)}
            \tilde{e}_{\alpha\to\beta}
    \label{eq:flex-pn-aggr}\\[4pt]
    \mathbf{H}^R_\beta
        &\leftarrow
        \tilde{\mathbf{H}}^R_\beta
        + \mathrm{MLP}_a\!\bigl(
            \mathbf{H}^R_\beta,\,
            \tilde{\mathbf{H}}^R_\beta
        \bigr)
    \label{eq:flex-pn-node}
\end{align}
For MLPs initialized near zero, this reduces to averaging neighboring sender representations, encouraging propagation by construction. These layers are used for inter-level mappings in the mesh hierarchy.

\paragraph{Flexible Interaction Networks.}
The standard Interaction Networks require all representations to share dimensionality~$d_z$. We relax this constraint by allowing separate dimensionalities $d_s$, $d_r$, and $d_e$ for sender nodes, receiver nodes, and edges, respectively. 
The edge MLP maps from $d_e + d_s + d_r$ to $d_r$, and the aggregation MLP maps from $d_r + d_r$ to $d_r$, so that the output matches the receiver node dimensionality. Because the edge message dimensionality~$d_r$ differs from the edge input dimensionality~$d_e$, the edge residual update in Eq.~\eqref{eq:in-edge} is omitted; only the receiver node representations~$\mathbf{H}^R$ are returned.
This is not a problem, as the updated edge representations are not needed where these layers are used in Njord.

\paragraph{Flexible Propagation Networks.}
Similarly, we extend Propagation Networks to heterogeneous dimensions. 
These layers are slightly more restrictive, as the updated receiver representation will always have the same dimensionality as the sender nodes.
Still, the key flexibility is to allow for a different edge dimensionality.
In the flexible Propagation Network the edge MLP maps from $d_e + d_s + d_r$ to $d_s$, and the aggregation MLP maps from $d_s + d_r$ to $d_s$, so that the output matches the sender node dimensionality. As with the Flexible Interaction Network, the edge residual update can be omitted without issues.

\subsection{Scaling to ocean grids}
\label{sec:app_scaling}
For comparison, the Graph-EFM weather model~\cite{oskarsson2024probabilistic} is applied on 29\,040 global grid nodes and 63\,784 regional grid nodes. Our global ocean model operates on 676\,736 grid nodes out of a 680 $\times$ 1440 bounding box (979~200 total); and the Baltic Sea regional model on 147\,701 grid nodes out of a 738 $\times$ 763 bounding box (563~094 total). In all cases the \gls{GNN} processes only the interior sea nodes. This order-of-magnitude increase in spatial scale produces grid-to-mesh and mesh-to-grid graphs with $\mathcal{O}(10^6)$ edges (Tables~\ref{tab:global_cluster}--\ref{tab:global_icosahedral}), making the capacity of the edge-embedding MLPs the dominant memory bottleneck. Since the static edge features are low-dimensional (3--4 features), we reduce the hidden and output dimensionality of the graph encoding and decoding edge MLPs from 256 (as in SeaCast~\cite{holmberg2025accurate}) to $d_e = 32$, with no noteworthy effect on forecast skill while yielding significant savings in compute and memory. Similarly, the grid and bottom-mesh-level node representations use dimensionality $d_g = 128$, compared to $d_z = 256$ for the mesh processing layers. Linear projection layers map between $d_g$ and $d_z$ at the encoder--processor and processor--decoder boundaries. Njord is configured to use 6 processing layers, which amount to 22M trainable parameters in total. All of these choices, together with gradient checkpointing~\cite{chen2016training} at each autoregressive step, enable training on the large ocean grids.

\clearpage

\section{Data Details}
\label{app:data-details}

Tables~\ref{tab:variables_global} and~\ref{tab:variables_baltic} detail the comprehensive set of variables used to train and evaluate the global ocean model and the Baltic Sea regional model, respectively. These encompass the internal physical state variables predicted by the model, alongside the external conditioning inputs, which include atmospheric forcings, lateral boundary conditions (for the regional model), and static geographic fields. We train the global model using the GLORYS12 global ocean reanalysis~\cite{jean2021copernicus} and finetune on operational GLO12 analysis data~\cite{lellouche2023evolution}. The regional model is trained on the \SI{2}{\kilo\meter} Baltic Sea Physics Reanalysis~\cite{madec2015nemo}, using GLORYS12 for lateral boundary forcing. During evaluation, regional boundaries are forced by GLO12 forecasts sourced from OceanBench~\cite{el2025oceanbench}, which also supplies our global baseline model data. For both configurations, surface atmospheric forcing uses the ERA5 reanalysis~\citep{hersbach2020era5} during training and operational 10-day IFS forecasts~\cite{ifs} during evaluation\footnote{Data usage and licensing: OceanBench data is provided under the EUPL-1.2 license. ERA5 data is provided by the Copernicus Climate Change Service under the ECMWF Copernicus License. IFS operational data is provided by ECMWF or through OceanBench. Oceanographic data was obtained using E.U. Copernicus Marine Service Information under the Copernicus Marine Service License (DOIs: 10.48670/moi-00021, 10.48670/moi-00016, 10.48670/moi-00013, 10.48670/moi-00010).}.

\begin{table}[tbh]
\centering
\caption{Variables, static fields, and forcing features for the global ocean dataset.}
\begin{tabular}{llll}
\toprule
& Abbreviation & Unit & Vertical level \\
\midrule
\textit{State variables} & & & \\
\midrule
Sea surface height above geoid & \texttt{zos} & m & Sea surface \\
Sea ice area fraction & \texttt{siconc} & -- & Sea surface \\
Sea ice thickness & \texttt{sithick} & m & Sea surface \\
Zonal sea water velocity & \texttt{uo} & m/s & 0, 47, 92, 222, 318, 541\,m \\
Meridional sea water velocity & \texttt{vo} & m/s & 0, 47, 92, 222, 318, 541\,m \\
Sea water salinity & \texttt{so} & PSU & 0, 47, 92, 222, 318, 541\,m \\
Sea water potential temperature & \texttt{thetao} & \textdegree C & 0, 47, 92, 222, 318, 541\,m \\
\midrule
\textit{Forcing fields} & & & \\
\midrule
Sea floor depth below geoid & \texttt{deptho} & m & Sea floor \\
Mean dynamic topography & \texttt{mdt} & m & Sea surface \\
Sine of longitude & \texttt{sin\_lon} & -- & -- \\
Cosine of longitude & \texttt{cos\_lon} & -- & -- \\
Sine of latitude & \texttt{sin\_lat} & -- & -- \\
Cosine of latitude & \texttt{cos\_lat} & -- & -- \\
Distance to coast & \texttt{coast\_dist} & m & -- \\
Sine of time of year & \texttt{sin\_t} & -- & -- \\
Cosine of time of year & \texttt{cos\_t} & -- & -- \\
\midrule
\textit{Atmospheric forcing} & & & \\
\midrule
2-meter air temperature & \texttt{sotemair} & \textdegree C & Sea surface \\
Zonal 10-meter wind & \texttt{sowinu10} & m/s & Sea surface \\
Meridional 10-meter wind & \texttt{sowinv10} & m/s & Sea surface \\
Downward shortwave radiation flux & \texttt{sosudosw} & W/m\textsuperscript{2} & Sea surface \\
Downward longwave radiation flux & \texttt{sosudolw} & W/m\textsuperscript{2} & Sea surface \\
Total precipitation rate & \texttt{sowaprec} & kg\,m\textsuperscript{-2}\,s\textsuperscript{-1} & Sea surface \\
2-meter dew point temperature & \texttt{sod2m} & \textdegree C & Sea surface \\
Mean sea level pressure & \texttt{somslpre} & Pa & Sea surface \\
\bottomrule
\end{tabular}
\label{tab:variables_global}
\end{table}

\begin{table}[tbh]
\centering
\caption{Variables, static fields, and forcing features for the Baltic Sea dataset.}
\begin{tabular}{llll}
\toprule
& Abbreviation & Unit & Vertical level \\
\midrule
\textit{State variables} & & & \\
\midrule
Sea level anomaly & \texttt{sla} & m & Sea surface \\
Sea ice area fraction & \texttt{siconc} & -- & Sea surface \\
Sea ice thickness & \texttt{sithick} & m & Sea surface \\
Zonal sea water velocity & \texttt{uo} & m/s & 1, 9, 28, 47, 91\,m \\
Meridional sea water velocity & \texttt{vo} & m/s & 1, 9, 28, 47, 91\,m \\
Sea water salinity & \texttt{so} & PSU & 1, 9, 28, 47, 91\,m \\
Sea water potential temperature & \texttt{thetao} & \textdegree C & 1, 9, 28, 47, 91\,m \\
\midrule
\textit{Forcing fields} & & & \\
\midrule
Sea floor depth below geoid & \texttt{deptho} & m & Sea floor \\
Mean dynamic topography & \texttt{mdt} & m & Sea surface \\
Projected x-coordinate & \texttt{x\_coord} & m & -- \\
Projected y-coordinate & \texttt{y\_coord} & m & -- \\
Distance to coast & \texttt{coast\_dist} & m & -- \\
Sine of time of year & \texttt{sin\_t} & -- & -- \\
Cosine of time of year & \texttt{cos\_t} & -- & -- \\
\midrule
\textit{Atmospheric forcing} & & & \\
\midrule
2-meter air temperature & \texttt{sotemair} & \textdegree C & Sea surface \\
Zonal 10-meter wind & \texttt{sowinu10} & m/s & Sea surface \\
Meridional 10-meter wind & \texttt{sowinv10} & m/s & Sea surface \\
Downward shortwave radiation flux & \texttt{sosudosw} & W/m\textsuperscript{2} & Sea surface \\
Downward longwave radiation flux & \texttt{sosudolw} & W/m\textsuperscript{2} & Sea surface \\
Total precipitation rate & \texttt{sowaprec} & kg\,m\textsuperscript{-2}\,s\textsuperscript{-1} & Sea surface \\
2-meter dew point temperature & \texttt{sod2m} & \textdegree C & Sea surface \\
Mean sea level pressure & \texttt{somslpre} & Pa & Sea surface \\
\midrule
\textit{Boundary forcing} & & & \\
\midrule
Sea surface height above geoid & \texttt{zos} & m & Sea surface \\
Zonal sea water velocity & \texttt{uo} & m/s & 0, 47, 92\,m \\
Meridional sea water velocity & \texttt{vo} & m/s & 0, 47, 92\,m \\
Sea water salinity & \texttt{so} & PSU & 0, 47, 92\,m \\
Sea water potential temperature & \texttt{thetao} & \textdegree C & 0, 47, 92\,m \\
\bottomrule
\end{tabular}
\label{tab:variables_baltic}
\end{table}

\clearpage

\section{Graph details}
\label{app:graph-details}

\subsection{Cluster-based graph construction}
\label{sec:app_graph_construction}
Previous global graph-based forecasting models have used icosahedral meshes~\cite{lam2023learning,oskarsson2024probabilistic} for constructing the graph. Icosahedral meshes are constructed by iteratively subdividing an icosahedron, with each subdivision quadrupling the number of nodes. This coarse refinement factor creates a practical limitation because the jumps in resolution between different splits are large, e.g., 7 splits produce 115\,016 nodes after masking out land, compared to 28\,753 for 6 splits. We had to restrict our comparison with icosahedral meshes to 6 splits as it was not possible to fit higher than that in memory during training. 
The icosahedral structure both limits the ability to choose the number of mesh nodes, and is poorly adapted to the irregular geometry of the ocean surface.
For better adapting to this geometry, we propose using spatial K-means clustering with Delaunay triangulation to construct hierarchical meshes.

For the global model, we use spherical K-means clustering of the ocean grid point 3D Cartesian coordinates, with latitude-based area weights to ensure equitable spatial coverage. Same-level edges are constructed via spherical Delaunay triangulation (computed as the convex hull of the 3D points on the unit sphere), followed by land-crossing edge filtering. The refinement factor between levels becomes a continuous parameter rather than a fixed quadrupling, enabling finer control over the mesh hierarchy. 
The first mesh level is obtained by clustering the $N$ ocean grid nodes into $N / r_0$ clusters, where $r_0$ is the grid-to-first-mesh refinement factor. Subsequent levels cluster the previous level's nodes by a refinement factor $r$.
In our global model, $r_0 = 20$ is used from the grid to the first mesh level and produces 33\,777 mesh nodes, compared to 28\,753 or 115\,016 for 6 or 7 icosahedral splits. Subsequent levels cluster the previous level's nodes by a refinement factor $r = 4$. We apply land-crossing edge filtering so that the meshes conform to coastlines, bays, and straits.

\begin{figure}[ht]
  \centering
  \includegraphics[width=.99\textwidth]{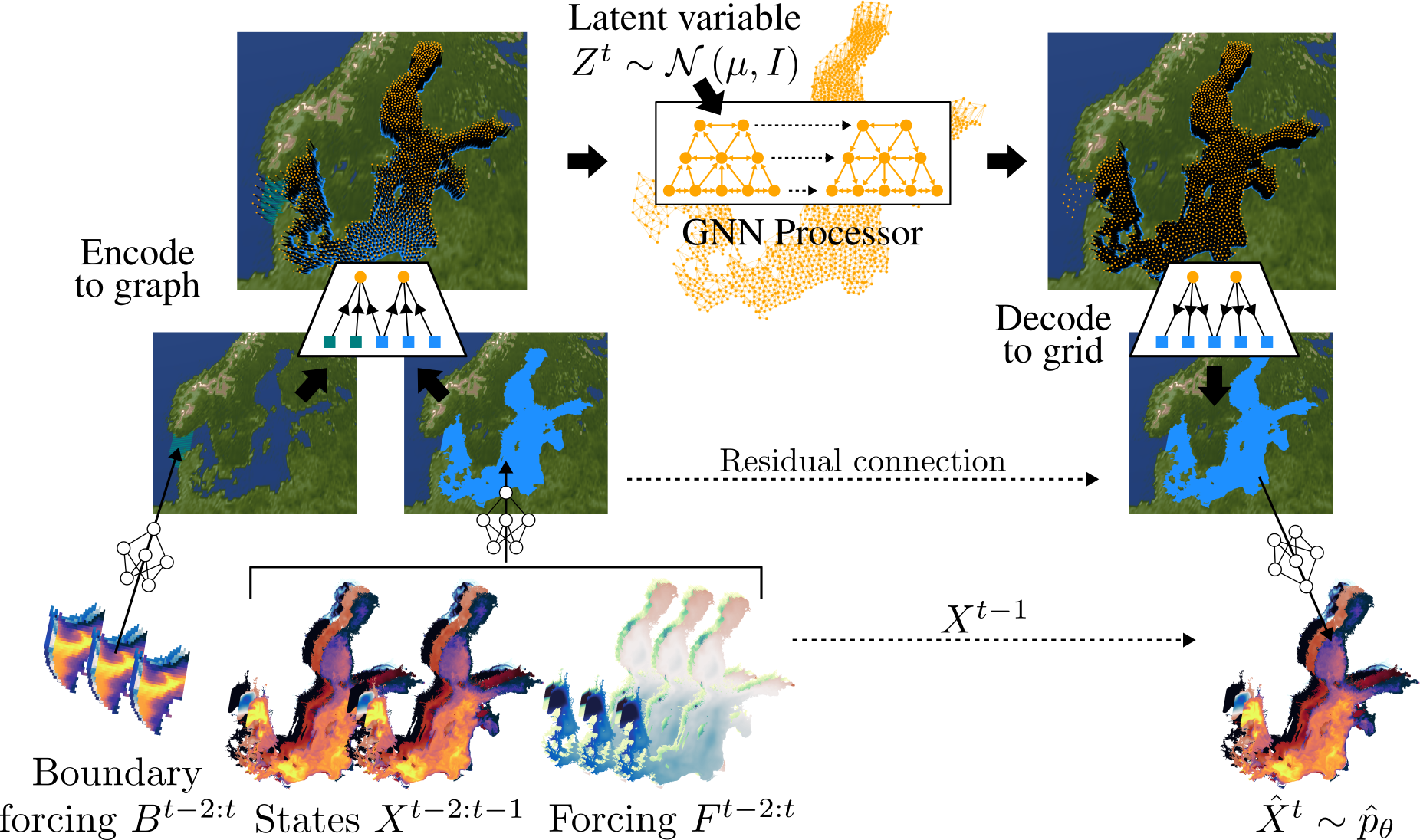}
  \caption{
  One-step prediction in the Njord-Baltic model. 
  Residuals are predicted at time $t$, which are then added to the previous state $X^{t-1}$ in order to produce the sample $\hat{X}^{t}$.
  }
  \label{fig:baltic_model_diagram}
\end{figure}

For regional graph-based models, quadrilateral meshes have been used~\cite{oskarsson2023graph, holmberg2024regional}. While more flexible in refinement, they do not necessarily line up well with irregular coastlines. We use Euclidean K-means applied to the ocean grid node positions. For the regional mesh the grid-to-first-mesh refinement factor is chosen as $r_0 = 9$. Subsequent levels also use a factor $r = 9$. This makes the mesh comparable in size to the quadrilateral mesh using a $3\times3$ coarsening factor. The Mesh edges at each level are constructed via 2D Delaunay triangulation of the cluster centers, and edges crossing land areas are filtered out. Inter-level edges connect each node to its nearest neighbor at the adjacent level.
The full Njord-Baltic framework, including the regional cluster graph, is shown in \cref{fig:baltic_model_diagram}. 

\begin{figure}[tbh]
  \centering
  \begin{subfigure}[b]{0.5\textwidth}
    \centering
    \includegraphics[width=\textwidth]{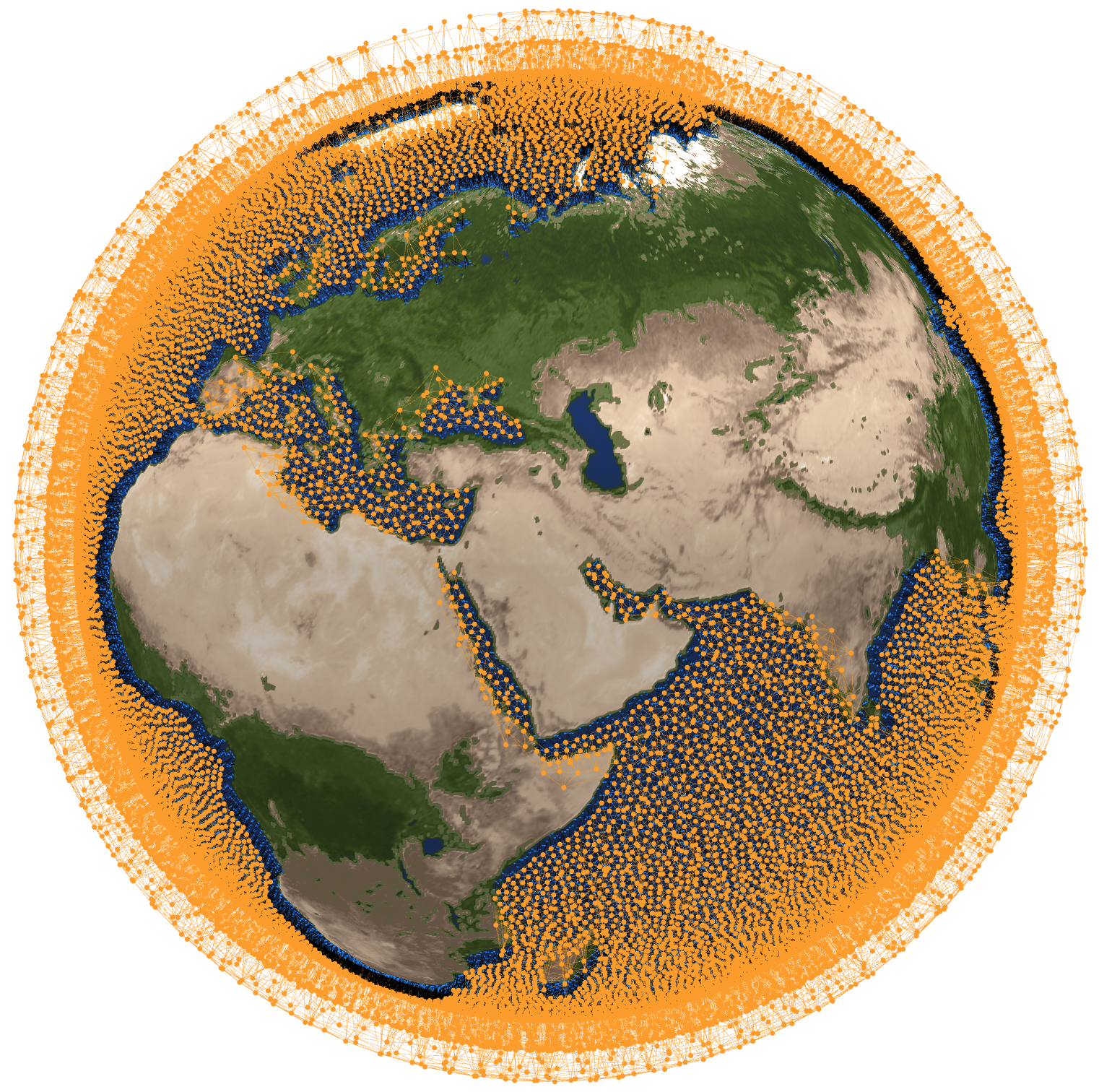}
    \caption{Global cluster graph.}
    \label{fig:global_cluster}
  \end{subfigure}%
  \hfill%
  \begin{subfigure}[b]{0.5\textwidth}
    \centering
    \includegraphics[width=\textwidth]{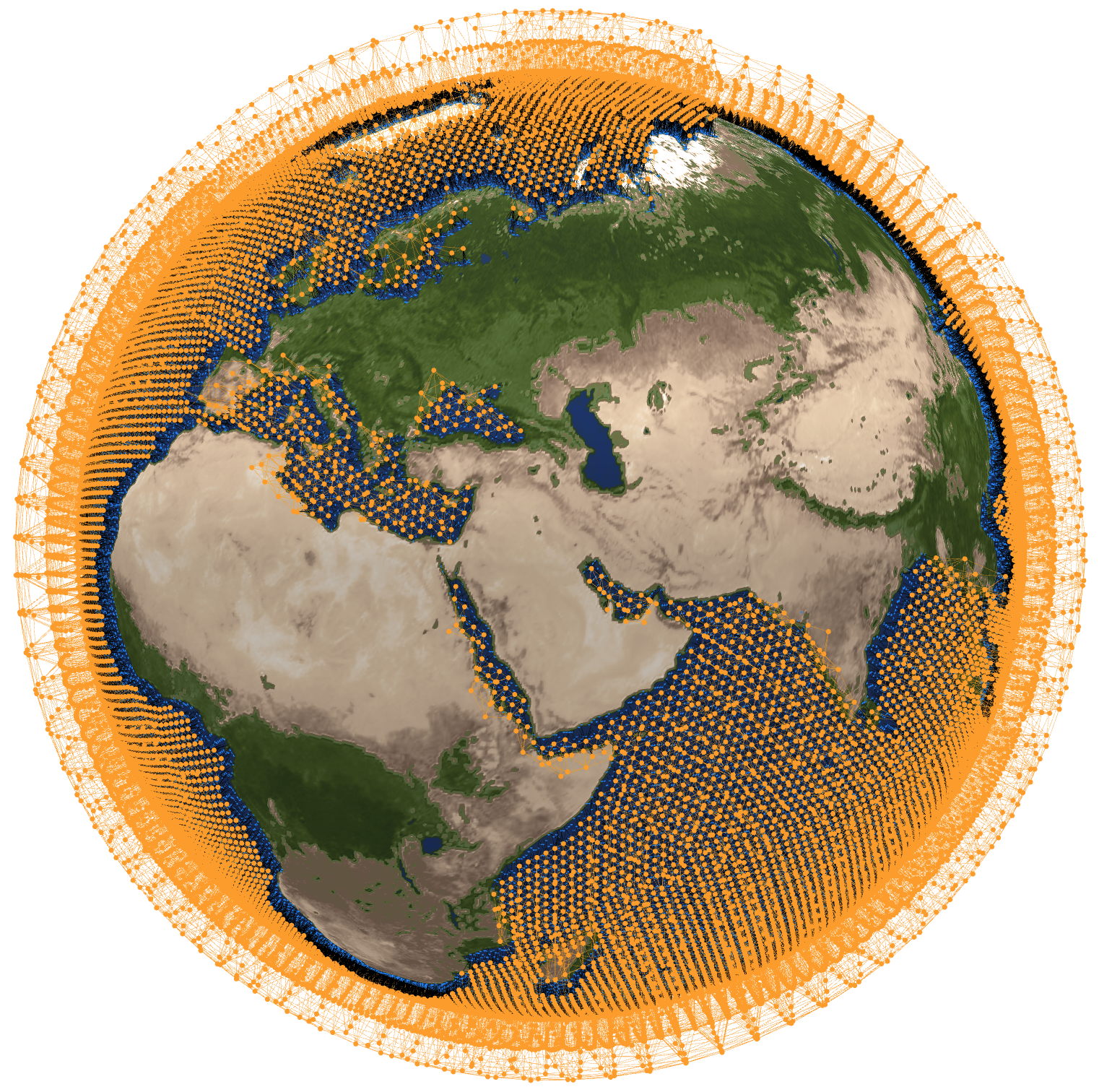}
    \caption{Global icosahedral graph.}
    \label{fig:global_icosahedral}
  \end{subfigure}
  \caption{Global graphs used by Njord, with grid nodes in blue, encoding/decoding edges in black, and the hierarchical meshes colored in yellow.}
  \label{fig:full_global_graphs}
\end{figure}

\begin{figure}[tbh]%
  \centering%
  \begin{subfigure}{0.48\textwidth}
    \includegraphics[width=\textwidth]{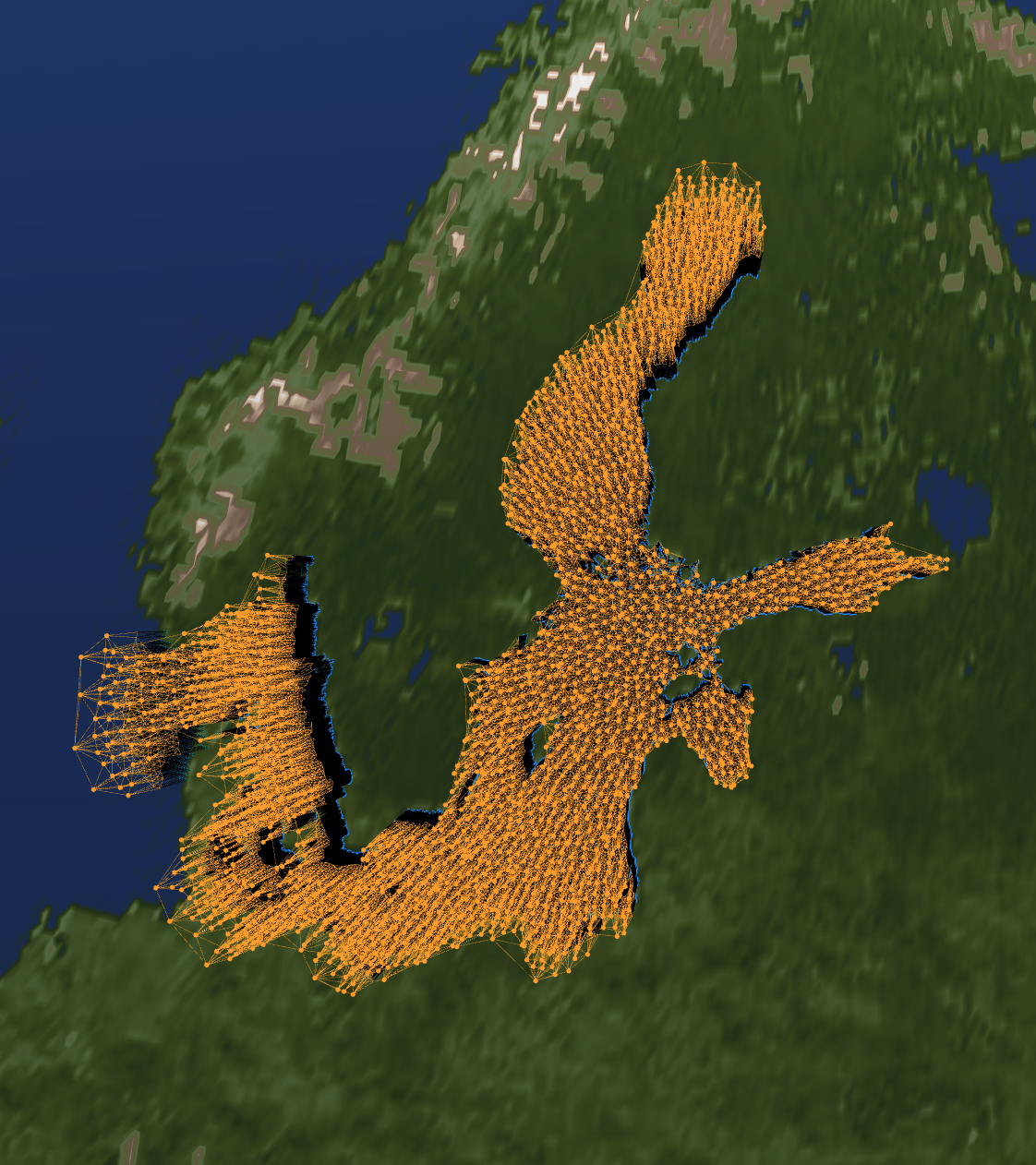}
    \caption{Baltic Sea cluster graph.}%
  \end{subfigure}
  \begin{subfigure}{0.48\textwidth}
    \includegraphics[width=\textwidth]{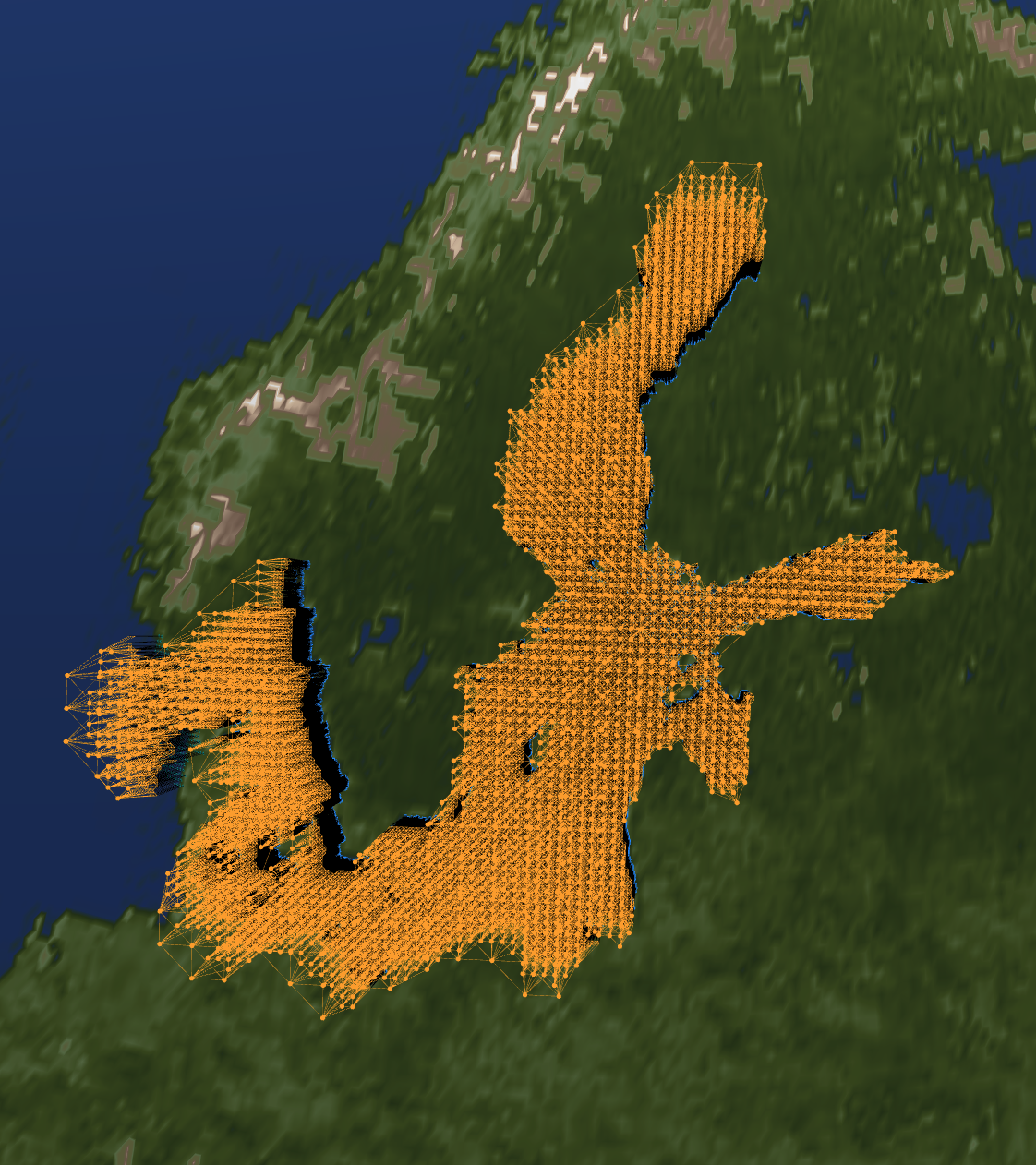}
    \caption{Baltic Sea quadrilateral graph.}%
  \end{subfigure}
  \caption{Regional graphs used by Njord, with grid nodes in blue, M2G and G2M edges in black, and the hierarchical meshes colored in yellow.}
  \label{fig:full_baltic_graphs}
\end{figure}

\begin{figure}[tbh]
  \centering
  \begin{subfigure}{.4\textwidth}
    \includegraphics[width=\textwidth]{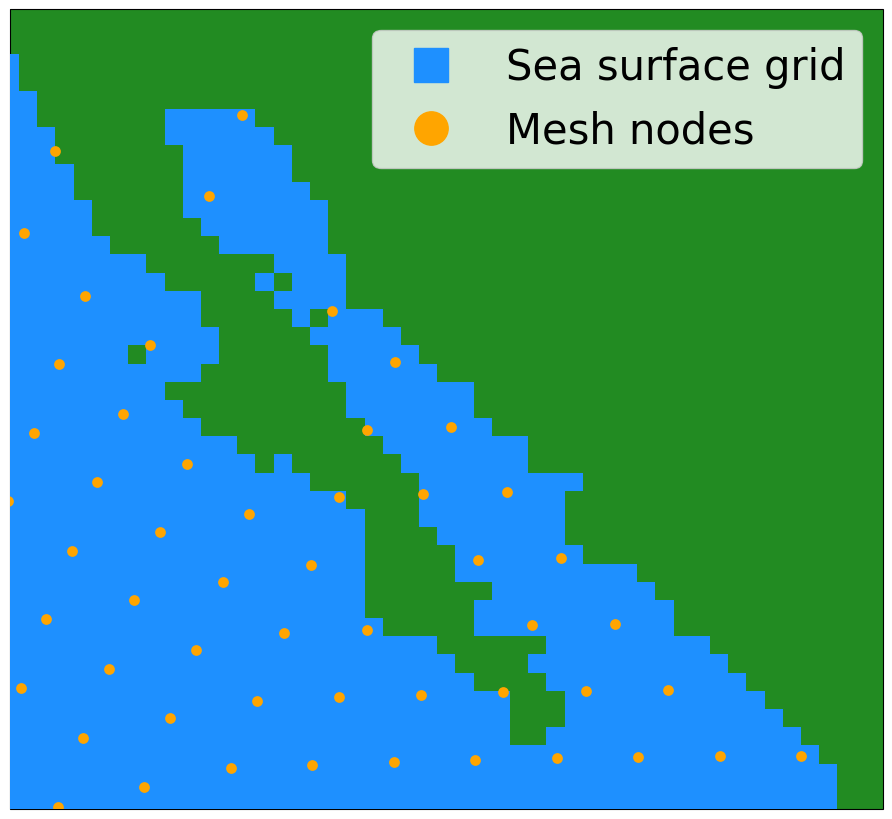}
    \caption{Icosahedral, level 0} 
  \end{subfigure}
  \begin{subfigure}{.4\textwidth}
    \includegraphics[width=\textwidth]{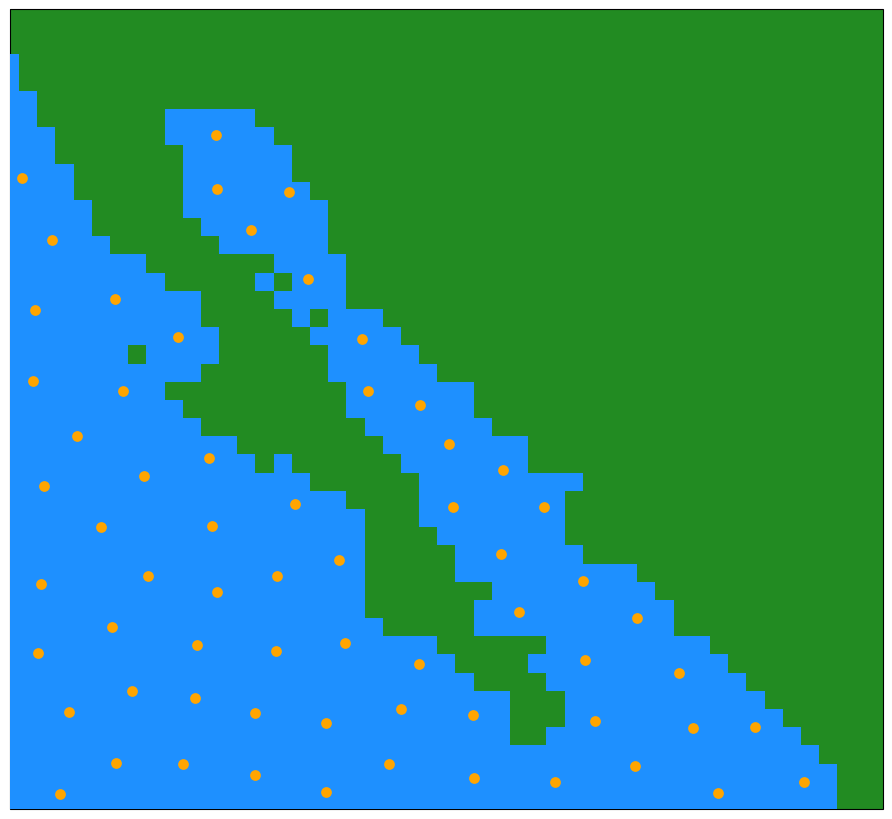}
    \caption{Cluster, level 0} 
  \end{subfigure}
  \begin{subfigure}{.4\textwidth}
    \includegraphics[width=\textwidth]{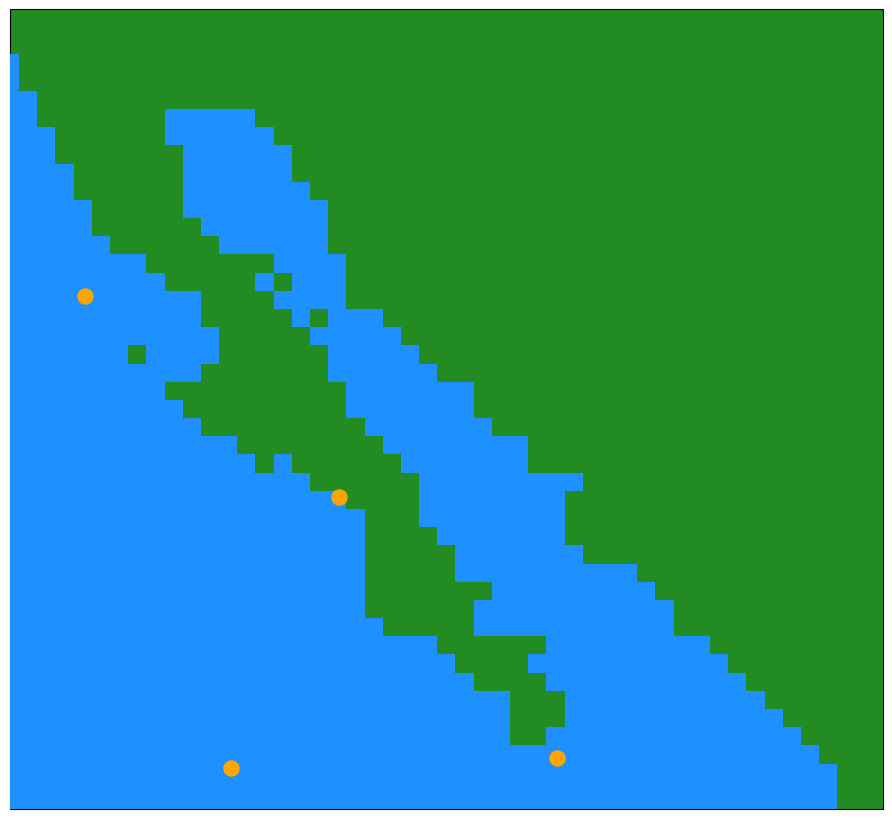}
    \caption{Icosahedral, level 2} 
  \end{subfigure}
  \begin{subfigure}{.4\textwidth}
    \includegraphics[width=\textwidth]{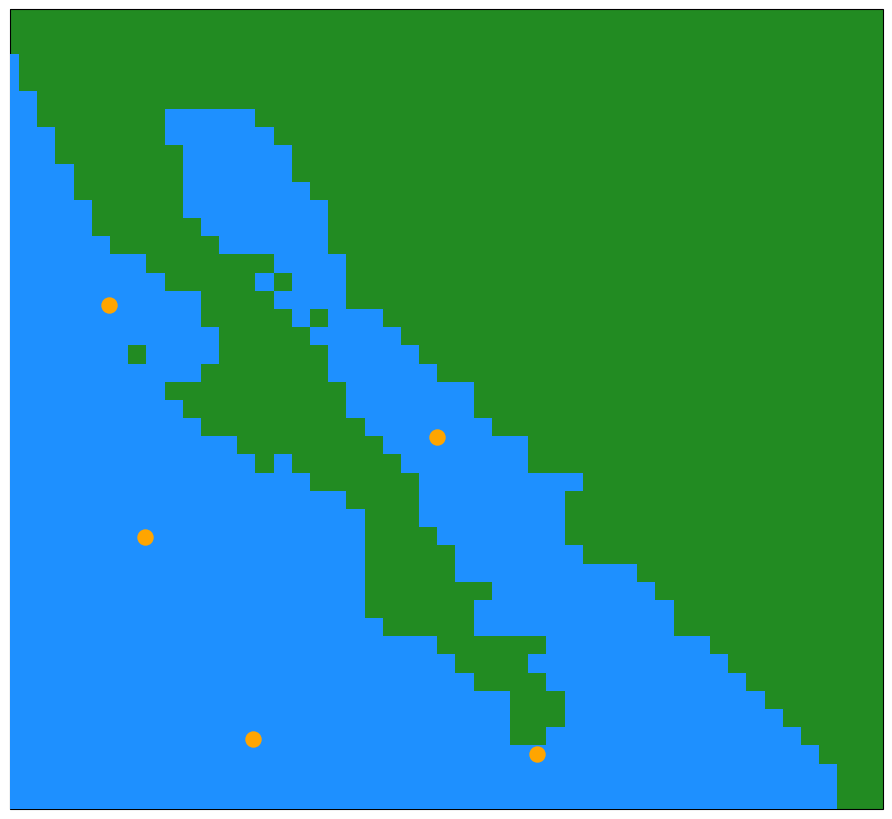}
    \caption{Cluster, level 2} 
  \end{subfigure}
  \caption{
  Example of mesh node placement in the Gulf of California (latitude $22^\circ$N--$33^\circ$N, longitude
  $117^\circ$W--$105^\circ$W).
  }
  \label{fig:graph_global_california}
\end{figure}

\begin{figure}[tbh]
    \centering
    \begin{subfigure}{.24\textwidth}
      \includegraphics[width=\textwidth]{figures/graph_geometry_examples/global/hierarchical_northern_red_sea_level0_legend.png}
      \caption{Icosahedral, level 0}
    \end{subfigure}
    \begin{subfigure}{.24\textwidth}
      \includegraphics[width=\textwidth]{figures/graph_geometry_examples/global/cluster_northern_red_sea_level0.png}
      \caption{Cluster, level 0}
    \end{subfigure}
    \begin{subfigure}{.24\textwidth}
      \includegraphics[width=\textwidth]{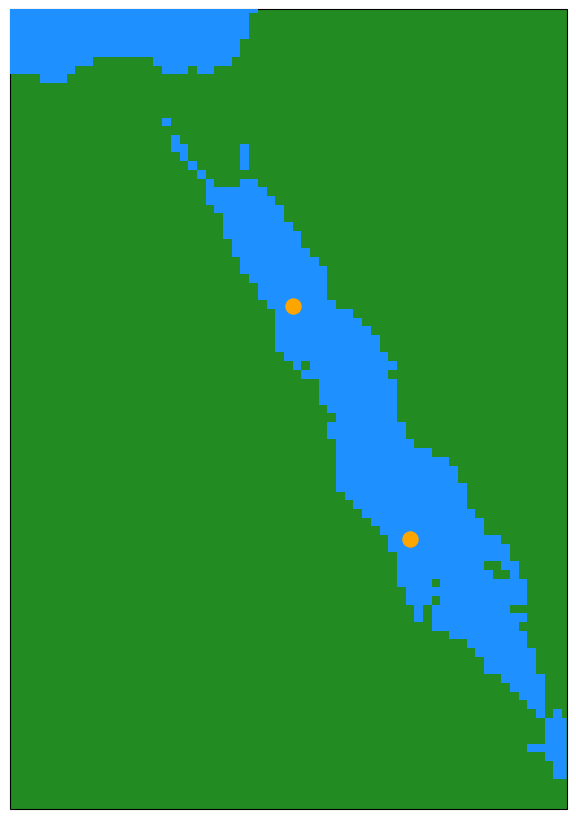}
      \caption{Icosahedral, level 2}
    \end{subfigure}
    \begin{subfigure}{.24\textwidth}
      \includegraphics[width=\textwidth]{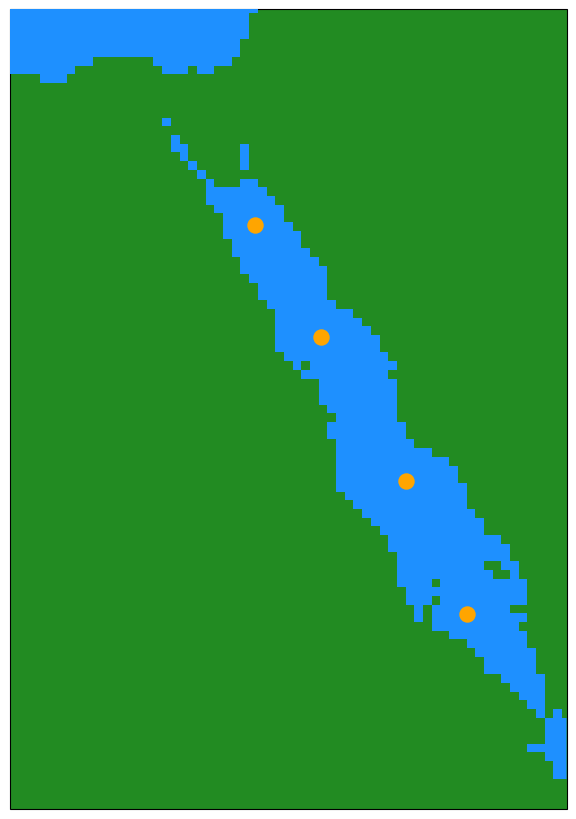}
      \caption{Cluster, level 2}
    \end{subfigure}
    \caption{
    Example of mesh node placement in the northern Red Sea and Suez Canal (latitude $10^\circ$N--$33^\circ$N,
  longitude
    $28^\circ$E--$44^\circ$E).
    }
    \label{fig:graph_global_red_sea}
\end{figure}

\begin{figure}[tbh]
\centering
\begin{subfigure}{.4\textwidth}
  \includegraphics[width=\textwidth]{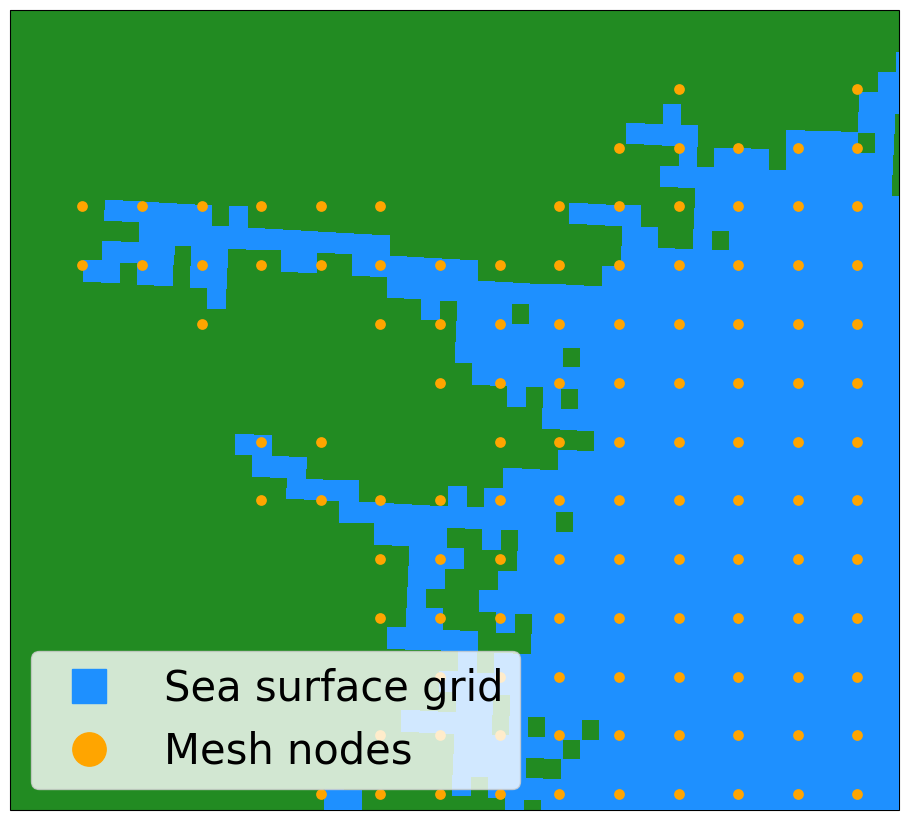}
  \caption{Quadrilateral, level 0}
\end{subfigure}
\begin{subfigure}{.4\textwidth}
  \includegraphics[width=\textwidth]{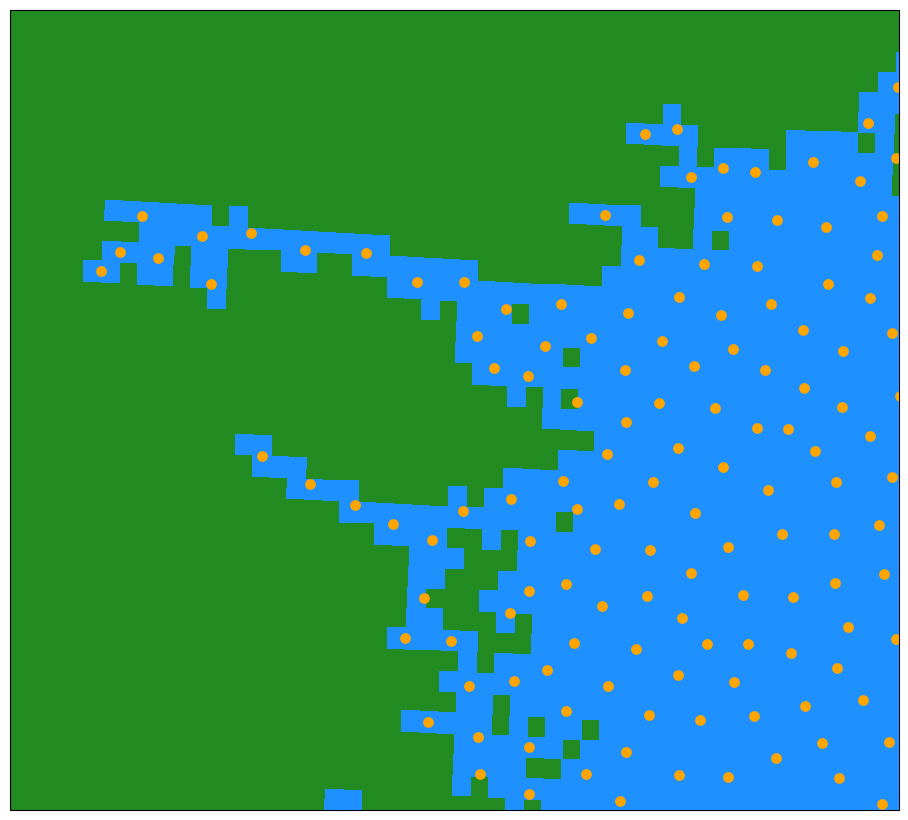}
  \caption{Cluster, level 0}
\end{subfigure}
\begin{subfigure}{.4\textwidth}
  \includegraphics[width=\textwidth]{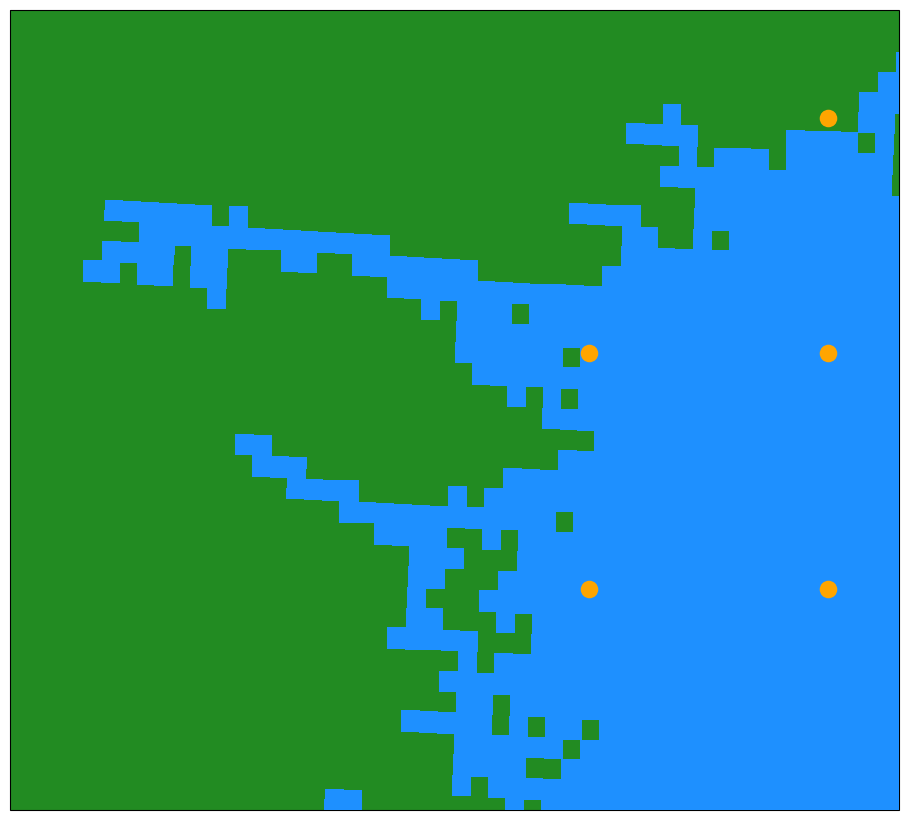}
  \caption{Quadrilateral, level 2}
\end{subfigure}
\begin{subfigure}{.4\textwidth}
  \includegraphics[width=\textwidth]{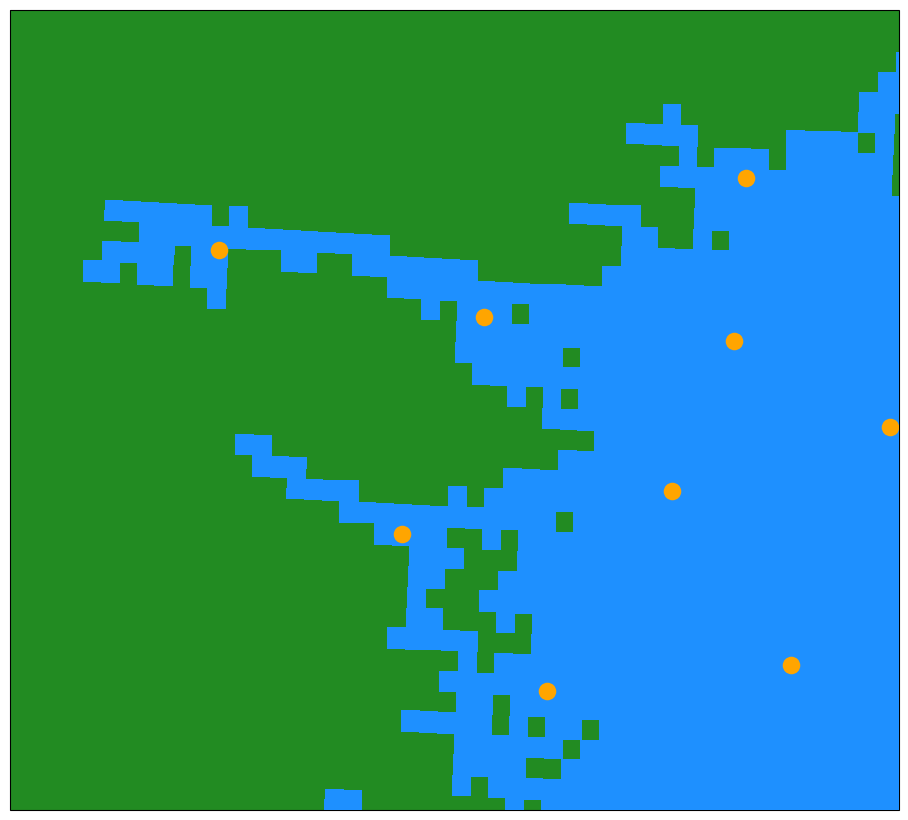}
  \caption{Cluster, level 2}
\end{subfigure}
\caption{
Example of mesh node placement in the Bråviken bay and Östergötland Archipelago, on the Swedish east coast.
}
\label{fig:graph_baltic_braviken}
\end{figure}

\begin{figure}[tbh]
\centering
\begin{subfigure}{.4\textwidth}
  \includegraphics[width=\textwidth]{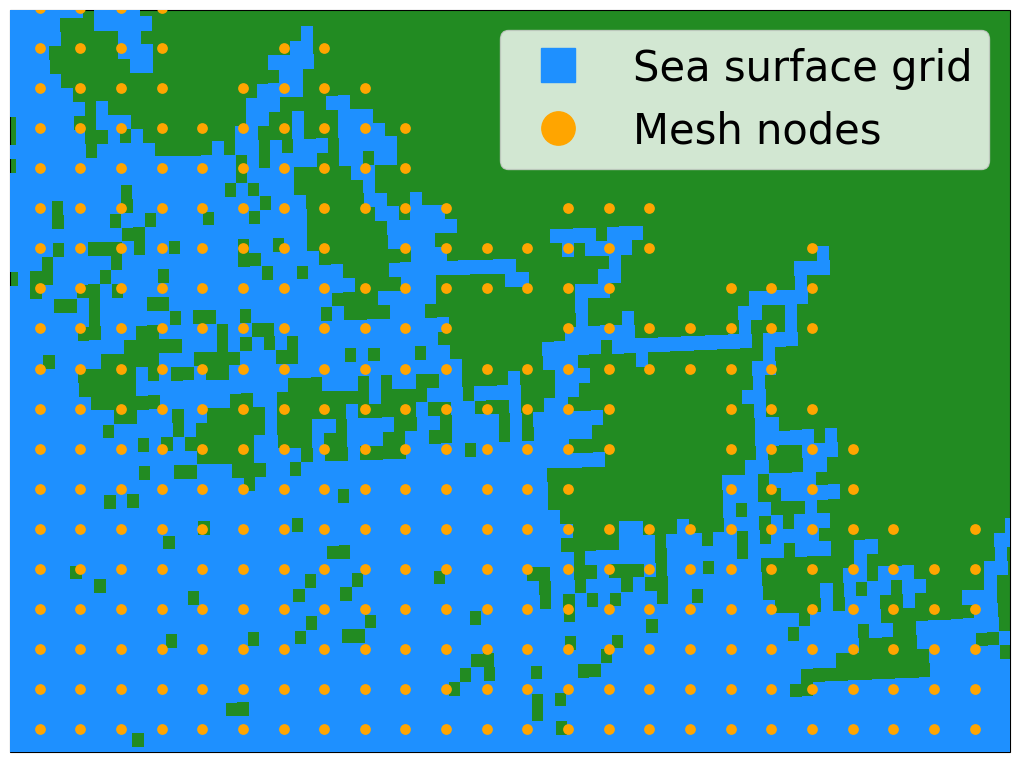}
  \caption{Quadrilateral, level 0}
\end{subfigure}
\begin{subfigure}{.4\textwidth}
  \includegraphics[width=\textwidth]{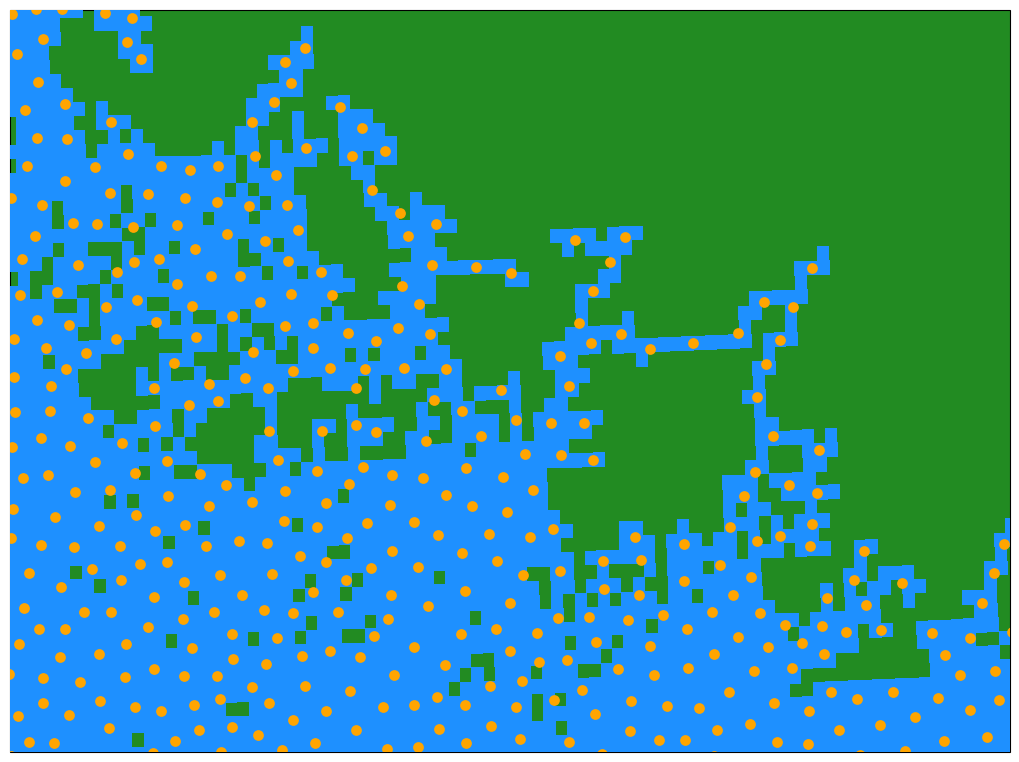}
  \caption{Cluster, level 0}
\end{subfigure}
\begin{subfigure}{.4\textwidth}
  \includegraphics[width=\textwidth]{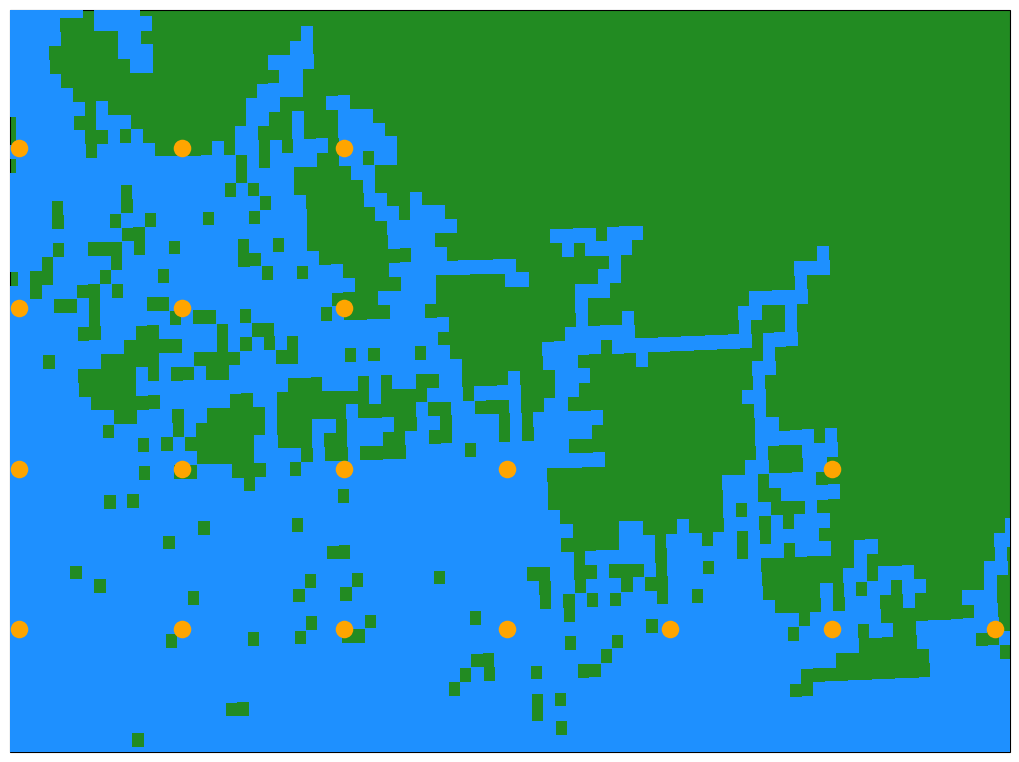}
  \caption{Quadrilateral, level 2}
\end{subfigure}
\begin{subfigure}{.4\textwidth}
  \includegraphics[width=\textwidth]{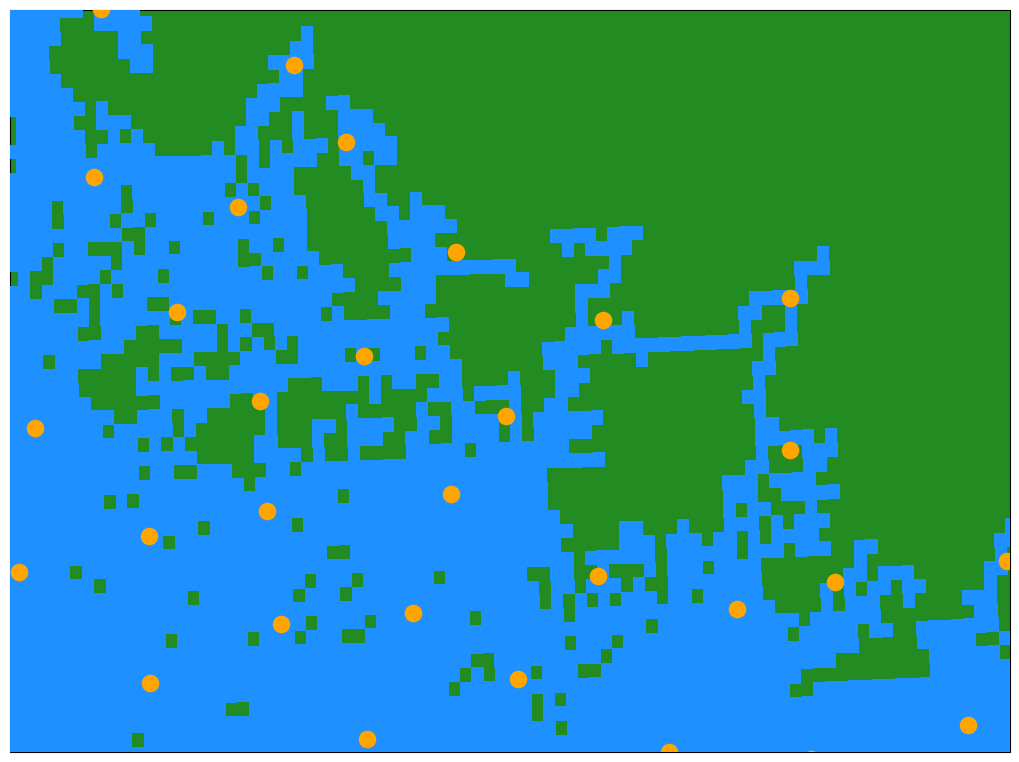}
  \caption{Cluster, level 2}
\end{subfigure}
\caption{
Example of mesh node placement in the Turku Archipelago in south-western Finland.
}
\label{fig:graph_baltic_turku}
\end{figure}

Our improved graph construction building on spatial clustering leads to graphs that better conform to the exact geometry of the sea surface. This is especially noticeable around complex coastlines, such as around islands and straits.
Here we provide additional illustrations of the difference between our cluster-based graphs and existing graph creation  methods.
\Cref{fig:full_global_graphs} shows the full graphs for the global ocean model, and \cref{fig:full_baltic_graphs} shows graphs for the Baltic Sea model.
For Njord-Baltic, all spatial processing and coordinate embeddings are defined in a Lambert Conformal Conic projection centered at (20\textdegree E, 60\textdegree N), which is used also for the graph.
We further show examples of mesh node placement for specific regions in \crefrange{fig:graph_global_california}{fig:graph_global_red_sea} and \crefrange{fig:graph_baltic_braviken}{fig:graph_baltic_turku}, for both nodes at the first mesh level $\mathcal{G}_0$ and the last $\mathcal{G}_2$.
These chosen regions serve as clear examples of how our graph creation leads to different node placements.
Nodes in the cluster-based graphs conform to the coastlines, and are nicely spaced throughout narrow straits.
For the higher graph level $\mathcal{G}_2$, the quadrilateral and icosahedral graphs can completely lack any mesh nodes in key areas, since there is nothing in the graph creation that favors placing nodes over sea rather than over land.

\subsection{Grid--mesh connections}
Grid-to-mesh (G2M) edges connect each grid node to all mesh nodes within a radius of $0.67\, d_m$, where $d_m$ is the mean edge length at the bottom mesh level. Mesh-to-grid (M2G) edges connect each interior grid node to its $k=3$ nearest mesh nodes. Edges crossing land are filtered out: an edge is removed if its midpoint is closer to a land grid node than to a sea grid node, or if the edge exceeds a maximum length threshold. For the regional model (projected coordinates), this threshold is \SI{20}{\kilo\meter}; for the global model (unit-sphere chord distance), it is 0.1, corresponding to approximately $5.7^\circ$ of great-circle arc.

\subsection{Static graph features}
Each edge in the graph carries a small set of static features that are embedded by an MLP before being used in the \gls{GNN} layers. For the regional model, edge features consist of the edge length and the 2D displacement vector $(\Delta x, \Delta y)$ in projected coordinates, yielding 3 features per edge. For the global model, edge features consist of the chord length and the 3D Cartesian displacement $(\Delta x, \Delta y, \Delta z)$ on the unit sphere, yielding 4 features per edge. Mesh node features at each hierarchy level are similarly embedded. For the regional model, these are the projected position $(x, y)$ and the Voronoi cell area (3 features). For the global model, node features are $(\sin\lambda, \cos\lambda, \sin\phi, \cos\phi)$, where $\lambda$ and $\phi$ are the longitude and latitude, plus the spherical Voronoi cell area (5 features). GraphCast~\cite{lam2023learning} uses $\cos\phi$, $\sin\lambda$, and $\cos\lambda$ as mesh node features; we additionally include $\sin\phi$, which is directly proportional to the Coriolis parameter $f = 2\Omega\sin\phi$ governing the influence of Earth's rotation on ocean currents. The Voronoi cell area for each mesh node is computed as the area of the corresponding Voronoi cell: in projected coordinates for the regional model and on the unit sphere (via spherical Voronoi tessellation) for the global model. Coastal nodes whose Voronoi cells extend over land are assigned zero area. All edge features are normalized by the longest mesh edge length, and mesh node features are min--max normalized across hierarchy levels.

\subsection{Graph statistics}

Tables~\ref{tab:global_cluster} through~\ref{tab:regional_quadrilateral} summarize the hierarchical node and edge counts for the various graph architectures evaluated in this work. For the global configuration, we have the K-means cluster mesh~\cref{tab:global_cluster}) and the icosahedral mesh~\ref{tab:global_icosahedral} across both the \SI{1}{\degree} pretraining and \SI{0.25}{\degree} finetuning resolutions. For the Baltic Sea, we use a cluster graph~\cref{tab:regional_cluster} for Njord and a quadrilateral mesh~\cref{tab:regional_quadrilateral} for SeaCast.

\begin{table}[tbh]
\centering
\caption{Number of nodes and edges in the global cluster graph.}
\begin{tabular}{lclrr}
\toprule
\textbf{Dataset} & \textbf{Resolution} & \textbf{Graph} & \textbf{Nodes} & \textbf{Edges} \\
\midrule
& & $\mathcal{G}_0$ & 33777 & 197348 \\
& & $\mathcal{G}_{0,1} / \mathcal{G}_{1,0}$ & - & 33777 \\
\multirow{3}{*}{Rea./Ana.}
& \multirow{3}{*}{\SI{1}{\degree} / \SI{0.25}{\degree}}
& $\mathcal{G}_1$ & 8409 & 48198 \\
& & $\mathcal{G}_{1,2} / \mathcal{G}_{2,1}$ & - & 8409 \\
& & $\mathcal{G}_2$ & 2088 & 11592 \\
\cmidrule(lr){3-5}
& & Total & 44274 & 341510 \\
\midrule
\multirow{3}{*}{Reanalysis}
& \multirow{3}{*}{\SI{1}{\degree}}
& $\mathcal{G}_\text{G2M}$ & - & 73077 \\
& & $\mathcal{G}_\text{M2G}$ & - & 126594 \\
& & Grid & 42348 & - \\
\midrule
\multirow{3}{*}{Reanalysis}
& \multirow{3}{*}{\SI{0.25}{\degree}}
& $\mathcal{G}_\text{G2M}$ & - & 1167675 \\
& & $\mathcal{G}_\text{M2G}$ & - & 2021392 \\
& & Grid & 676736 & - \\
\midrule
\multirow{3}{*}{Analysis}
& \multirow{3}{*}{\SI{0.25}{\degree}}
& $\mathcal{G}_\text{G2M}$ & - & 1165790 \\
& & $\mathcal{G}_\text{M2G}$ & - & 2016228 \\
& & Grid & 675219 & - \\
\bottomrule
\end{tabular}
\label{tab:global_cluster}
\end{table}

\begin{table}[tbh]
\centering
\caption{Number of nodes and edges in the global icosahedral graph.}
\begin{tabular}{lclrr}
\toprule
\textbf{Dataset} & \textbf{Resolution} & \textbf{Graph} & \textbf{Nodes} & \textbf{Edges} \\
\midrule
& & $\mathcal{G}_0$ & 28753 & 166354 \\
& & $\mathcal{G}_{0,1} / \mathcal{G}_{1,0}$ & - & 28753 \\
\multirow{3}{*}{Rea./Ana.} 
& \multirow{3}{*}{\SI{1}{\degree} / \SI{0.25}{\degree}} 
& $\mathcal{G}_1$ & 7194 & 40662 \\
& & $\mathcal{G}_{1,2} / \mathcal{G}_{2,1}$ & - & 7194 \\
& & $\mathcal{G}_2$ & 1817 & 9862 \\
\cmidrule(lr){3-5}
& & Total & 37764 & 288772 \\
\midrule
\multirow{3}{*}{Reanalysis} 
& \multirow{3}{*}{\SI{1}{\degree}} 
& $\mathcal{G}_\text{G2M}$ & - & 67914 \\
& & $\mathcal{G}_\text{M2G}$ & - & 126408 \\
& & Grid & 42348 & - \\
\midrule
\multirow{3}{*}{Reanalysis} 
& \multirow{3}{*}{\SI{0.25}{\degree}} 
& $\mathcal{G}_\text{G2M}$ & - & 1088358 \\
& & $\mathcal{G}_\text{M2G}$ & - & 2016626 \\
& & Grid & 676736 & - \\
\midrule
\multirow{3}{*}{Analysis} 
& \multirow{3}{*}{\SI{0.25}{\degree}} 
& $\mathcal{G}_\text{G2M}$ & - & 1086477 \\
& & $\mathcal{G}_\text{M2G}$ & - & 2011528 \\
& & Grid & 675219 & - \\
\bottomrule
\end{tabular}
\label{tab:global_icosahedral}
\end{table}

\begin{table}[tbh]
\centering
\footnotesize 
\setlength{\tabcolsep}{3pt} 

\begin{minipage}{0.37\textwidth}
\centering
\caption{Number of nodes and edges in the Baltic Sea cluster graph.}
\begin{tabular}{lrr}
\toprule
\textbf{Graph} & \textbf{Nodes} & \textbf{Edges} \\
\midrule
$\mathcal{G}_0$ & 16358 & 92332 \\
$\mathcal{G}_{0,1} / \mathcal{G}_{1,0}$ & - & 16358 \\
$\mathcal{G}_1$ & 1767 & 9404 \\
$\mathcal{G}_{1,2} / \mathcal{G}_{2,1}$ & - & 1767 \\
$\mathcal{G}_2$ & 193 & 920 \\
\midrule
Total & 18318 & 138906 \\
\midrule
$\mathcal{G}_\text{G2M}$ & - & 301460 \\
$\mathcal{G}_\text{M2G}$ & - & 436104 \\
Grid & 147701 & - \\
\bottomrule
\end{tabular}
\label{tab:regional_cluster}
\end{minipage}
\hspace{0.08\textwidth}
\begin{minipage}{0.37\textwidth}
\centering
\caption{Number of nodes and edges in the Baltic Sea quadrilateral graph.}
\begin{tabular}{lrr}
\toprule
\textbf{Graph} & \textbf{Nodes} & \textbf{Edges} \\
\midrule
$\mathcal{G}_0$ & 16274 & 123264 \\
$\mathcal{G}_{0,1} / \mathcal{G}_{1,0}$ & - & 16274 \\
$\mathcal{G}_1$ & 1804 & 12787 \\
$\mathcal{G}_{1,2} / \mathcal{G}_{2,1}$ & - & 1804 \\
$\mathcal{G}_2$ & 203 & 1220 \\
\midrule
Total & 18281 & 173427 \\
\midrule
$\mathcal{G}_\text{G2M}$ & - & 302661 \\
$\mathcal{G}_\text{M2G}$ & - & 429540 \\
Grid & 147701 & - \\
\bottomrule
\end{tabular}
\label{tab:regional_quadrilateral}
\end{minipage}
\end{table}
\clearpage

\section{Training Details}
\label{app:training-details}
\label{sec:training_details}

\subsection{Loss functions}
The base loss function accounts for the ocean's bathymetric structure, following SeaCast~\cite{holmberg2025accurate}:
\begin{equation}
    \mathcal{L}_{\mathrm{base}} = \frac{1}{T} \sum_{t=1}^{T} \sum_{i=1}^{C} \lambda_i \sum_{l=1}^{L_i} \frac{w_l}{N_l} \sum_{v=1}^{N_l} a_v\, \ell\left(\hat{X}^{t}_{v,i},\, X^{t}_{v,i},\, \hat{\sigma}^t_{v,i}\right)
    \label{eq:masked-loss}
\end{equation}
where $T$ is the number of autoregressive rollout steps, $C$ is the number of variables, $L_i$ is the number of depth levels for variable $i$, $N_l$ is the number of ocean grid nodes at depth level $l$, $a_v$ is the latitude--longitude area weight for grid cell~$v$ (normalized to unit mean), $w_l$ is a depth-level weight that we configure as $1/L_i$ for depth-resolved variables and 0.5 for all surface-level variables, and lastly $\lambda_i$ is the inverse variance of one-step time differences for variable $i$. The per-entry loss~$\ell$ is the \gls{ELBO} and \gls{CRPS} for Njord.

The \gls{CRPS} term in Eq.~\eqref{eq:total-loss} employs the almost-fair \gls{CRPS} (af\gls{CRPS}) estimator introduced by~\cite{lang2026aifs}, which interpolates between the biased and unbiased (fair) \gls{CRPS} estimators~\cite{ferro2014fair}.
The loss can be written as:
\begin{equation}
    \mathcal{L}_{\mathrm{\gls{CRPS}}} = \frac{1}{M}\sum_{m=1}^{M} |\hat{x}_m - x| - \frac{1-\varepsilon}{2M(M-1)}\sum_{m=1}^{M}\sum_{m'=1}^{M} |\hat{x}_m - \hat{x}_{m'}|, \qquad \varepsilon = \frac{1-\alpha}{M}
    \label{eq:crps}
\end{equation}
where $\hat{x}_m$ is one scalar dimension from an ensemble member $\hat{X}^t_m$ sampled from $\hat{p}_\theta$, $x$ is the target, and $\alpha \in (0, 1]$ controls the interpolation ($\alpha = 1$ recovers the fair \gls{CRPS}). We use $\alpha = 0.95$ and $M = 2$ ensemble members during training. The \gls{CRPS} loss is normalized per variable by dividing by the per-variable standard deviation and reduced using the same spatial mask and area weighting as Eq.~\eqref{eq:masked-loss}.

\subsection{Training curriculum}

\paragraph{Global schedule.}

The global Njord model follows a two-stage resolution training schedule, transitioning from coarse to fine resolution as detailed in Table~\ref{tab:global_training_schedule}. The model is first pretrained on \SI{1}{\degree} resolution data for 325 epochs using cosine learning rate annealing. Subsequently, it is finetuned on \SI{0.25}{\degree} resolution data for an additional 165 epochs, utilizing a 5-epoch linear warmup followed by cosine decay. Because Njord's grid-to-mesh encoder's uses Flexible Propagation Networks (Eq.~\eqref{eq:flex-pn-aggr}) with mean aggregation no rescaling of incoming messages used for similar models with sum aggregation~\cite{price2025probabilistic} is needed when moving to higher resolution.

Finetuning with the af\gls{CRPS} loss requires sampling two more trajectories, which makes it more expensive to train. Fortunately you can get away with doing this for only a few epochs. For the global case which has a much larger grid than the regional model, we especially pay attention to this by choosing a high $\lambda_\mathrm{\gls{CRPS}}=10^6$ for 5 epochs only. The model is further fine-tuned on analysis data as seen in \cref{tab:global_training_schedule}. Note when looking at epochs and GPUh for analysis fine-tuning that it consists of less training data (1 year) compared to reanalysis (28 years).

\begin{table}[ht]
\centering
\caption{Training schedule and hyperparameter configuration for the global Njord model. Pretraining (\SI{1}{\degree}) follows a cosine annealing schedule from $10^{-3}$ to $10^{-5}$, while finetuning (\SI{0.25}{\degree}) incorporates a 5-epoch linear warmup from $10^{-5}$ to $10^{-4}$ followed by cosine decay from $10^{-4}$ to $10^{-5}$. Finetuning on the analysis dataset uses a constant $10^{-5}$ learning rate.}
\label{tab:global_training_schedule}
\begin{tabular}{ccccccc}
\toprule
Dataset & Resolution & Epochs & $\lambda_{\mathrm{KL}}$ & $\lambda_{\mathrm{\gls{CRPS}}}$ & Unrolling $T$ & GPUh \\
\midrule
Reanalysis & \SI{1}{\degree}    & 100 & 0   & 0        & 1 & 600 \\
Reanalysis & \SI{1}{\degree}    & 200 & 0.1 & 0        & 1 & 1300 \\
Reanalysis & \SI{1}{\degree}    & 25  & 0.1 & 0        & 2 & 320 \\
\midrule
Reanalysis & \SI{0.25}{\degree} & 5   & 0   & 0        & 1 & 110 \\
Reanalysis & \SI{0.25}{\degree} & 150 & 0.1 & 0        & 1 & 3300 \\
Reanalysis & \SI{0.25}{\degree} & 5   & 0.1 & 0        & 2 & 210 \\
Reanalysis & \SI{0.25}{\degree} & 5   & 0.1 & $10^{6}$ & 2 & 530 \\
\midrule
Analysis & \SI{0.25}{\degree}   & 100  & 0.1 & $10^{6}$ & 2 & 590 \\
Analysis & \SI{0.25}{\degree}   & 100  & 0.1 & $10^{7}$ & 7 & 1970 \\
\bottomrule
\end{tabular}
\end{table}

\paragraph{Regional schedule.}
The Njord-Baltic model is trained for 350 epochs using a staged curriculum and cosine learning rate annealing. As shown in Table~\ref{tab:baltic_training_schedule}, the process begins with 100 epochs of pure autoencoder training ($T=1, \lambda_{\mathrm{KL}}=0$) to establish the base representation. We then introduce the KL divergence term ($\lambda_{\mathrm{KL}}=0.1$) for 200 epochs to align the prior with the approximate posterior. The final 50 epochs focus on temporal consistency and calibration: first by unrolling the model to two steps ($T=2$) for 25 epochs, and then by incorporating the \gls{CRPS} loss ($\lambda_{\mathrm{\gls{CRPS}}} = 10^4$) for the remaining 25 epochs to optimize the ensemble spread. Lastly, Njord-Baltic is finetuned for 50 epochs on 1 year of analysis data.

\begin{table}[ht]
\centering
\caption{Training schedule and hyperparameter configuration for the Njord-Baltic model, using a cosine learning rate annealing schedule from $10^{-3}$ to $10^{-5}$ over 350 epochs. Finetuning on the analysis dataset uses a constant $10^{-5}$ learning rate.}
\label{tab:baltic_training_schedule}
\begin{tabular}{cccccc}
\toprule
Dataset & Epochs & $\lambda_{\mathrm{KL}}$ & $\lambda_{\mathrm{\gls{CRPS}}}$ & Unrolling $T$ & GPUh \\
\midrule
Reanalysis & 100 & 0   & 0        & 1 & 400 \\
Reanalysis & 200 & 0.1 & 0        & 1 & 870 \\
Reanalysis & 25  & 0.1 & 0        & 2 & 200 \\
Reanalysis & 25  & 0.1 & $10^{4}$ & 2 & 520 \\
\midrule
Analysis & 50 & 0.1 & $10^{4}$ & 2 & 40 \\
\bottomrule
\end{tabular}
\end{table}

\paragraph{Deterministic baseline.} We train SeaCast as a baseline using the 3-level icosahedral mesh in the global setting, and quadrilateral mesh in the regional setting. Note that SeaCast is a regional model but we generalize to the globe. SeaCast is configured with 3 processing layers and latent dimension 256~\cite{holmberg2024regional}, and trained for 175 epochs with the weighted MSE loss and cosine learning rate annealing where 150 epochs are single-step training, followed by 25 epochs with two-step autoregressive rollout to improve temporal stability. The learning rate scheduler is 5 epoch linear warmup from $10^{-5}$ to $10^{-4}$ followed by cosine decay from $10^{-4}$ to $10^{-5}$. In the global setting, SeaCast is further trained on \SI{0.25}{\degree} data for 60 epochs, where 50 are 1-step prediction, and the last 10 are 2-step prediction with 5 epoch linear warmup from $10^{-5}$ to $10^{-4}$ followed by cosine decay from $10^{-4}$ to $10^{-5}$. Lastly it is finetuned on analysis data for 75 epochs, where 50 epochs are 2-step training and the last 25 epochs use 7 steps, with a constant $10^{-5}$ learning rate. In the regional setting, the SeaCast model is finetuned on Baltic Sea analysis data for 25 epochs with 2-step prediction and a constant $10^{-5}$ learning rate.

\paragraph{Implementation details.} All models are optimized with AdamW~\cite{loshchilov2019decoupled} with $\beta_1=0.9$, $\beta_2=0.95$ and a weight decay of $0.1$. Gradient checkpointing~\cite{chen2016training} is employed at each autoregressive step to fit the model into GPU memory for global training at \SI{0.25}{\degree} resolution. The training uses bfloat 16 precision and evaluation is performed with float 32 precision. All models are trained using 64 AMD MI250X GPUs, except global \SI{0.25}{\degree} reanalysis training which ran on 128 AMD MI250X GPUs. The GPUs each have 64GB VRAM.

\clearpage

\section{Additional results}
\label{sec:additional_results}

\subsection{Input and forcing steps}
\label{sec:inputs_ablation}

\begin{figure}[tbh]
    \centering
    \includegraphics[width=\textwidth]{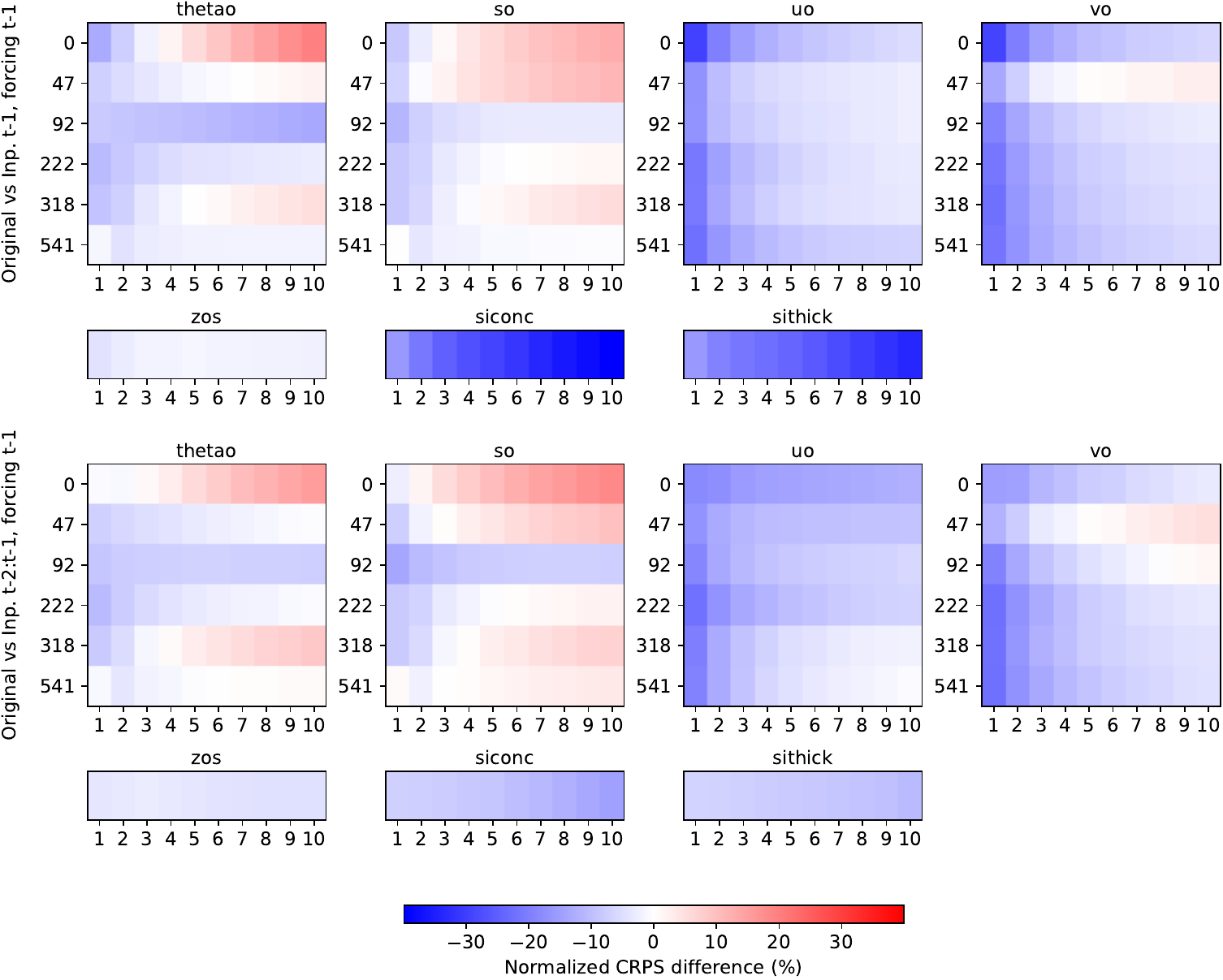}
    \caption{Ensemble mean \gls{CRPS} scorecards. The heatmaps display the relative difference between Njord with 2 input steps and 3 forcing steps versus using one two input steps and 1 forcing step and 1 input step and 1 forcing step across all ocean variables. Blue indicates better performance by the original approach with 2 input steps and 3 forcing steps.}
    \label{fig:inp_crps}
\end{figure}

We study the effect of the temporal context provided to Njord by comparing the default configuration, which uses two input states $(t{-}2, t{-}1)$ and three forcing steps $(t{-}2, t{-}1, t)$, against two ablations: (i) the same two input states with a single forcing step at $t{-}1$, and (ii) a single input state and forcing step at $t{-}1$. This ablation compares models trained using the \SI{1}{\degree} pretraining schedule. Performance is reported per variable, depth and lead time as the normalized \gls{RMSE} difference between the target configuration $A$ and the baseline $B$ as $\Delta_{\text{CRPS}} \;=\; \frac{\text{CRPS}_{A}-\text{CRPS}_{B}}{\text{CRPS}_{B}}$. Negative (blue) values therefore indicate that the default configuration outperforms the ablation, while positive (red) values indicate degradation. \Cref{fig:inp_crps} show that the default configuration with two input steps and three forcing steps is broadly preferred. Sea ice predictions specifically seems to benefit  most from having two input steps.

\clearpage

\subsection{Graph type}
\label{sec:graph_type_ablation}

\begin{figure}[tbh]
    \centering
    \includegraphics[width=\textwidth]{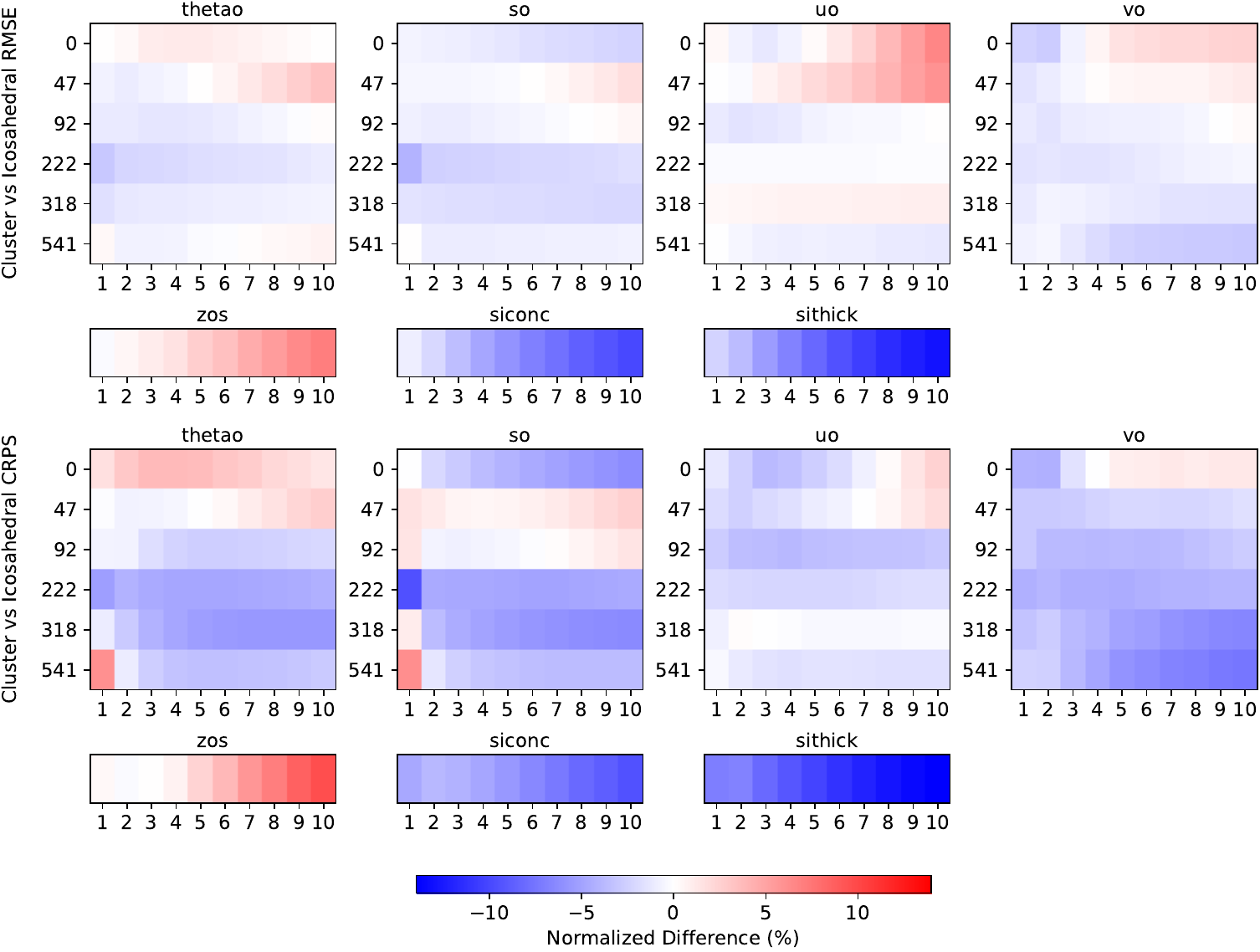}
    \caption{The heatmaps display the relative difference in \gls{RMSE} and \gls{CRPS} between Njord trained with the global cluster graph and the icosahedral graph across all ocean variables.}
    \label{fig:scorecard_cluster_vs_icosahedral}
\end{figure}

The global cluster graph generally achieves lower \gls{RMSE} and \gls{CRPS} compared to the icosahedral graph as seen in \cref{fig:scorecard_cluster_vs_icosahedral}. In this experiment two Njord models with the cluster and icosahedral graphs in~\crefrange{tab:global_cluster}{tab:global_icosahedral} are trained on \SI{1}{\degree} reanalysis data according to the pretraining schedule in~\cref{tab:global_training_schedule}. The slight performance advantage from the cluster graph is partly attributed to the higher spatial density of the cluster graph, which utilizes 33,777 mesh nodes, whereas the three-layer (6, 5, 4 split) icosahedral graph contains 28,753 nodes. As demonstrated in \crefrange{fig:graph_global_california}{fig:graph_global_red_sea}, the K-means clustering approach also ensures that mesh nodes are evenly distributed over sea areas, maintaining coverage even within narrow bays and complex coastal geometries. The comparison could be made even stronger by training on higher resolution data with both meshes, but it becomes unnecessarily computationally expensive to do so for this ablation.

The cluster graph has the added benefit that it allows for a continuous selection of node counts, where one can choose the mesh resolution to what fits in GPU memory. In contrast, icosahedral graphs are constrained by discrete subdivision levels, resulting in significant jumps in resolution between splits. For example, after masking land areas, an increase from 6 to 7 splits results in a roughly fourfold increase in nodes, from 28,753 to 115,016, that then did not fit in memory anymore with a high hidden dimension.

\clearpage

\subsection{Sea ice treatment}
\label{sec:sea_ice_comparison}

We evaluate three strategies for enforcing sea ice constraints:

\paragraph{Unconstrained.}
The unconstrained model predicts state increments $\Delta X^t$ directly, with the next state obtained as $\hat{X}^{t+1} = X^t + \Delta X^t$. No bounds are enforced during training or inference. We observe that this leads to unphysical sea ice values already at short lead times, with concentrations outside $[0, 1]$ and negative thicknesses appearing at moderate rollout lengths.

\paragraph{Output clamping.}
To constrain the sea ice variables to realistic bounds we apply smooth invertible activation functions in the residual update itself. For variables with both a lower and upper bound (here \texttt{siconc} with bounds $[0, 1]$), we use a rescaled sigmoid:
\begin{equation}
    f_{\mathrm{sig}}(x) = x^L + (x^U - x^L)\,\sigma(x), \qquad f_{\mathrm{sig}}^{-1}(y) = \sigma^{-1}\!\!\left(\frac{y - x^L}{x^U - x^L}\right)
\end{equation}
For variables with only a lower bound (here \texttt{sithick} with $x^L = 0$), we use a shifted softplus: $f_{\mathrm{sp}}(x) = x^L + \mathrm{softplus}(x)$, with corresponding inverse. The clamped next state is then computed as
\begin{equation}
    \hat{X}^{t+1}_v = f\!\left(f^{-1}(X^t_v) + \Delta X^t_v\right)
    \label{eq:clamped-update}
\end{equation}
This formulation operates in the unconstrained latent space of $f^{-1}$, adds the predicted increment there, and maps back through $f$. Both $f$ and $f^{-1}$ are smooth and differentiable everywhere.

\paragraph{Density channel.}
Purely using a soft clamping may lead to accumulation of small deviations from zero ice over time, so in addition to the clamping we adopt a density channel mechanism~\cite{gordon2019convolutional} used previously for handling missing wave data in the Aurora foundation model~\cite{bodnar2025foundation}. In Aurora, each ocean wave variable receives its own density channel indicating whether a measurement is present (1) or absent (0), allowing the model to represent the absence of wave data. In our case a single binary density channel $d \in \{0, 1\}$ is constructed from \gls{SIC}: $d = \mathbb{I}[\text{SIC} > 0]$. This channel is appended to the model state, and predicted alongside all other variables. During autoregressive rollout, the predicted density logit $\hat{d}_{\mathrm{raw}}$ is passed through a sigmoid and thresholded at~$0.5$. Where the predicted density falls below this threshold (indicating no ice), the density channel, \gls{SIC}, and \gls{SIT} are all set to their normalized-zero values in the feedback state passed to the next step. Where density exceeds the threshold, the density channel is set to its normalized-one value. This ensures that the model receives a clean, zero-ice input at ice-free locations rather than a small residual value that can accumulate. The raw predictions before applying the threshold are used for loss computation.

\begin{figure}[tbh]
    \centering
    \includegraphics[width=\textwidth]{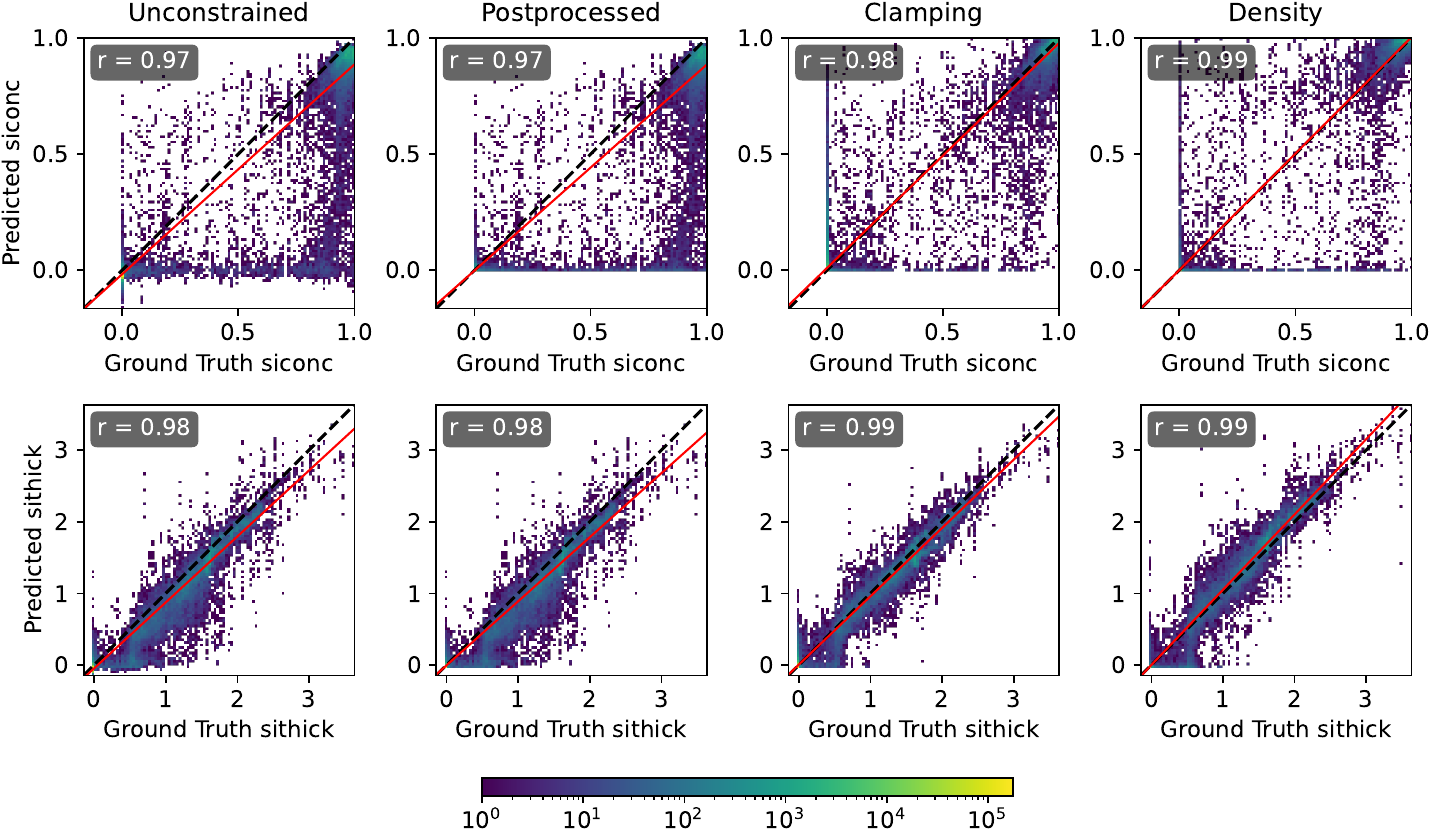}
    \caption{Log-scaled scatter density heatmaps evaluating predicted versus observed \gls{SIC} and \gls{SIT} at a 30-day lead time. Each panel includes a 1:1 reference line (dashed), a linear line of best fit (solid red), and the Pearson correlation coefficient ($r$).}
    \label{fig:ice_heatmaps}
\end{figure}

\begin{figure}[tbh]
    \centering
    \includegraphics[width=\textwidth]{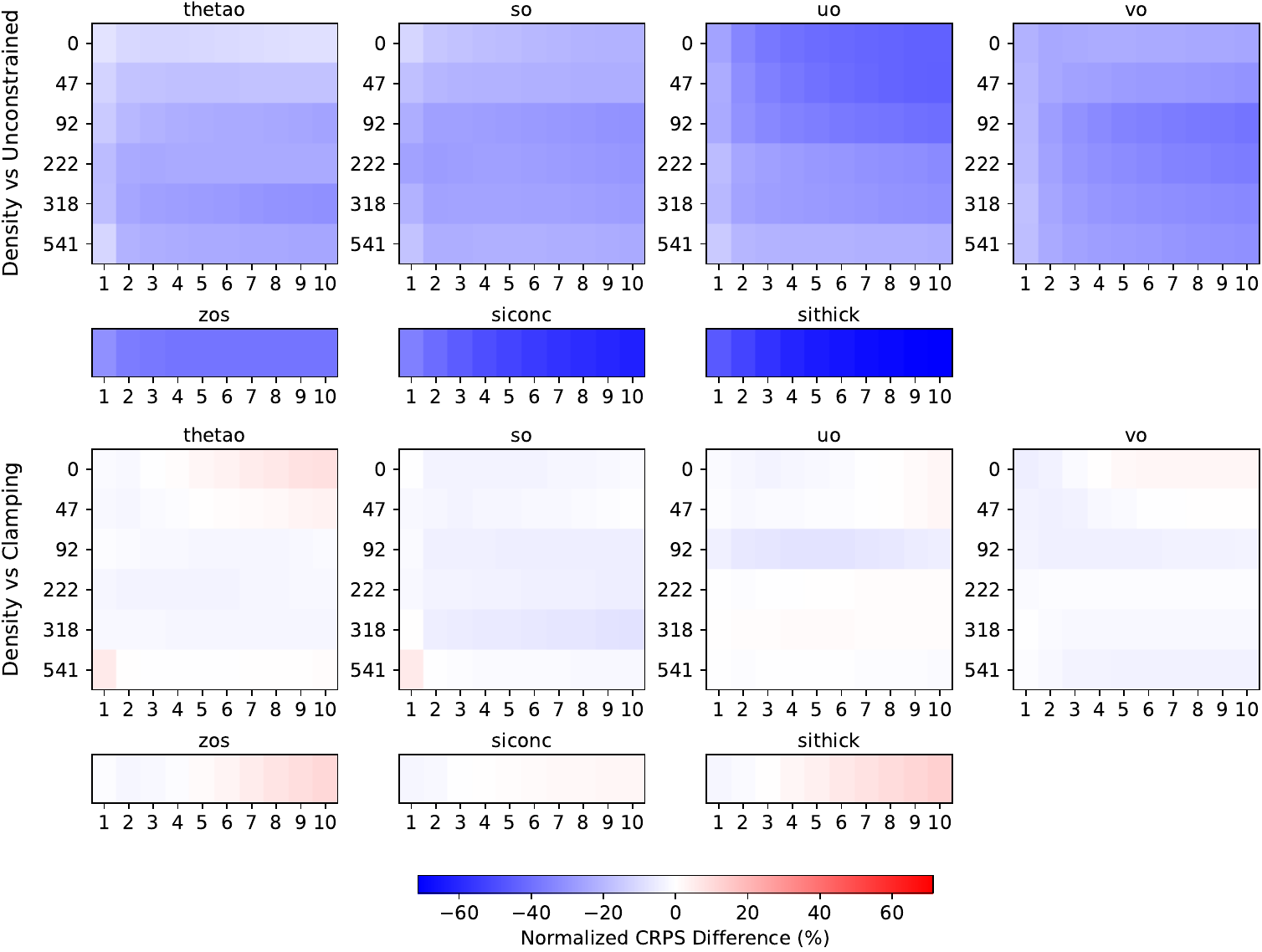}
    \caption{Ensemble mean \gls{CRPS} scorecards. The heatmaps display the relative difference between the density + clamping model versus the unconstrained and clamping only approaches across all ocean variables. Blue indicates better performance by the density + clamping approach.}
    \label{fig:scorecard_crps}
\end{figure}

The unconstrained variation misses the structure of sea ice near the Antarctic, compared to clamping or density + clamping, denoted as just density. Postprocessing to sea ice bounds looks fairly good, but the correlation remains lower than when clamping is used in \cref{fig:ice_heatmaps}. Purely clamping leads to an unfavorable buildup of \gls{SIC} near the equator, present in all ensemble members at high lead times. Clamping + density leads to better \gls{CRPS} compared to the unconstrained approach as seen in \cref{fig:scorecard_crps}, by a fair amount which is a bit surprising, considering it should affect mainly sea ice. Purely clamping leads to a very comparable \gls{CRPS} to when the density channel is used on top.

\begin{figure}[tbh]
    \centering
    \includegraphics[width=\textwidth]{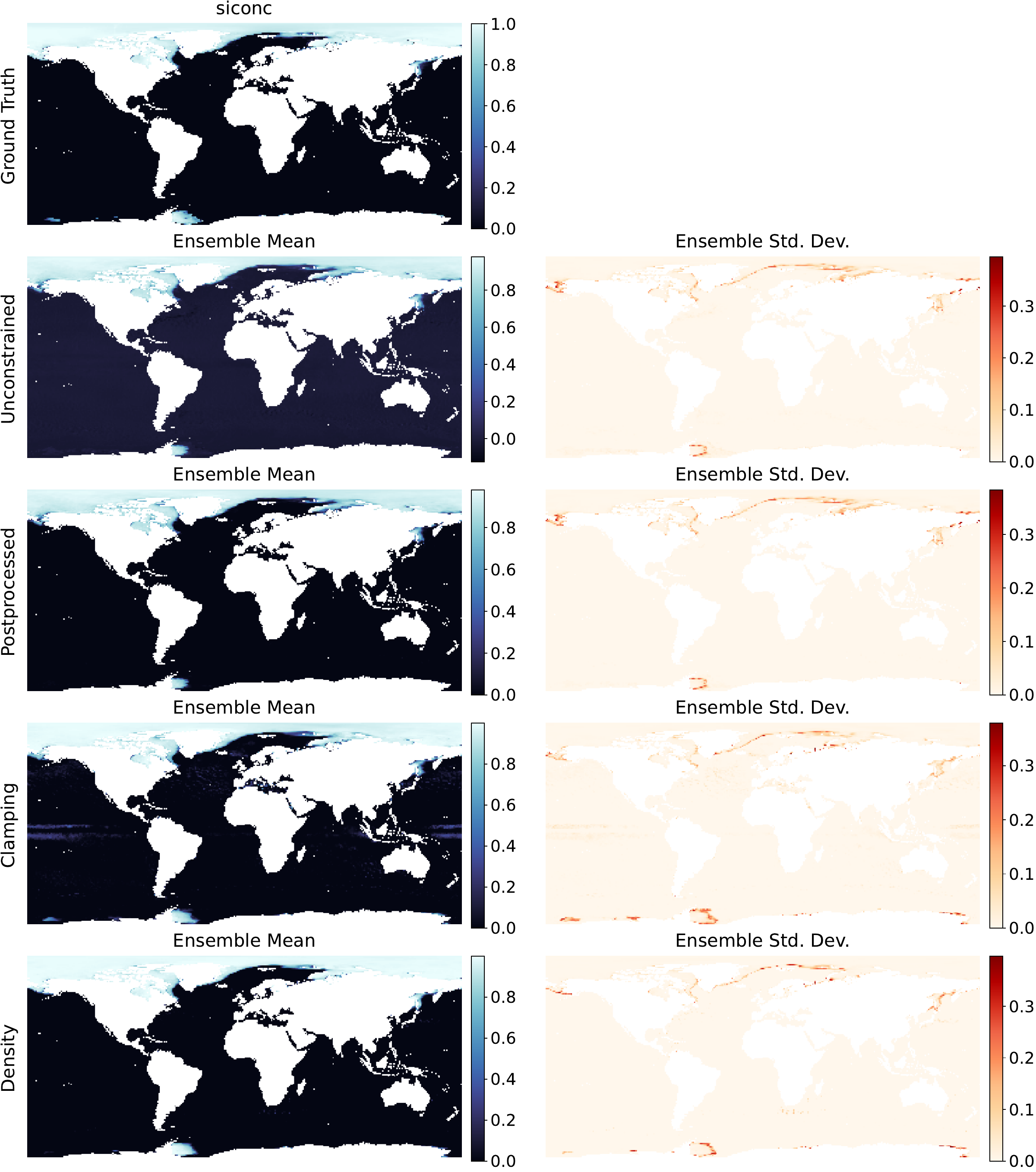}
    \caption{Spatial evaluation of \gls{SIC} at a 30-day lead time. The panels compare the ground truth against the ensemble mean and ensemble standard deviation for different ways of handling ice boundaries.}
    \label{fig:map_siconc}
\end{figure}

\clearpage

\subsection{OceanBench evaluation}
\label{sec:oceanbench_eval}

To assess the global forecasting capabilities of Njord, we evaluate it using the OceanBench framework~\cite{el2025oceanbench}. OceanBench provides a standardized benchmark for data-driven ocean forecasting by comparing models against the operational physics-based system, GLO12~\cite{lellouche2023evolution}, as well as state-of-the-art deep learning baselines including GLONET~\cite{aouni2025glonet}, WenHai~\cite{cui2025forecasting}, and XiHe~\cite{wang2024xihe}. In OceanBench the native resolution of each model is used when comparing to the analysis or reanalysis targets.

\paragraph{Evaluation on the GLORYS12 reanalysis track.} 
\cref{tab:glorys12_track} presents the evaluation against the independent GLORYS12 reanalysis~\cite{jean2021copernicus} track\footnote{
We use the updated evaluation procedure and values from the live OceanBench webpage (\url{https://oceanbench.lab.dive.edito.eu}) as of April 2026, or version 0.0.3 of the associated Python package. These have been updated after the publication of the original paper \citep{el2025oceanbench}. 
}. In this setting, Njord demonstrates competitive performance compared to the physics-based GLO12 system. Njord shows particular strength forecasting zonal and meridional currents, where it consistently outperforms GLO12 (indicated by the blue cells), as well as temperature down to \SI{50}{m}. Njord is skillful at predicting geostrophic currents at the surface. Geostrophic currents provide a diagnostic of large-scale ocean circulation and transport. Under the geostrophic approximation, these currents are derived directly from forecasted SSH. Because Njord maintains highly accurate and stable predictions of SSH, this fidelity translates directly into superior geostrophic current forecasts.

\begin{table}[tbph]
    \centering
    \caption{Scorecard for the GLORYS12 reanalysis track. Colors represent the normalized \gls{RMSE} difference with respect to the GLO12 operational baseline. Blue cells indicate that the model outperforms GLO12, while red indicates higher error.}
    \renewcommand{\arraystretch}{1.2}
    \setlength{\tabcolsep}{3pt}
    \input{tables/glorys12_track}
    \label{tab:glorys12_track}
\end{table}

\paragraph{Evaluation on the GLO12 analysis track.}
\cref{tab:glo12_track} presents the evaluation on the GLO12 analysis track. Because the GLO12 forecast model and the GLO12 analysis share the exact same underlying physical parameterizations, it achieves the lowest \gls{RMSE} when evaluated against its own analysis fields that are produced through a weekly data assimilation cycle applied to GLO12 forecasts. Especially if the observations are sparse, which they generally are in the global ocean, it can be difficult to outperform the physical simulator on this benchmark. On the other hand, the machine learning models are biased to perform better than GLO12 on the reanalysis benchmark. A more independent view of performance is shown when comparing to observations in \cref{tab:observation_track} and \cref{fig:sst_rmse,fig:sst_spatial_rmse}, where the machine learning models generally show favorable results compared to GLO12.

\begin{table}[tbph]
    \centering
    \caption{Scorecard for the GLO12 analysis track. Colors represent the normalized \gls{RMSE} difference with respect to the GLO12 operational baseline. Blue cells indicate that the model outperforms GLO12, while red indicates higher error.}
    \renewcommand{\arraystretch}{1.2}
    \setlength{\tabcolsep}{3pt}
    \input{tables/glo12_track}
    \label{tab:glo12_track}
\end{table}

\paragraph{Evaluation on the observation track.}
\cref{tab:observation_track} presents the evaluation of the models against the \textit{in-situ} observations curated within the IV-TT CLASS-4 framework. At the surface and near-surface layers, Njord demonstrates strong predictive skill, consistently outperforming the operational GLO12 baseline when evaluated against surface drifting buoys measurements of \gls{SST}, shallow Argo (global array of autonomous ocean profiling floats) measurements of 0--5m temperature and salinity, and drifters measurements of 15m currents. However, Njord's performance degrades relative to the baseline at intermediate depth layers, which is particularly evident when compared against deeper Argo profiles for 100--300m temperature and salinity. This decrease in skill at depth is an expected limitation, as Njord currently models fewer vertical depth levels than the other models. Consequently, this lower vertical resolution provides a less granular representation of the 3D ocean state, which becomes apparent when validating against observational data spread across the water column.

\begin{table}[tbph]
    \centering
    \caption{Scorecard for the observation track within the IV-TT CLASS-4 framework. Colors represent the normalized \gls{RMSE} difference with respect to the GLO12 operational baseline. Blue indicate that the model outperforms GLO12 against observations, while red indicates a higher error.}
    \input{tables/observation_track}
    \label{tab:observation_track}
\end{table}

\clearpage

\subsection{Evaluation against SST observations}

To evaluate the models' \gls{SST} forecasts, we used the potential temperature of the uppermost ocean layer, benchmarking these predictions against global ocean adjusted \gls{SST}~\cite{cmems2026sst}.
\begin{wrapfigure}{r}{0.42\textwidth}
    \centering
    \includegraphics[width=\linewidth]{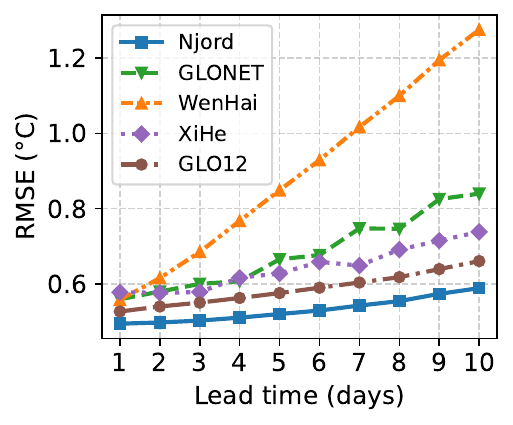}
    \caption{Global \gls{RMSE} of \gls{SST} by forecast lead time, where Njord has the lowest error compared to satellite measurements.}
    \label{fig:sst_rmse}
\end{wrapfigure}
The dataset merges multi-sensor satellite observations into a Level-3 global grid. \cref{fig:sst_rmse} shows the globally averaged \gls{RMSE} for \gls{SST} over a 10-day forecast horizon where the models are interpolated to the \SI{0.1}{\degree} \gls{SST} grid. Njord maintains the lowest \gls{RMSE} across all lead times. The other machine learning models show higher error growth, and WenHai especially so, which is interesting considering it has atmospheric forcing that e.g. GLONET lacks. By day 10, Njord's \gls{RMSE} is below 0.6\textdegree C, remaining lower than the other machine learning models and the GLO12 baseline.

\begin{figure}[htbp]
    \centering
    \includegraphics[width=\textwidth]{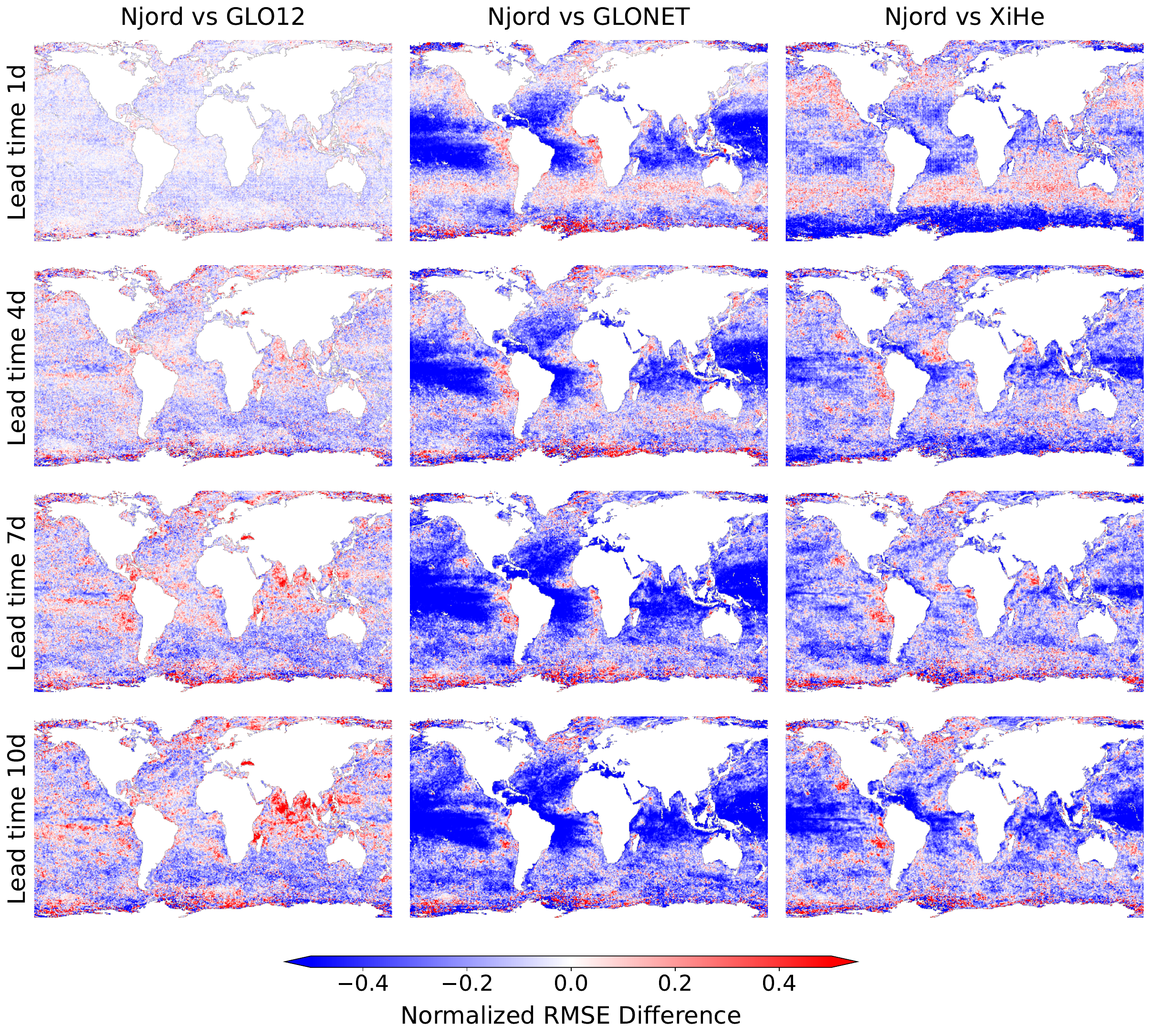}
    \caption{Spatial distribution of normalized \gls{RMSE} difference for \gls{SST} between Njord ensemble mean and three baselines. Blue indicates lower error for Njord.}
    \label{fig:sst_spatial_rmse}
\end{figure}

\cref{fig:sst_spatial_rmse} maps the normalized \gls{RMSE} difference for lead times of 1, 4, 7, and 10 days. Compared to GLONET and XiHe, Njord shows lower error across most ocean basins. Compared to GLO12, Njord performs slightly better across the global ocean. Note that GLO12 and XiHe operate at 3 times higher resolution.

\clearpage

\subsection{Global metrics}

We evaluate Njord on a global scale against the operational physics-based GLO12 model~\cite{lellouche2023evolution}, and deep learning baselines including GLONET~\cite{aouni2025glonet}, our global variant of SeaCast~\cite{holmberg2025accurate}, and a Persistence forecast. The ground truth used for verification is the global analysis. We use this as reference, because the initial conditions are from the same product, and the only time a high \gls{SSR} in training translated to equally high \gls{SSR} during evaluation was when we initialized and evaluated on the same source of data. The \gls{RMSE} results are expected to be similar to what is shown in \cref{tab:glo12_track}, but here all models are evaluated on the same \SI{0.25}{\degree} resolution. To do this GLO12 is downsampled from its native \SI{0.083}{\degree} resolution using bilinear interpolation. GLONET is like Njord a \SI{0.25}{\degree} resolution model. Note that this model is expected to behave differently because it is trained only on reanalysis. SeaCast, on the other hand, provides a deterministic baseline trained in a similar manner as Njord.

Ensemble metrics for Njord, specifically the \gls{SSR}, the \gls{CRPS}, and the \gls{RMSE} of the ensemble mean, are compared against the deterministic baselines. For the deterministic models (GLO12, GLONET, WenHai, XiHe, SeaCast, and Persistence), \gls{MAE} is shown alongside Njord's \gls{CRPS} as a comparable deterministic reference. Mathematically, the \gls{CRPS} evaluates the distance between the predictive \gls{CDF} and the empirical \gls{CDF} of the observation. In the deterministic limit, where the predictive distribution is a Dirac delta function (a point mass) at the predicted value, the \gls{CRPS} reduces exactly to the \gls{MAE}.
We use the unbiased \gls{CRPS} estimator, corresponding to  $\alpha = 1$ in \cref{eq:crps}.
The \gls{SSR} computation includes the finite ensemble size correction from \citet{ssr_correction}. To account for varying grid cell areas, the metrics are weighted by the cosine of the latitude, normalized to unit mean.

\begin{figure}[tbh]
    \centering
    \includegraphics[width=.89\textwidth]{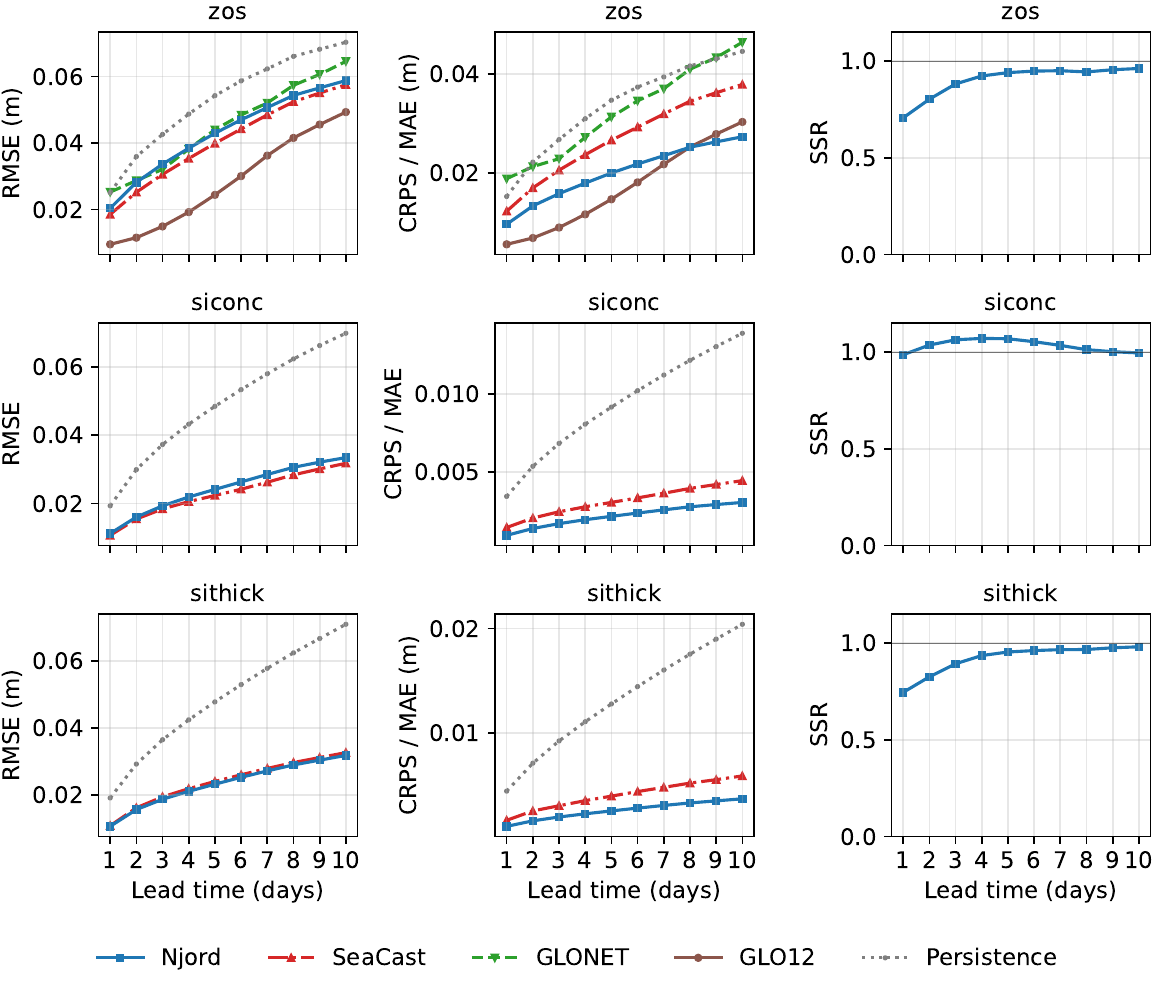}
    \caption{Surface variables: \gls{SSH}, \gls{SIC}, and \gls{SIT}. Columns from left to right show \gls{RMSE}, \gls{CRPS}, and \gls{SSR}.}
    \label{fig:global_metrics_surface}
\end{figure}

\begin{figure}[tbh]
    \centering
    \includegraphics[height=0.95\textheight, keepaspectratio]{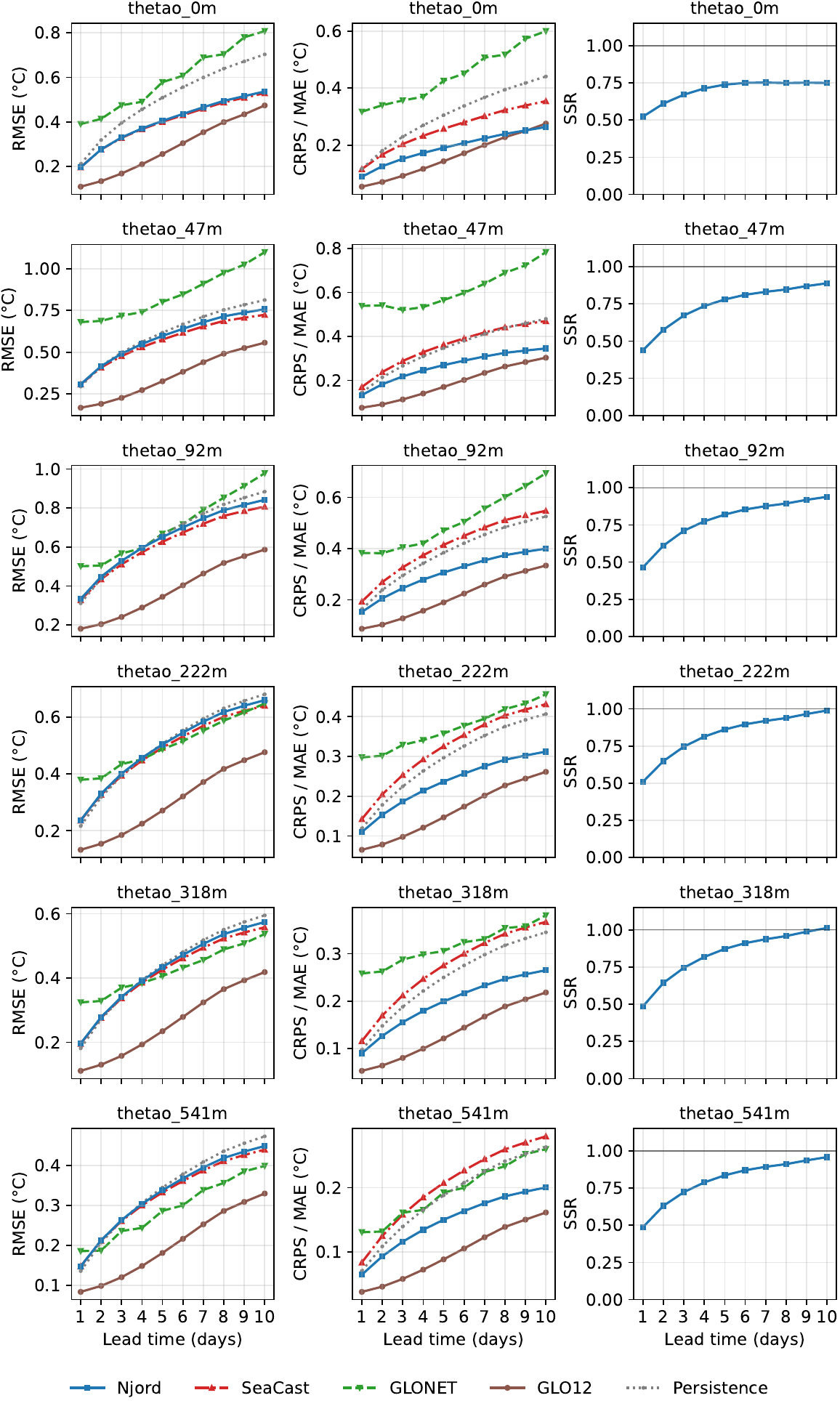}
    \caption{Temperature at six different depths. Columns from left to right show \gls{RMSE}, \gls{CRPS}, and \gls{SSR}.}
    \label{fig:global_metrics_thetao}
\end{figure}

\begin{figure}[tbh]
    \centering
    \includegraphics[height=0.96\textheight, keepaspectratio]{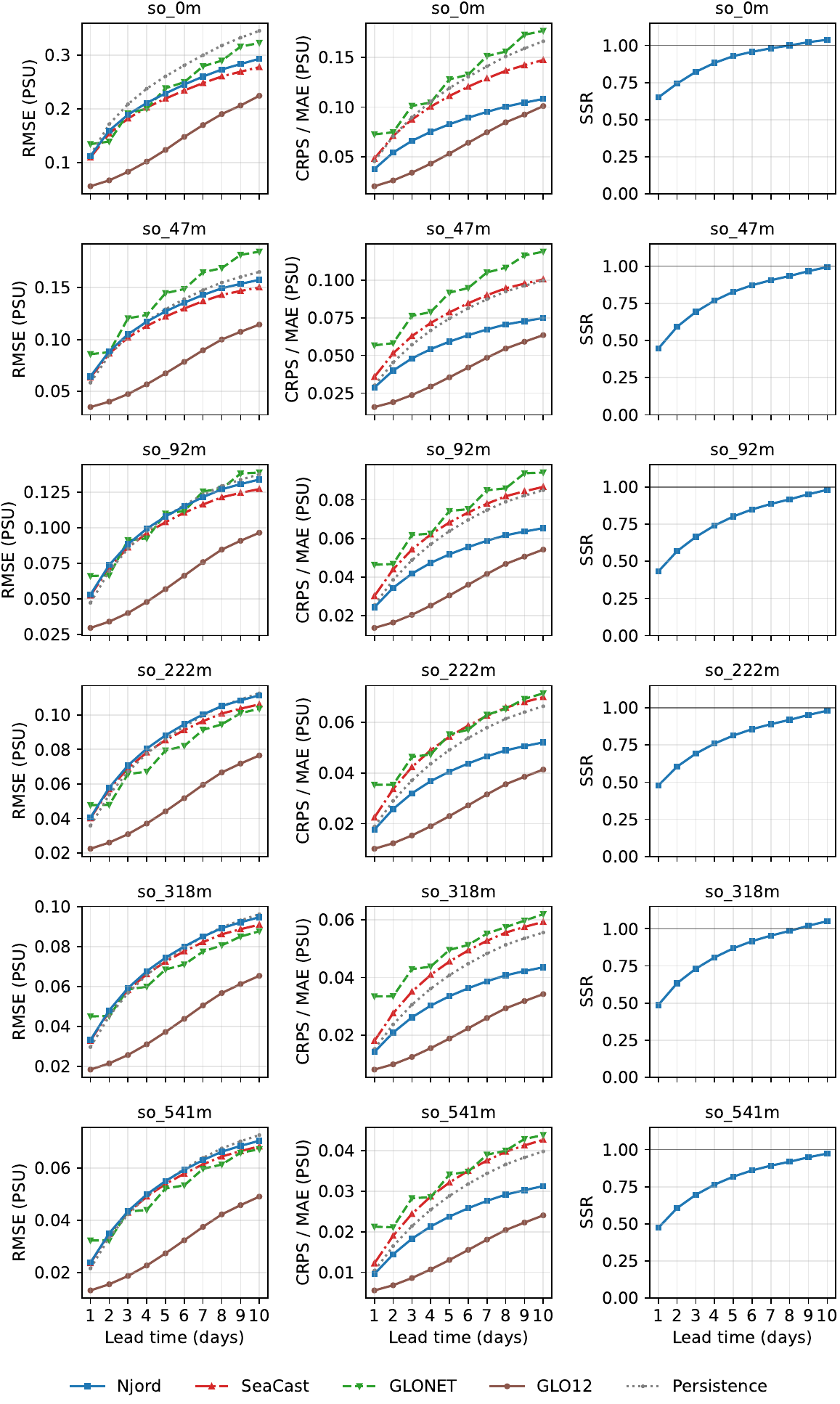}
    \caption{Salinity at six different depths. Columns from left to right show \gls{RMSE}, \gls{CRPS}, and \gls{SSR}.}
    \label{fig:global_metrics_so}
\end{figure}

\begin{figure}[tbh]
    \centering
    \includegraphics[height=0.95\textheight, keepaspectratio]{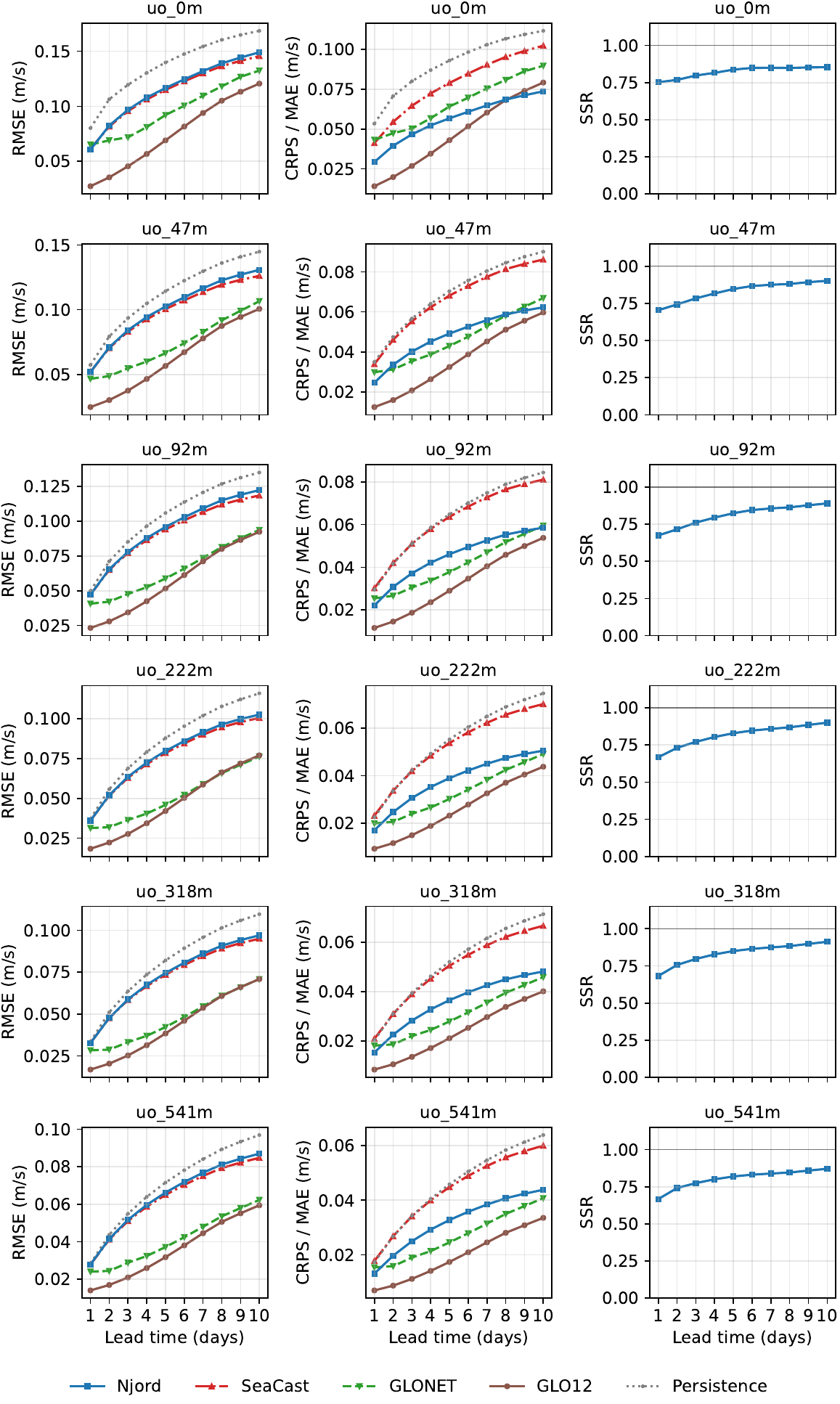}
    \caption{Zonal current at six different depths. Columns from left to right show \gls{RMSE}, \gls{CRPS}, and \gls{SSR}.}
    \label{fig:global_metrics_uo}
\end{figure}

\Crefrange{fig:global_metrics_surface}{fig:global_metrics_uo} report these metrics per variable and lead time. The figure panels are organized into three columns: \gls{SSR} on the left, \gls{CRPS} (or \gls{MAE}) in the middle, and RMSE on the right. The \gls{SSR} is defined as the ratio between the standard deviation of the ensemble and the \gls{RMSE} of the ensemble mean; values close to one indicate a well-calibrated ensemble. Because zonal and meridional currents exhibit very similar error accumulation patterns, only the zonal components are shown here.

\clearpage

\subsection{Ensemble size comparison}

\begin{figure}[tbh]
    \centering
    \includegraphics[width=\textwidth]{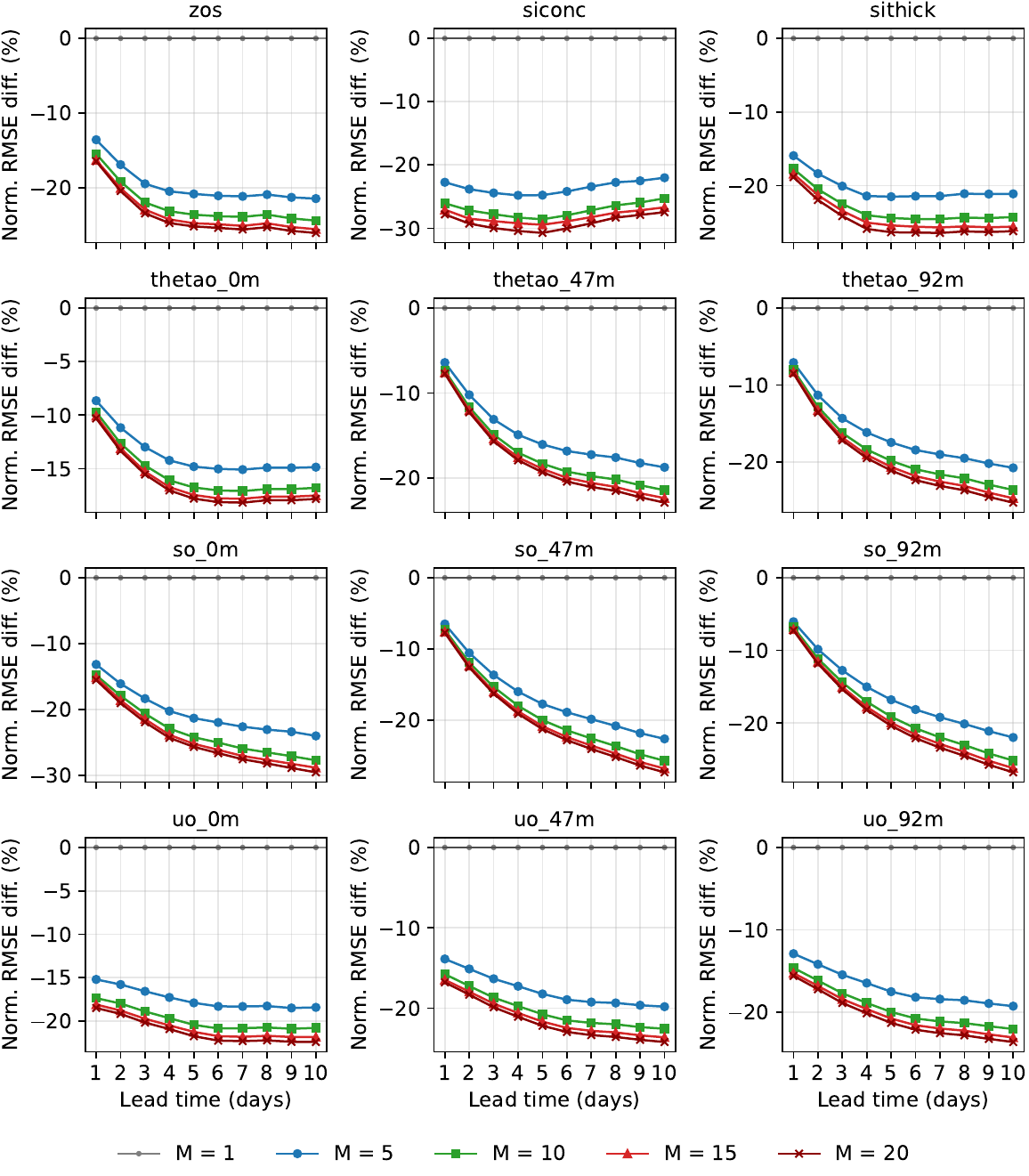}
    \caption{Normalized \gls{RMSE} difference for various variables and depth levels, comparing ensemble sizes $M \in \{5, 10, 15, 20\}$ against a baseline ensemble size of $M=1$. Values below zero indicate a reduction in \gls{RMSE} relative to the baseline.}
    \label{fig:ensemble_size_rmse_ratio}
\end{figure}

\Cref{fig:ensemble_size_rmse_ratio} demonstrates the impact of varying ensemble size ($M$) on Njord's predictive performance, measured as the relative \gls{RMSE} difference compared to a baseline of $M = 1$. Increasing the ensemble size to $M = 20$ yields a systematic \gls{RMSE} reduction across all evaluated variables and depth levels. While larger ensembles produce more accurate deterministic mean forecasts by better sampling predictive uncertainty, these gains must be weighed against the linear increase in computational cost.

\clearpage

\subsection{Global forecasts}
\label{sec:global_example_forecasts}

\begin{figure}[tbh]
    \centering
    \includegraphics[width=\textwidth]{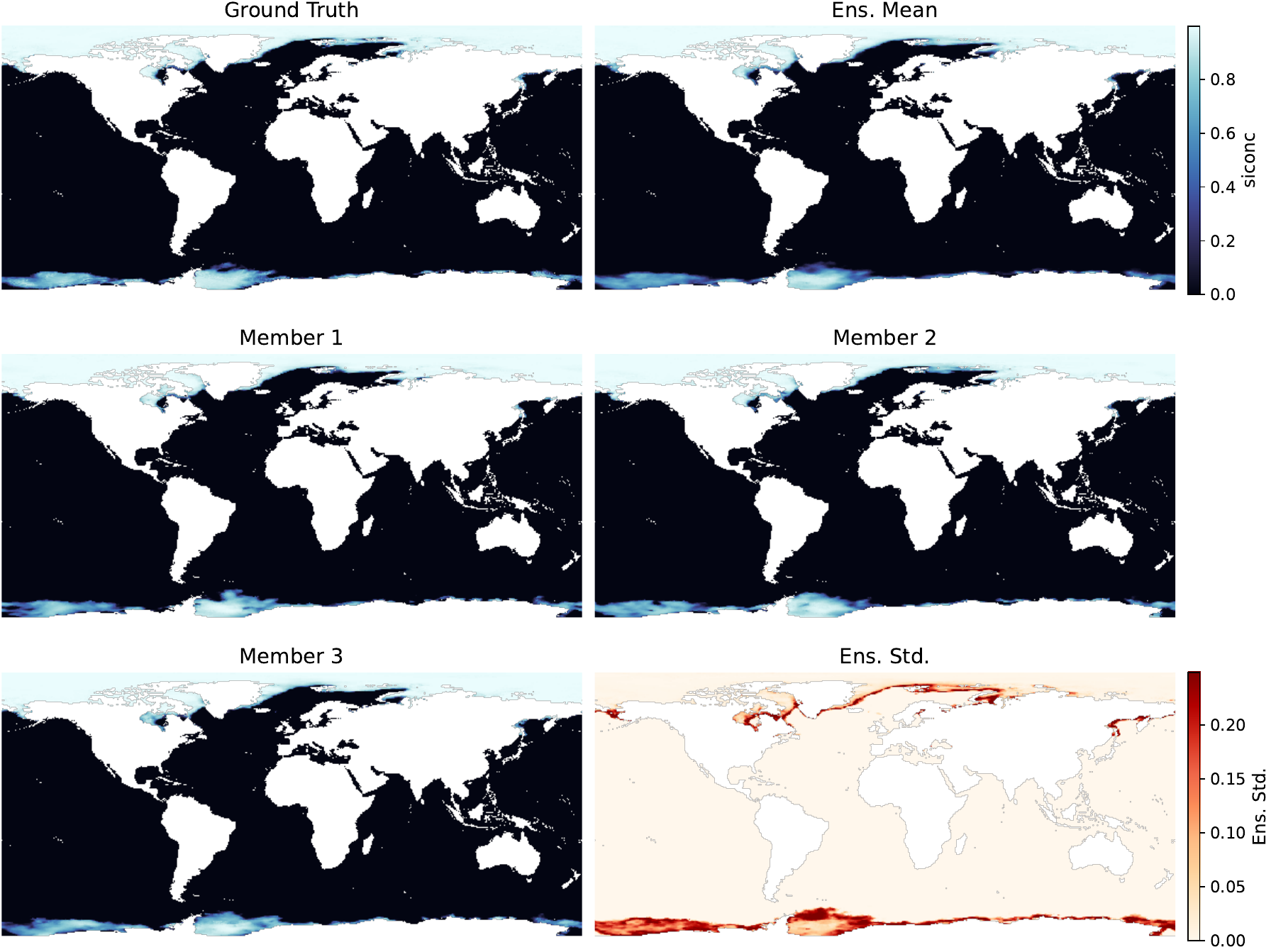}
    \caption{Sea ice concentration at lead time 10\,d, init 2024-12-24.}
    \label{fig:global_example_siconc}
\end{figure}

\begin{figure}[tbh]
    \centering
    \includegraphics[width=\textwidth]{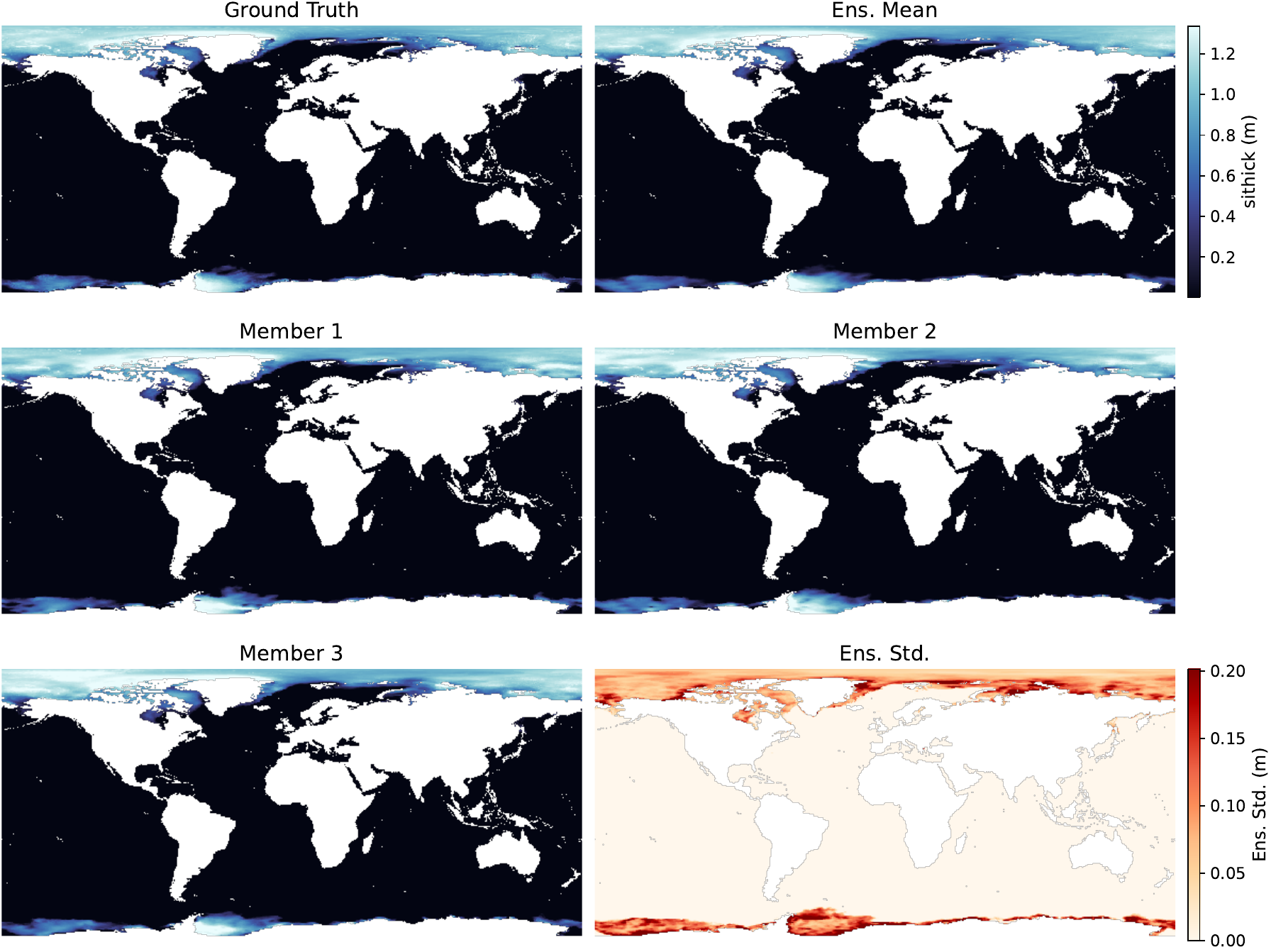}
    \caption{Sea ice thickness at lead time 10\,d, init 2024-12-24.}
    \label{fig:global_example_sithick}
\end{figure}

\begin{figure}[p]
    \centering
    \includegraphics[width=\textwidth]{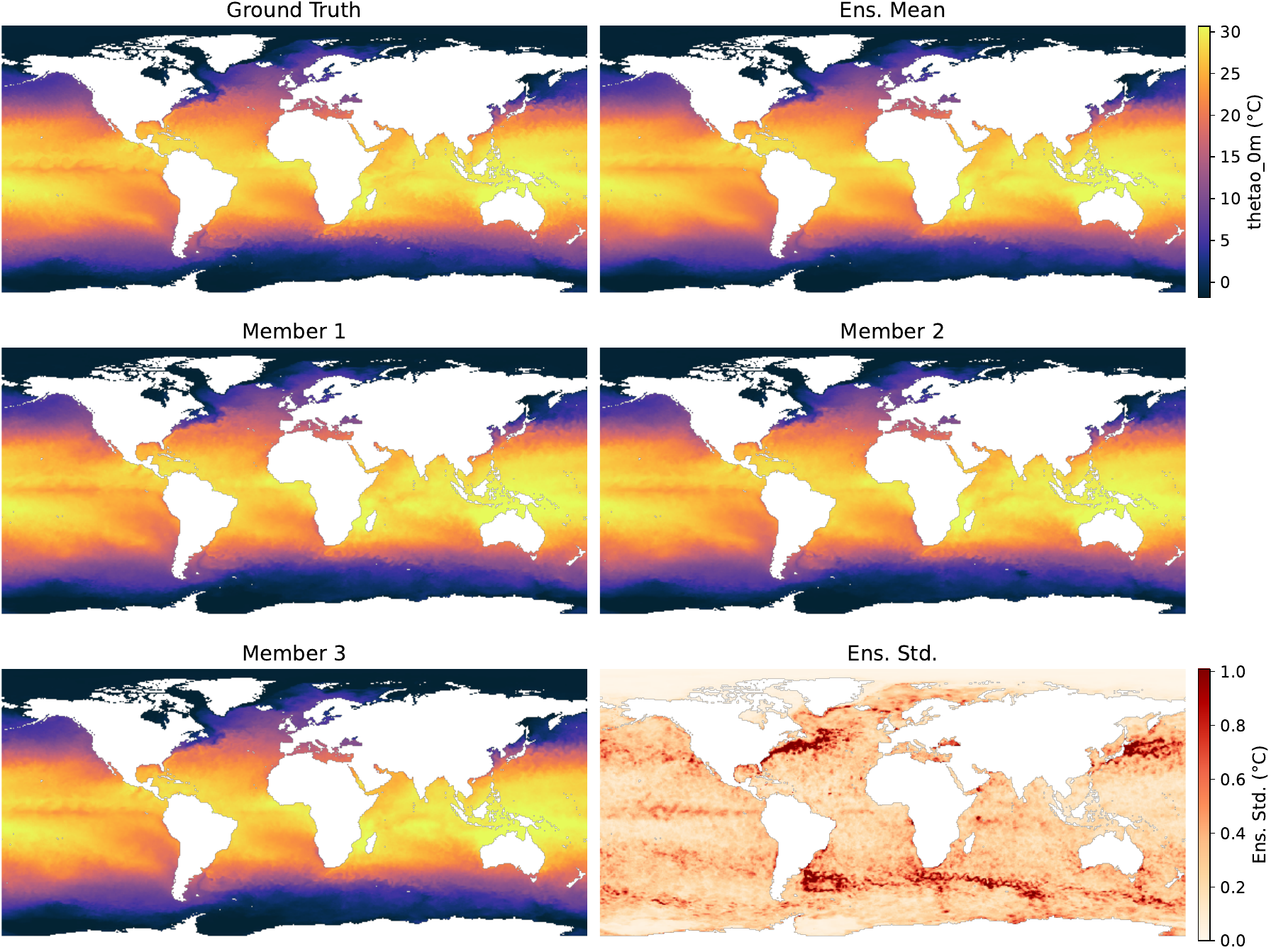}
    \caption{Temperature at the surface, lead time 10\,d, init 2024-12-24.}
    \label{fig:global_example_thetao_0m}
\end{figure}

\begin{figure}[p]
    \centering
    \includegraphics[width=\textwidth]{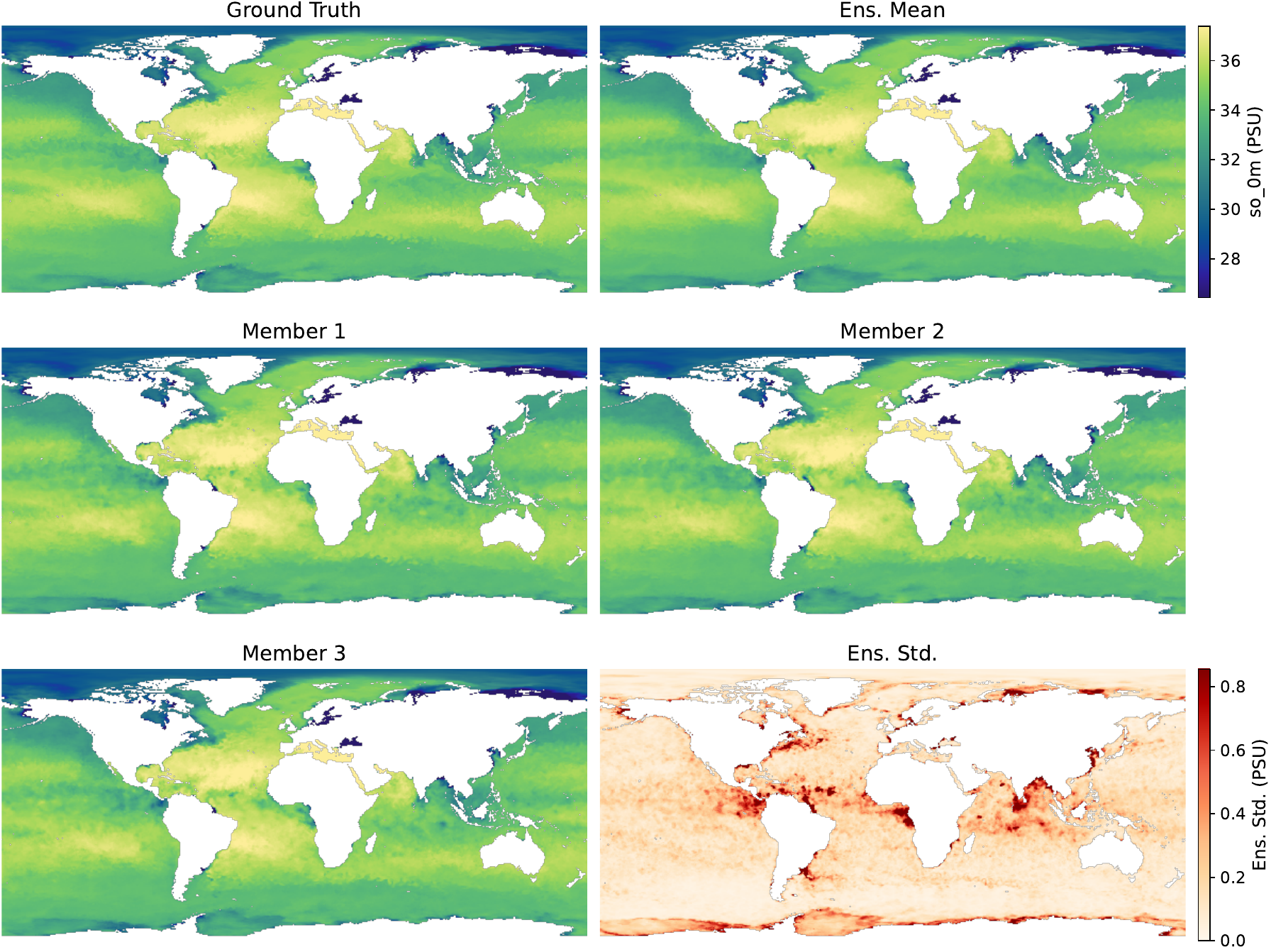}
    \caption{Salinity at the surface, lead time 10\,d, init 2024-12-24.}
    \label{fig:global_example_so_0m}
\end{figure}

\begin{figure}[p]
    \centering
    \includegraphics[width=\textwidth]{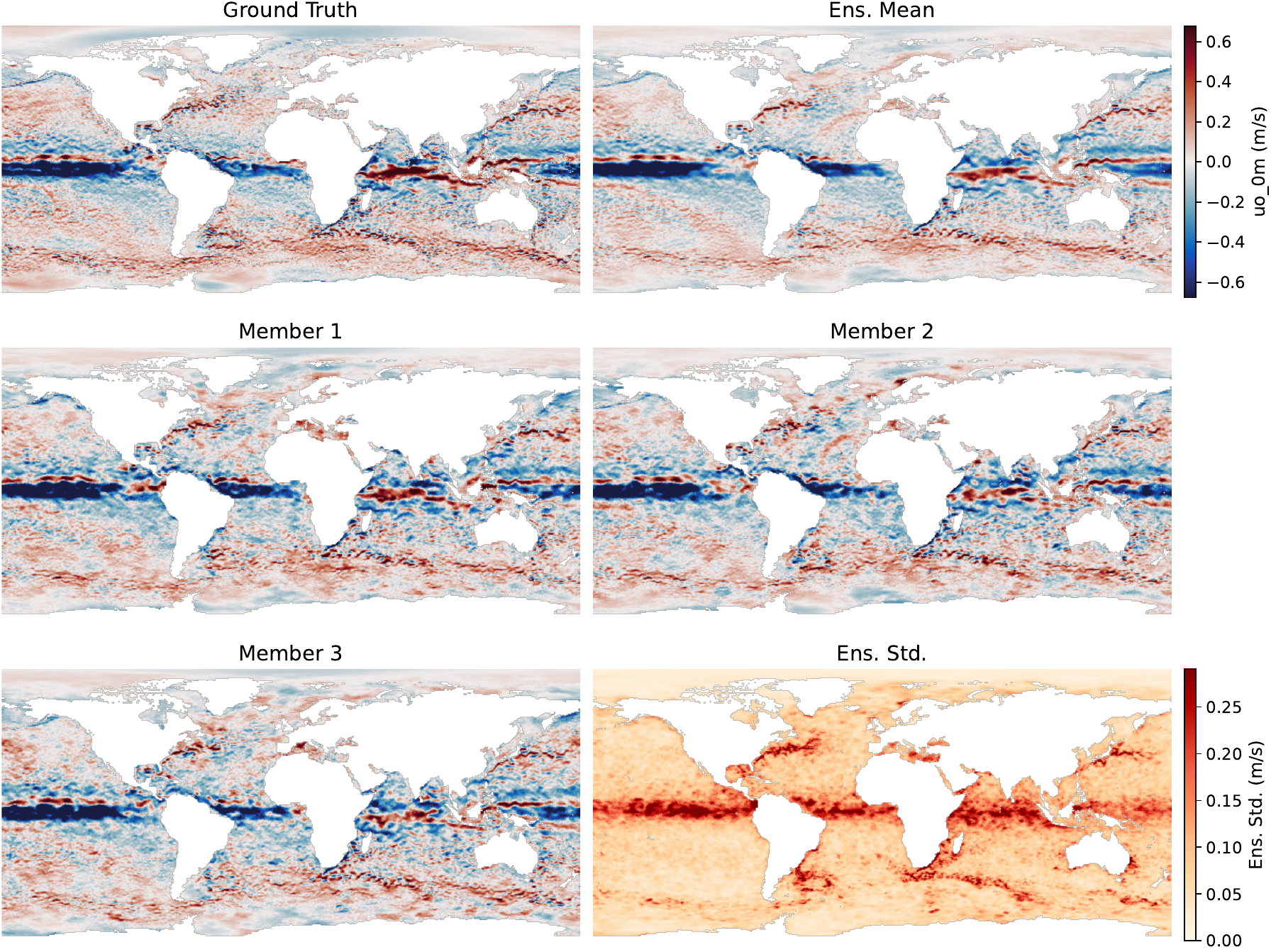}
    \caption{Zonal current at the surface, lead time 10\,d, init 2024-12-24.}
    \label{fig:global_example_uo_0m}
\end{figure}

\begin{figure}[p]
    \centering
    \includegraphics[width=\textwidth]{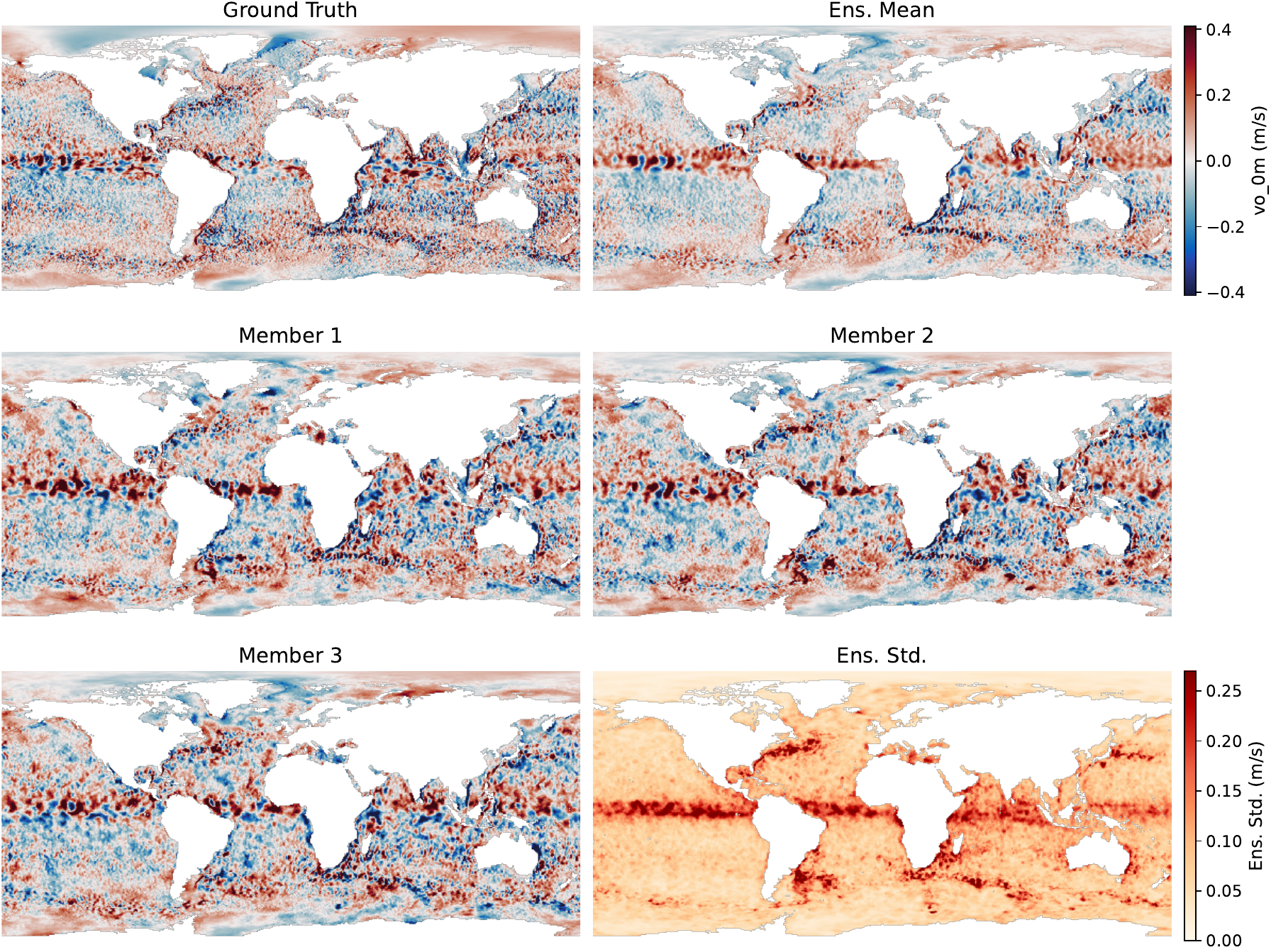}
    \caption{Meridional current at the surface, lead time 10\,d, init 2024-12-24.}
    \label{fig:global_example_vo_0m}
\end{figure}

\begin{figure}[p]
    \centering
    \includegraphics[width=\textwidth]{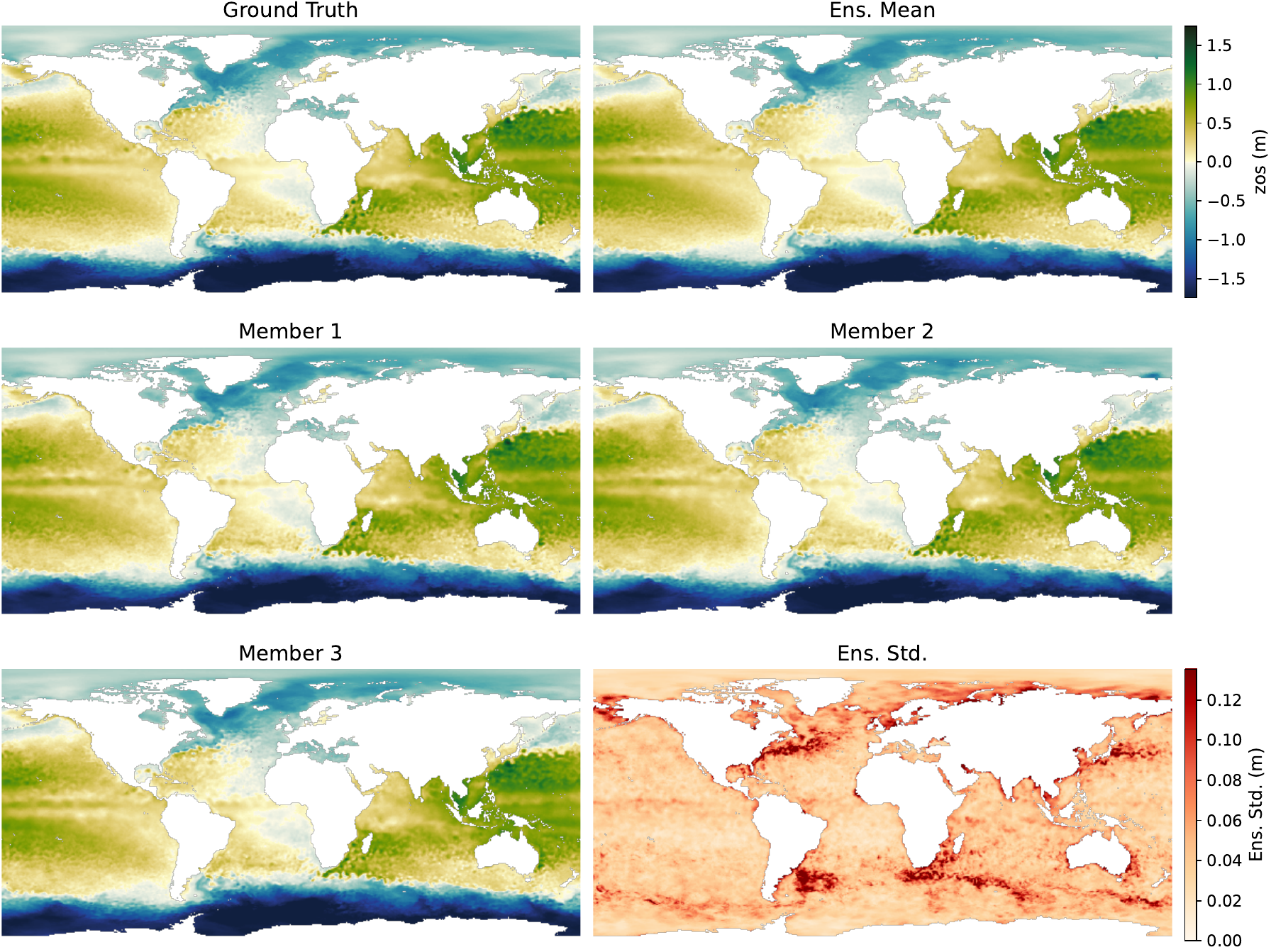}
    \caption{Sea surface height at lead time 10\,d, init 2024-12-24.}
    \label{fig:global_example_zos}
\end{figure}

To illustrate the qualitative behavior of Njord on a global scale, we present example ensemble forecasts initialized on 24 December 2024 at a lead time of 10 days in \crefrange{fig:global_example_siconc}{fig:global_example_zos}. For each variable, we show the analysis target, the ensemble mean, the ensemble standard deviation, and three individual ensemble members.

The individual ensemble members appear sharp and exhibit noticeable variability, whereas the ensemble mean is smoother due to averaging. Sea-ice fields display well-defined edges and are exactly zero in ice-free regions, reflecting the use of clamping and a dedicated density channel. The ensemble standard deviation is elevated near the ice edge, where its position varies between members, and for \gls{SIT}, it is also elevated within regions where ice is present. 

For potential temperature and \gls{SSH}, the ensemble spread is most pronounced in dynamically active regions characterized by sharp thermal fronts and turbulent eddies. This is particularly evident along major western boundary currents, such as the Gulf Stream and Kuroshio, as well as the Agulhas Retroflection and the Antarctic Circumpolar Current. In these regions, small spatial disagreements between members regarding the exact placement of a meandering current or a newly formed eddy translate into high local variance.

In contrast, the uncertainty patterns for salinity and ocean currents are governed by distinctly different physical drivers. Salinity deviations are heavily dominated by freshwater dynamics, with the highest ensemble spread localized around massive river outflows, such as the Amazon, Congo, and Ganges-Brahmaputra plumes. In these areas, slight variations in predicted coastal winds or surface currents drastically shift the floating lenses of fresh water. For velocity, particularly the zonal current, the ensemble standard deviation features a striking band of uncertainty directly across the Equator. This highlights the model's spread in resolving Tropical Instability Waves and the chaotic shear between opposing equatorial currents, compounding the variance already seen in eddy-rich boundary regions.

\clearpage

\subsection{Regional metrics}

We evaluate Njord on the Baltic Sea against the deterministic SeaCast baseline and a persistence forecast. For both Njord-Baltic and SeaCast we report two variants: a \emph{reanalysis} model, pretrained on reanalysis only, and an \emph{analysis} model, additionally finetuned on operational analysis. All systems are initialized from the Baltic Sea analysis, forced with the IFS 10-day atmospheric forecast at the surface, and constrained at the lateral boundary by the GLO12 10-day ocean forecast from OceanBench. GLO12 forecasts covering the Baltic Sea are also used as a reference in the comparison. While it retains some large-scale skill, it lacks the spatial resolution required to capture fine-scale dynamics in the Baltic Sea. Interpolation of GLO12 to the regional grid also results in some missing values in narrow coastal regions which are not used in the error calculation. The ground truth used for verification is the Baltic Sea analysis. Ensemble metrics for Njord-Baltic: \gls{RMSE} of the ensemble mean, \gls{CRPS}, and \gls{SSR}, are compared against deterministic \gls{RMSE} and \gls{MAE} for SeaCast and persistence. The SeaCast/persistence \gls{MAE} is
shown alongside \gls{CRPS} as a deterministic reference.

\begin{figure}[tbh]
    \centering
    \includegraphics[width=\textwidth]{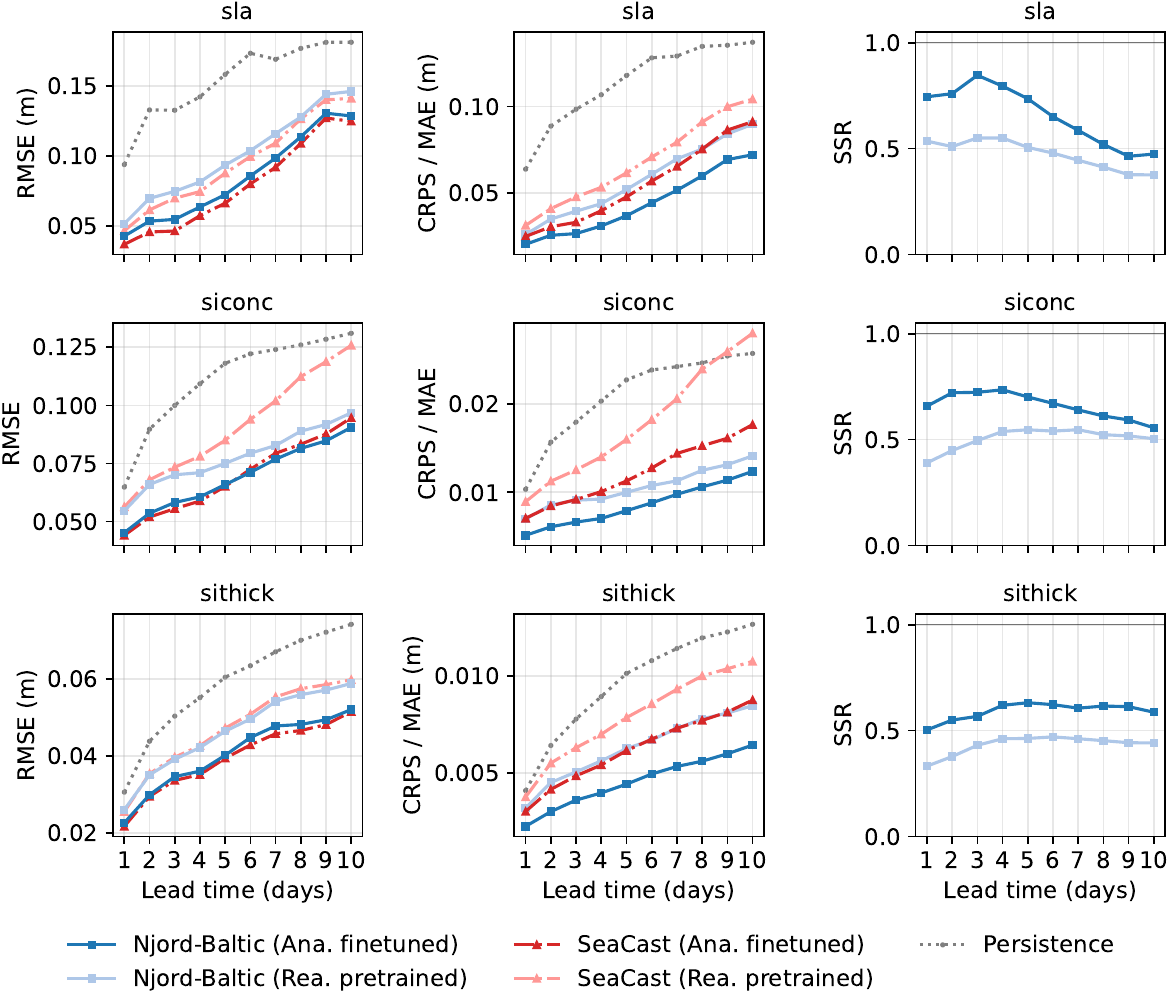}
    \caption{Surface variables: \gls{SLA}, \gls{SIC}
    and \gls{SIT}. Reanalysis variants are shown dashed and
    analysis variants solid.}
    \label{fig:baltic_metrics_surface}
\end{figure}

\begin{figure}[tbh]
    \centering
    \includegraphics[width=\textwidth]{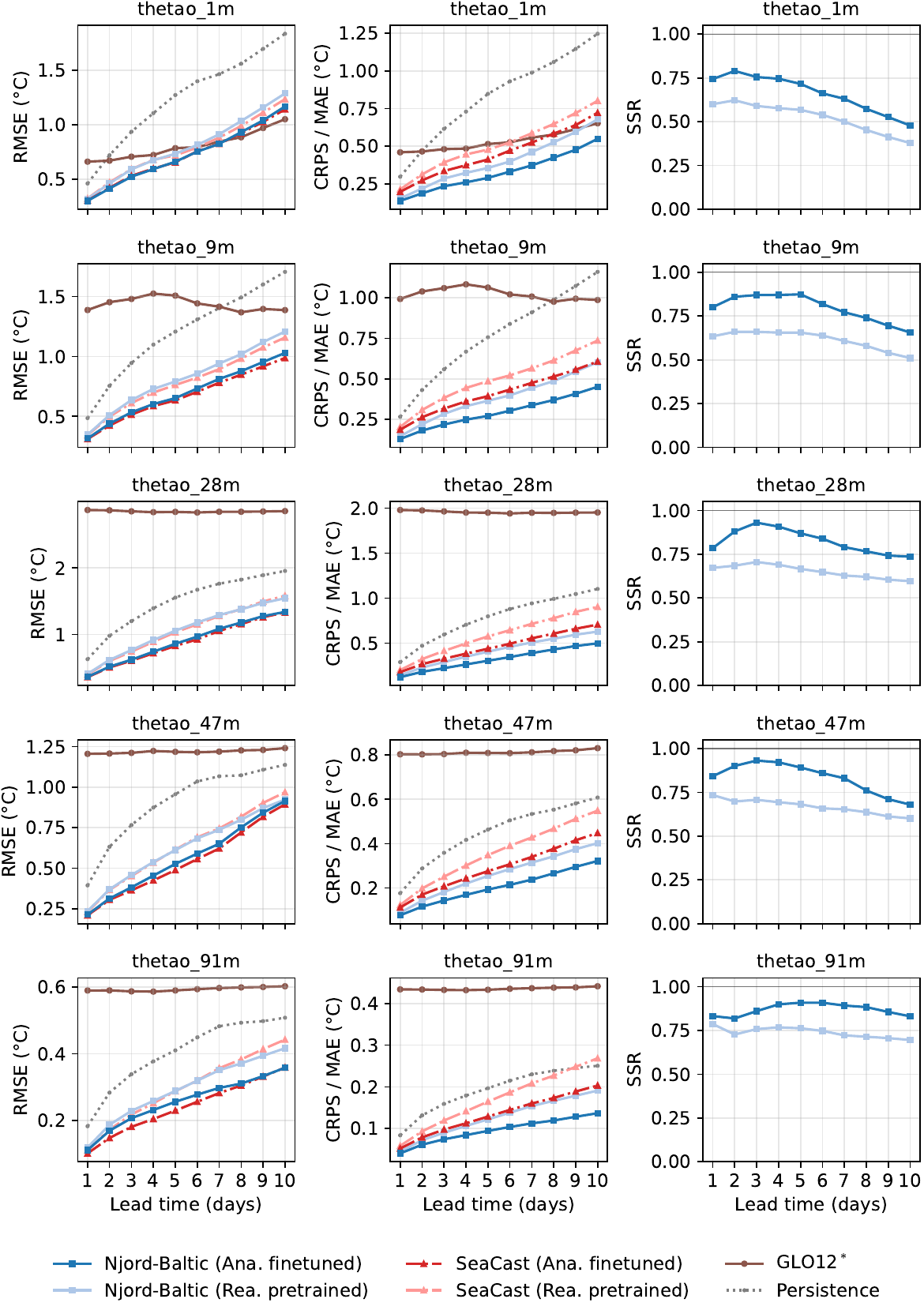}
    \caption{Temperature at 1, 9, 28, 47 and 91\,m depth.}
    \label{fig:baltic_metrics_thetao}
\end{figure}

\begin{figure}[tbh]
    \centering
    \includegraphics[width=\textwidth]{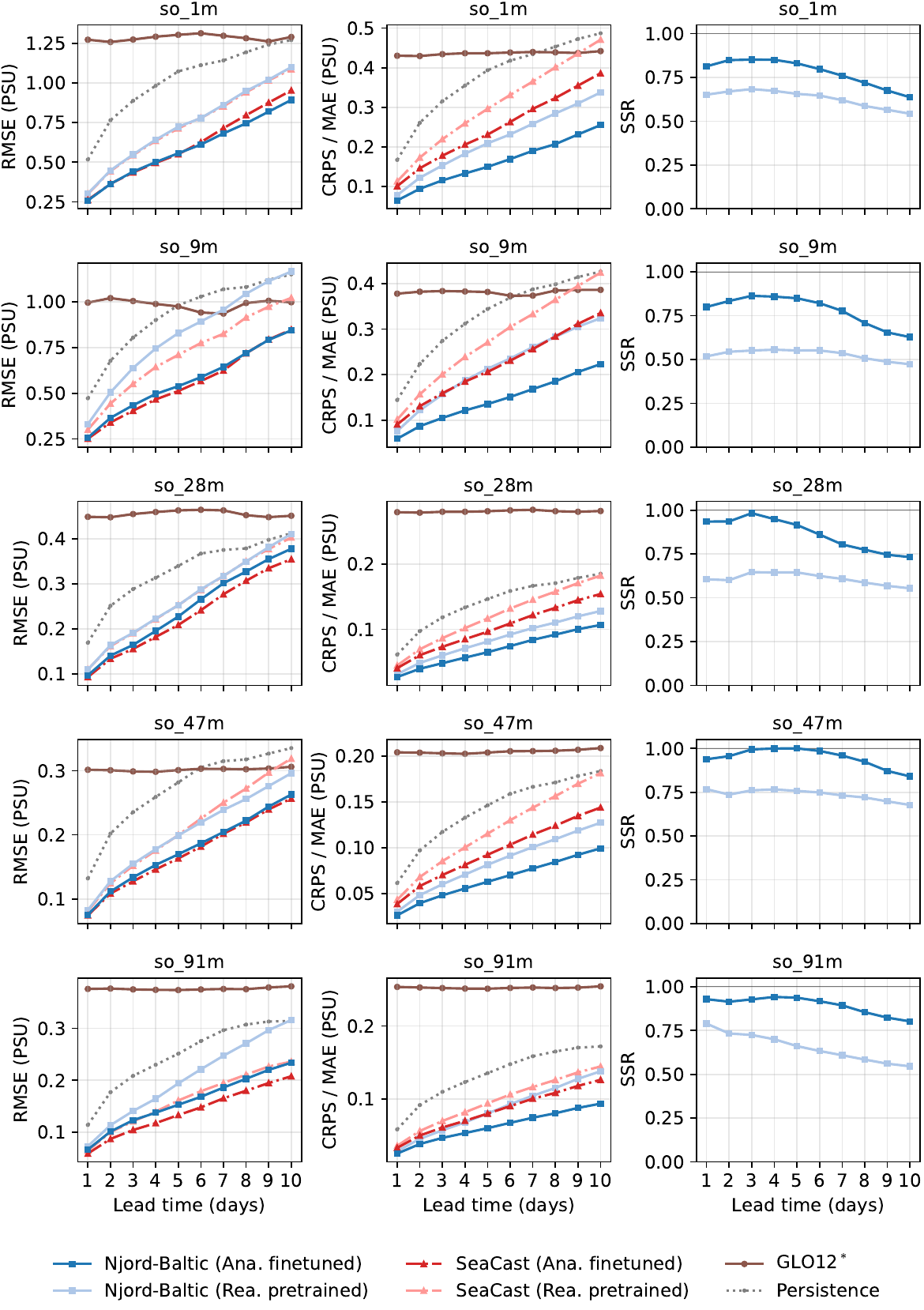}
    \caption{Salinity at 1, 9, 28, 47 and 91\,m depth.}
    \label{fig:metrics_so}
\end{figure}

\begin{figure}[tbh]
    \centering
    \includegraphics[width=\textwidth]{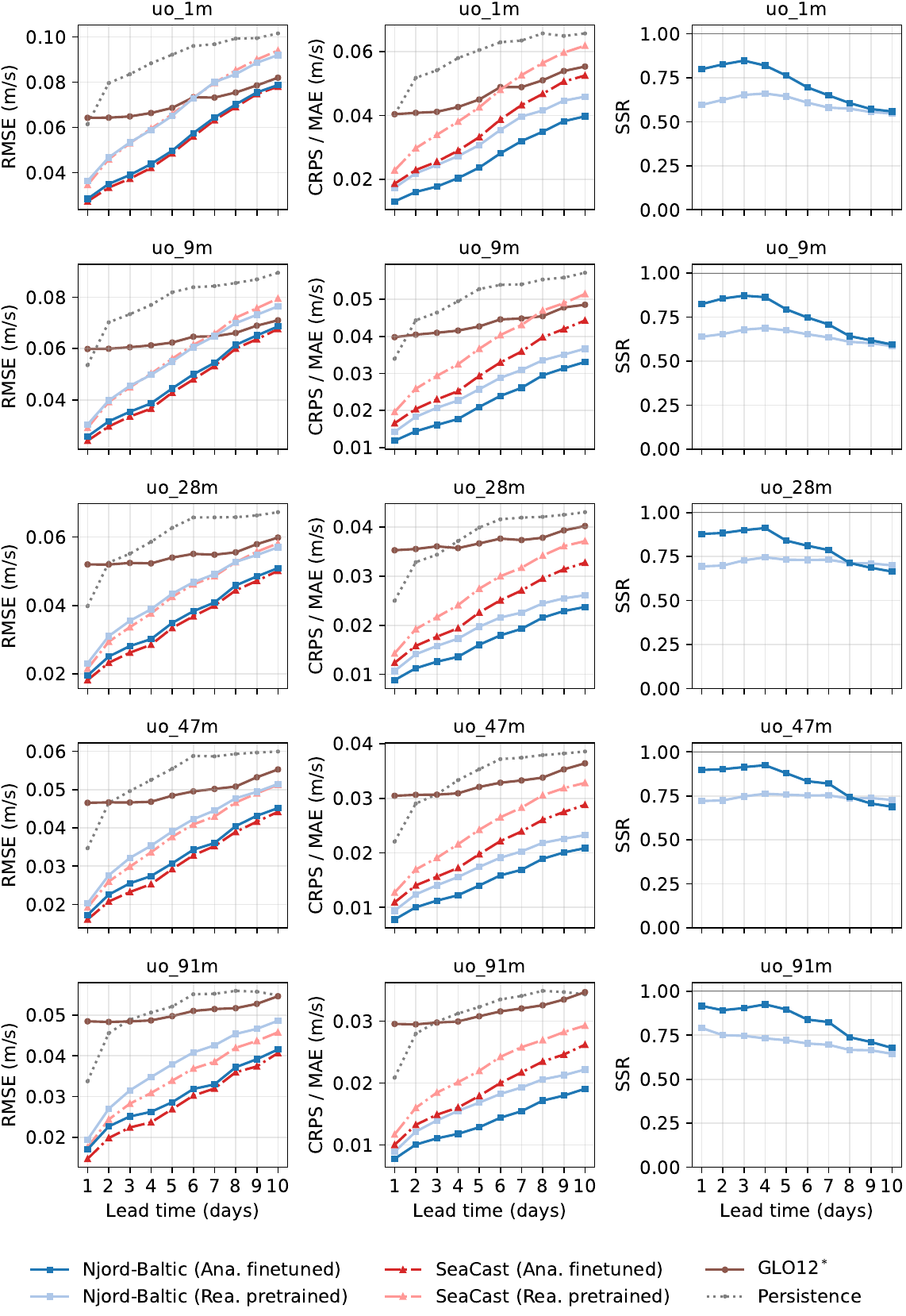}
    \caption{Meridional current at 1, 9, 28, 47 and 91\,m depth.}
    \label{fig:baltic_metrics_uo}
\end{figure}

\Crefrange{fig:baltic_metrics_surface}{fig:baltic_metrics_uo} report metrics per variable and lead time. The \gls{SSR} is defined as the ratio between the standard deviation of the ensemble and the \gls{RMSE} of the ensemble mean; values close to one indicate a well-calibrated ensemble. Because zonal and meridional currents exhibit similar error accumulation patterns, only the zonal components are shown here for brevity.

\clearpage

\clearpage

\subsection{Regional forecasts}

\begin{figure}[tbh]
    \centering
    \includegraphics[width=\textwidth]{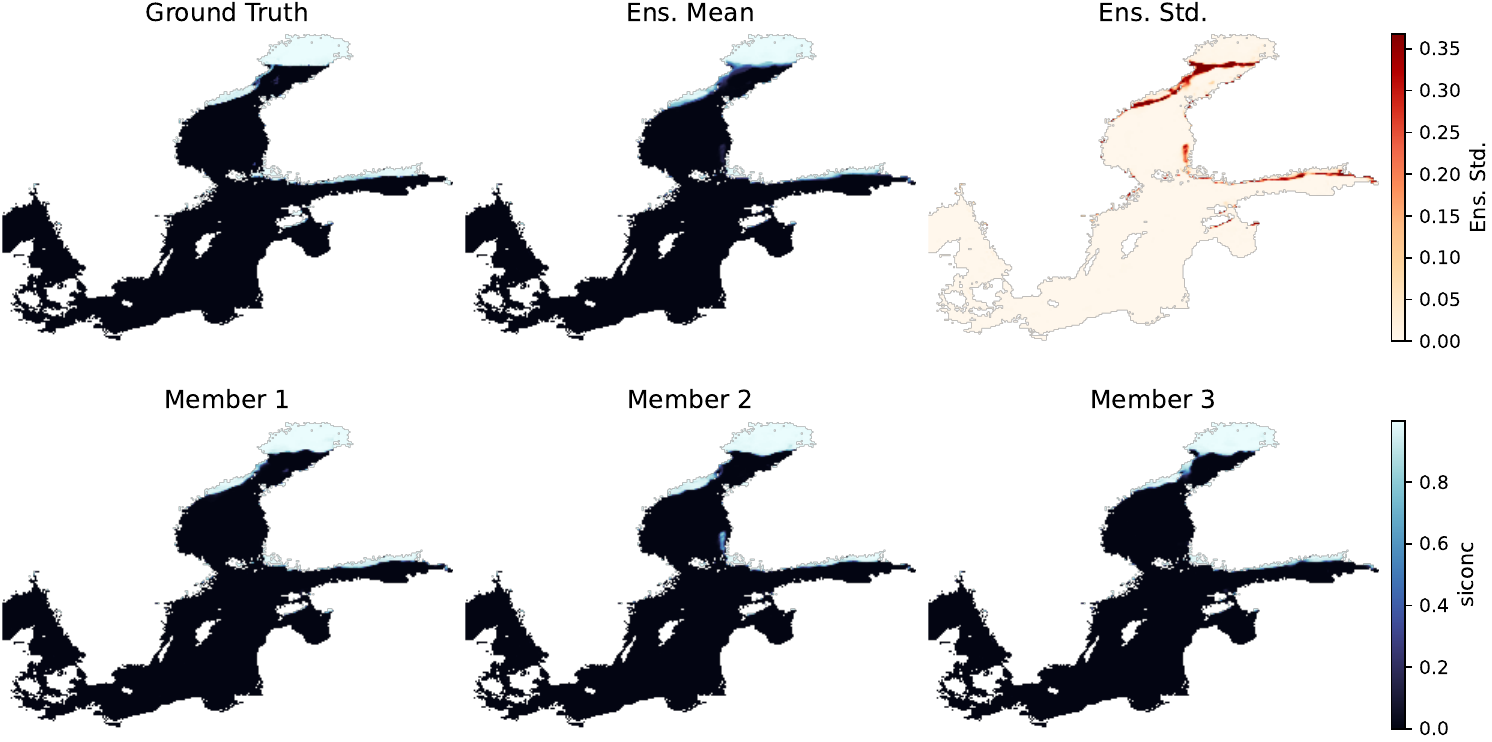}
    \caption{Sea ice concentration at lead time 10\,d, init 2024-02-20.}
    \label{fig:example_siconc}
\end{figure}

\begin{figure}[tbh]
    \centering
    \includegraphics[width=\textwidth]{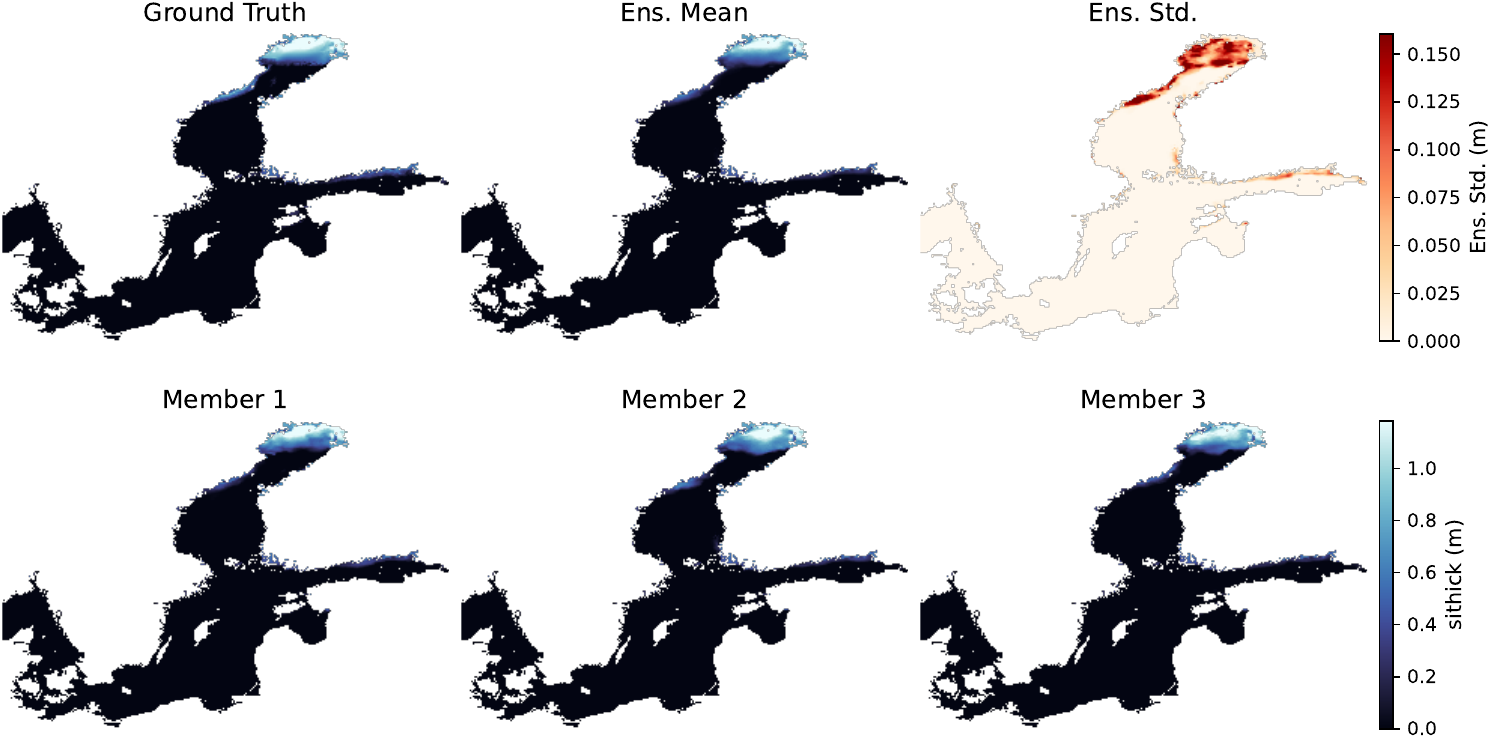}
    \caption{Sea ice thickness at lead time 10\,d, init 2024-02-20.}
    \label{fig:example_sithick}
\end{figure}

\begin{figure}[p]
    \centering
    \includegraphics[width=\textwidth]{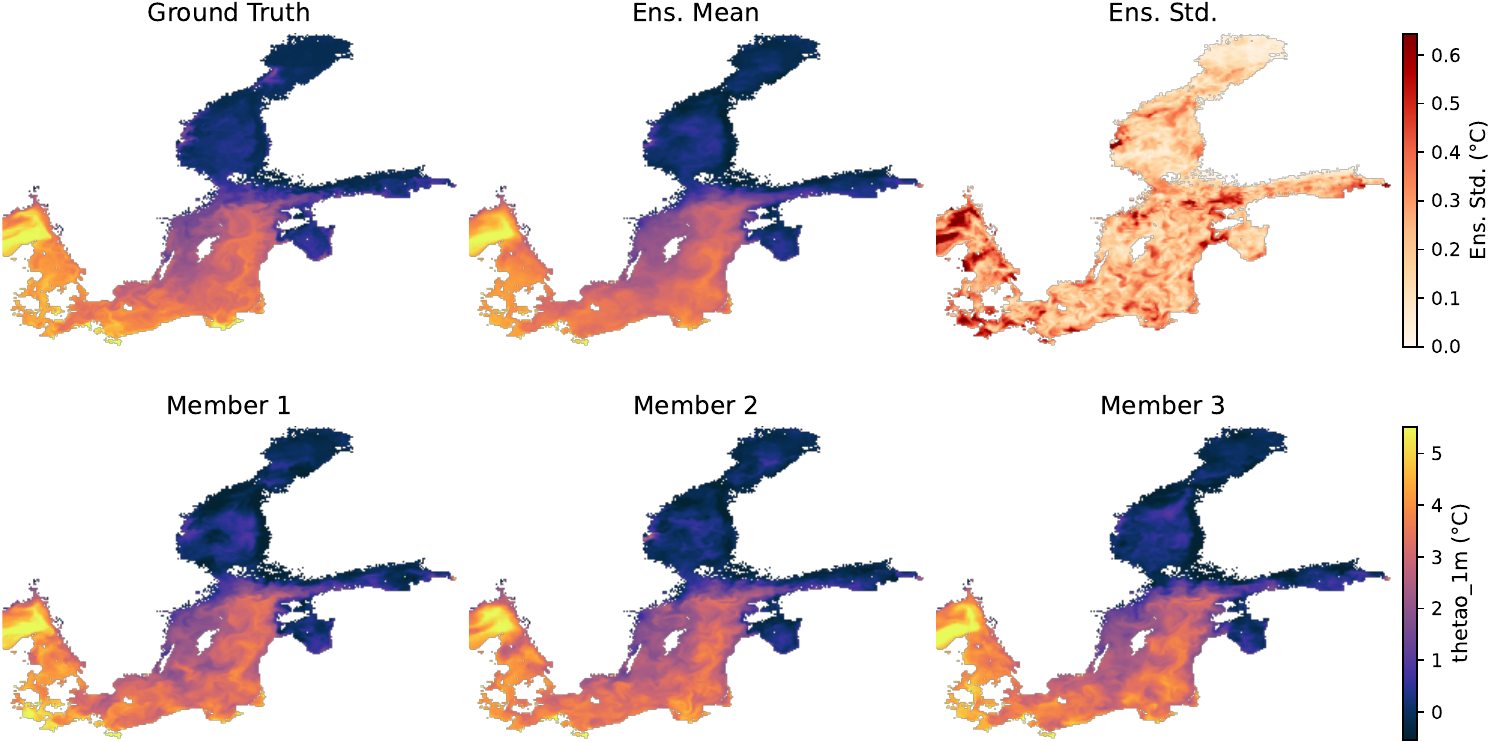}
    \caption{Temperature at the surface, lead time 10\,d, init 2024-02-20.}
    \label{fig:example_thetao_1m}
\end{figure}

\begin{figure}[p]
    \centering
    \includegraphics[width=\textwidth]{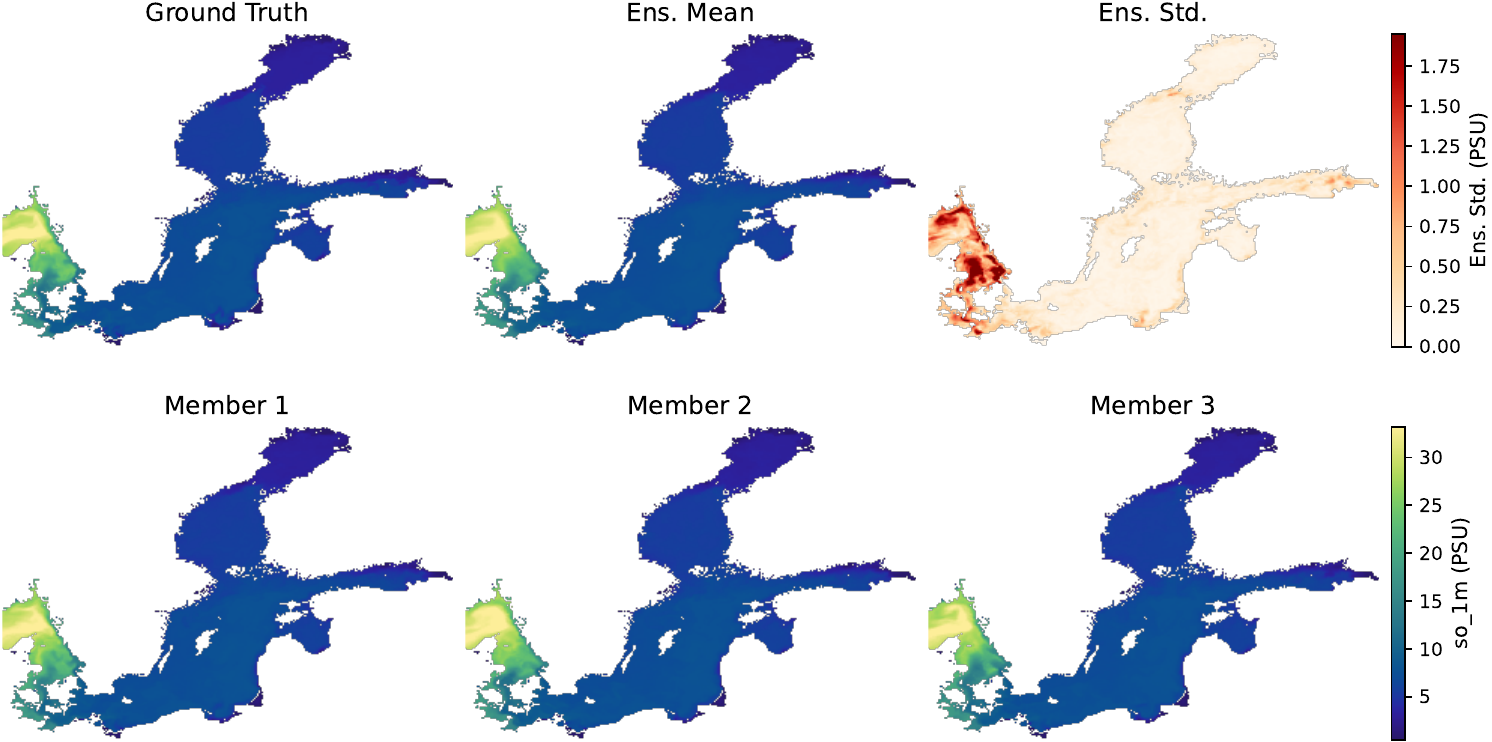}
    \caption{Salinity at the surface, lead time 10\,d, init 2024-02-20.}
    \label{fig:example_so_1m}
\end{figure}

\begin{figure}[p]
    \centering
    \includegraphics[width=\textwidth]{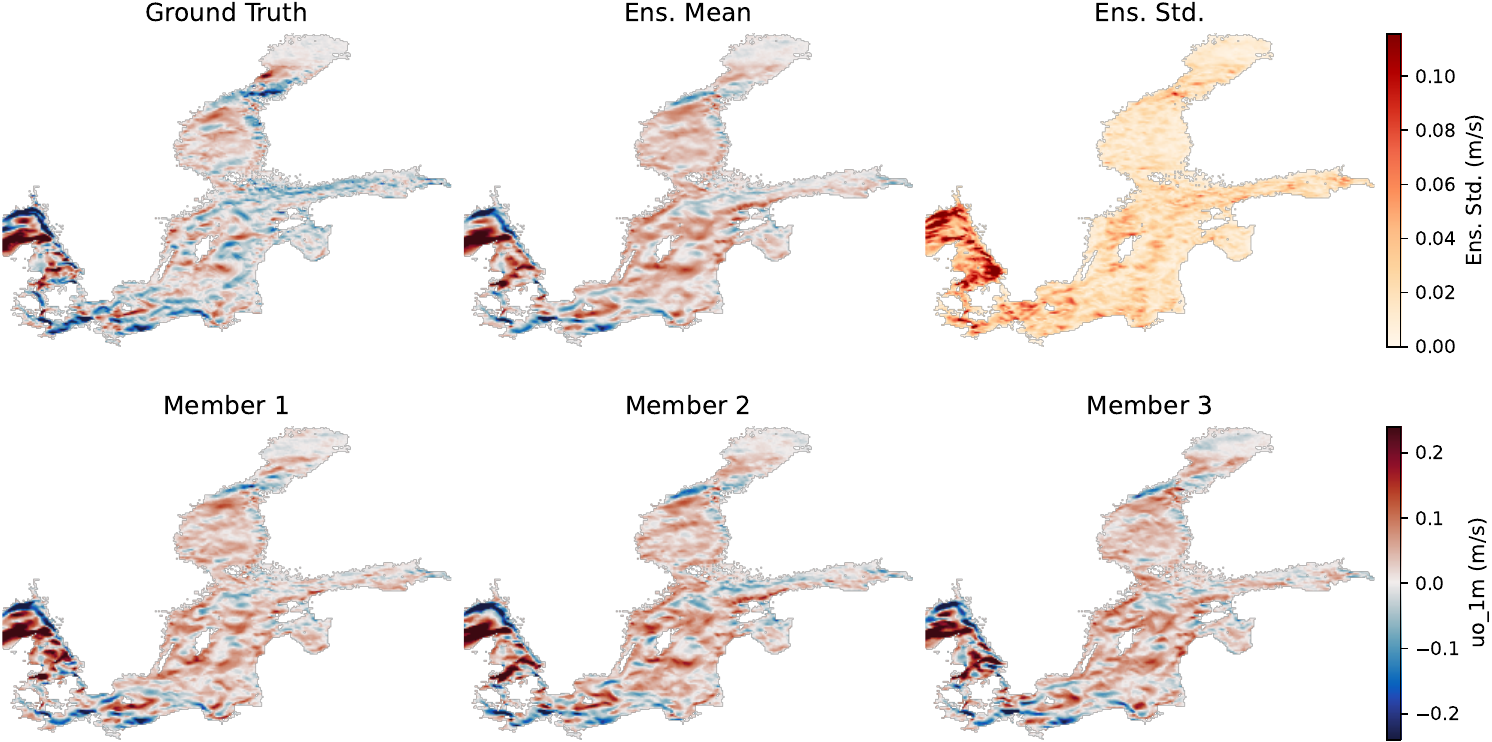}
    \caption{Zonal current at the surface, lead time 10\,d, init 2024-02-20.}
    \label{fig:example_uo_1m}
\end{figure}

\begin{figure}[p]
    \centering
    \includegraphics[width=\textwidth]{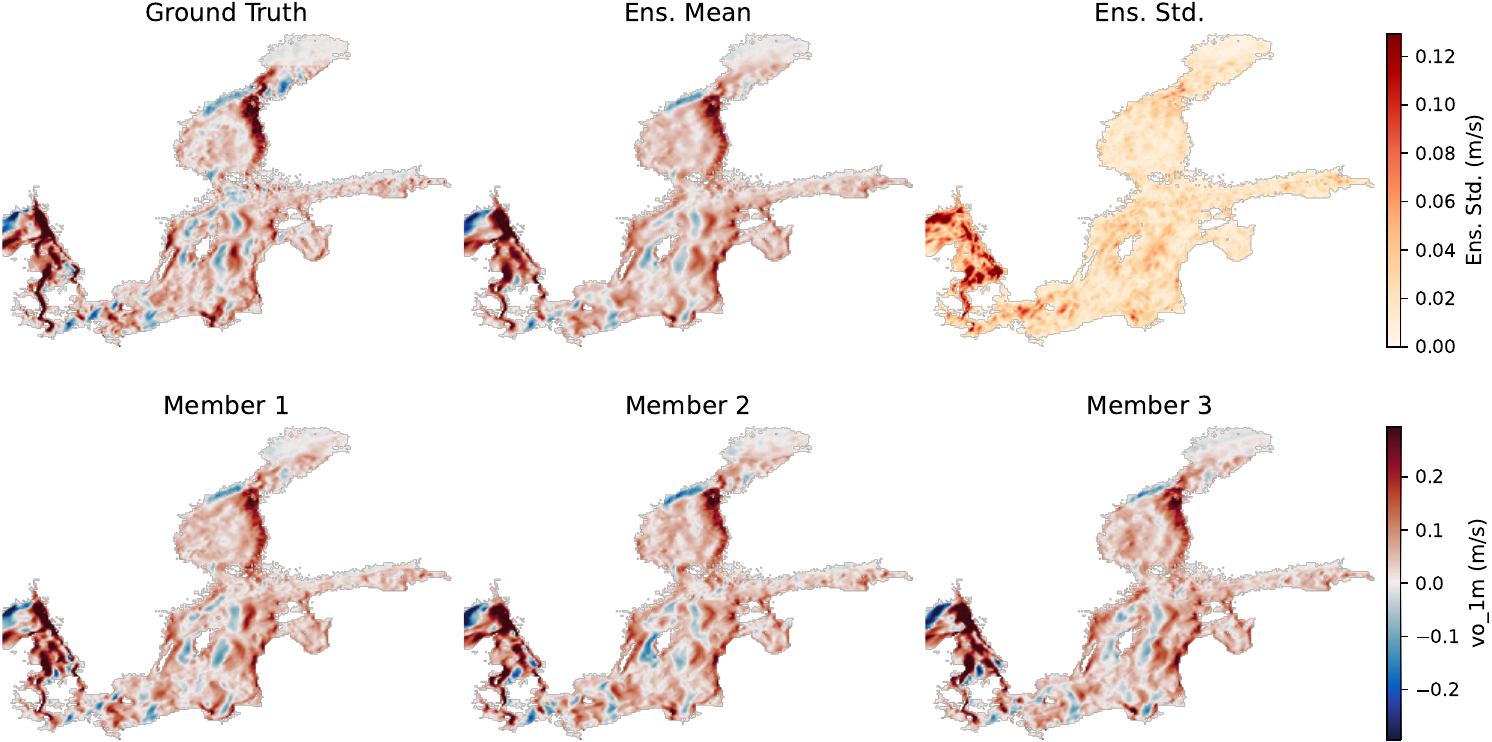}
    \caption{Meridional current at the surface, lead time 10\,d, init 2024-02-20.}
    \label{fig:example_vo_1m}
\end{figure}

\begin{figure}[p]
    \centering
    \includegraphics[width=\textwidth]{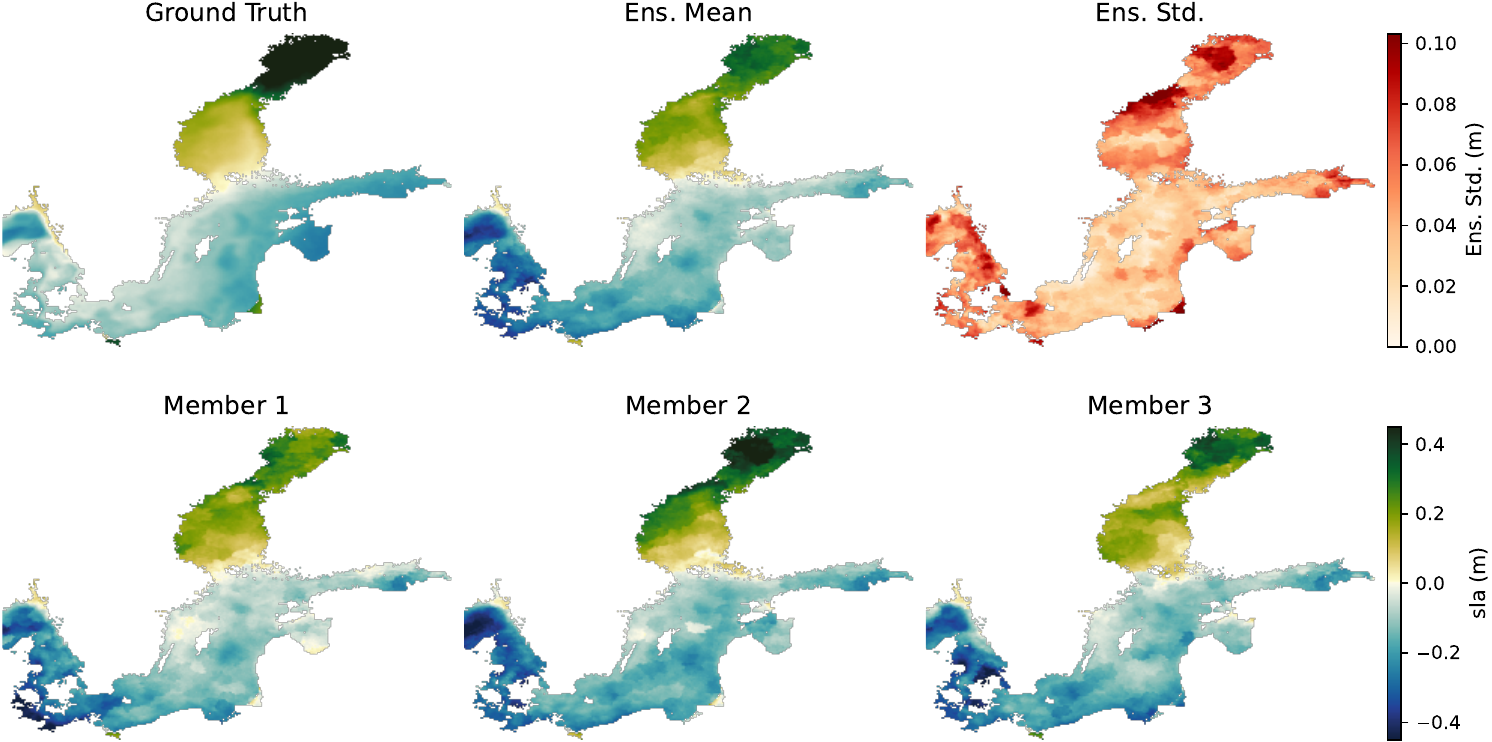}
    \caption{Sea level anomaly at lead time 10\,d, init 2024-02-20.}
    \label{fig:example_sla}
\end{figure}

To illustrate the qualitative behavior of Njord-Baltic in a high-resolution regional context, we present example ensemble forecasts for the Baltic Sea, initialized on 20 February 2024 at a lead time of 10 days, in \crefrange{fig:example_siconc}{fig:example_sla}. For each variable, we show the analysis target, the ensemble mean, the ensemble standard deviation, and three individual ensemble members.

The individual ensemble members appear sharp and exhibit noticeable variability, whereas the ensemble mean is smoother due to averaging. Sea-ice fields display well-defined edges and are exactly zero in ice-free regions, reflecting the use of clamping and a dedicated density channel. For this late-winter date, ice is primarily confined to the Bay of Bothnia and the Gulf of Finland. The ensemble standard deviation clearly highlights the marginal ice zone, marking the uncertainty in the exact location of the ice edge. For \gls{SIT}, uncertainty also extends into the interior of the ice pack, reflecting ensemble disagreements on dynamic thickening processes such as ridging.

The physical drivers of uncertainty for salinity and potential temperature are distinctly visible. Salinity variance is overwhelmingly concentrated in the Skagerrak and Kattegat straits. This transition zone is highly dynamic, as dense, saline North Sea water forcefully mixes with the fresh, brackish outflow of the Baltic; small ensemble disagreements on the exact timing, volume, or extent of these inflows create massive local variance. Potential temperature uncertainty is also elevated in these straits but extends further into the Baltic Proper, reflecting complex thermal mixing fronts and internal mesoscale eddies.

Surface currents exhibit widespread uncertainty across the entire basin. Because the Baltic Sea's surface circulation is heavily wind-driven, ensemble spread in these velocities reflects the chaotic, rapid response of surface waters to varying meteorological forcing across the members. This variance naturally peaks in the narrow, high-flow bottlenecks of the Danish straits.

Finally, SLA shows elevated uncertainty not just in the straits, but also at the northern and eastern extremities of the basin. In a shallow, enclosed sea, water levels are highly sensitive to wind stress piling water up against the coasts (storm surges and seiches). Furthermore, some visual artifacts remain apparent in the SLA standard deviation. This could potentially arise because SLA fields are derived from interpolating sparse along-track satellite observations, resulting in noisy targets that may require higher temporal resolution and denser training samples to model smoothly.

\clearpage
\input{checklist.tex}

\end{document}

%% file: acronyms.tex
\newacronym{RNN}{RNN}{Recurrent Neural Network}
\newacronym{CNN}{CNN}{Convolutional Neural Network}
\newacronym[plural=GNNs, longplural={Graph Neural Networks}]{GNN}{GNN}{Graph Neural Network}
\newacronym{MPNN}{MPNN}{Message Passing Neural Network}
\newacronym{GRU}{GRU}{Gated Recurrent Unit}
\newacronym{MLP}{MLP}{Multi-Layer Perceptron}
\newacronym{VAE}{VAE}{Variational Auto-Encoder}
\newacronym{CVAE}{CVAE}{Conditional Variational Auto-Encoder}

\newacronym{RMSE}{RMSE}{Root Mean Squared Error}
\newacronym{MSE}{MSE}{Mean Squared Error}
\newacronym{MAE}{MAE}{Mean Absolute Error}
\newacronym{CRPS}{CRPS}{Continuous Ranked Probability Score}
\newacronym{ELBO}{ELBO}{Evidence Lower Bound}
\newacronym{SSR}{SSR}{Spread-Skill Ratio}
\newacronym{CDF}{CDF}{Cumulative Distribution Function}

\newacronym{ODE}{ODE}{Ordinary Differential Equation}
\newacronym{PDE}{PDE}{Partial Differential Equation}
\newacronym{SDE}{SDE}{Stochastic Differential Equation}

\newacronym{GP}{GP}{Gaussian Process}
\newacronym{GMRF}{GMRF}{Gaussian Markov Random Field}

\newacronym{ROM}{ROM}{Regional Ocean Model}
\newacronym{SST}{SST}{Sea Surface Temperature}
\newacronym{SSH}{SSH}{Sea Surface Height}
\newacronym{SSS}{SSS}{Sea Surface Salinity}
\newacronym{SLA}{SLA}{Sea Level Anomaly}
\newacronym{SIT}{SIT}{Sea Ice Thickness}
\newacronym{SIC}{SIC}{Sea Ice Concentration}
\newacronym{S}{S}{Salinity}
\newacronym{T}{T}{Temperature}
\newacronym{U}{U}{Zonal Current}
\newacronym{V}{V}{Meridional Current}

%% file: tables/glorys12_track.tex
\resizebox{\textwidth}{!}{%
\begin{tabular}{l cccc c cccc c cccc c cccc}
\toprule
& \multicolumn{4}{c}{\begin{tabular}{@{}c@{}}\textbf{Zonal geo. current}\end{tabular}} && \multicolumn{4}{c}{\begin{tabular}{@{}c@{}}\textbf{Meridional geo. current}\end{tabular}} && \multicolumn{4}{c}{\textbf{SSH}} && \multicolumn{4}{c}{} \\
& \multicolumn{4}{c}{RMSE (m/s)} && \multicolumn{4}{c}{RMSE (m/s)} && \multicolumn{4}{c}{RMSE (m)} && \multicolumn{4}{c}{} \\
\cmidrule(lr){2-5} \cmidrule(lr){7-10} \cmidrule(lr){12-15}
Lead time (days) & 1 & 4 & 7 & 10 && 1 & 4 & 7 & 10 && 1 & 4 & 7 & 10 && & & & \\
\cmidrule{1-15}
GLO12 & \cellcolor{gray!15}0.131 & \cellcolor{gray!15}0.137 & \cellcolor{gray!15}0.144 & \cellcolor{gray!15}0.151 & & \cellcolor{gray!15}0.123 & \cellcolor{gray!15}0.129 & \cellcolor{gray!15}0.137 & \cellcolor{gray!15}0.143 & & \cellcolor{gray!15}0.069 & \cellcolor{gray!15}0.072 & \cellcolor{gray!15}0.077 & \cellcolor{gray!15}0.082 & & & & & \\
GLONET & \cellcolor{red!35}0.151 & \cellcolor{red!45}0.186 & \cellcolor{red!45}0.234 & \cellcolor{red!45}0.277 & & \cellcolor{red!45}0.177 & \cellcolor{red!45}0.218 & \cellcolor{red!45}0.289 & \cellcolor{red!45}0.323 & & \cellcolor{red!21}0.075 & \cellcolor{red!18}0.077 & \cellcolor{red!21}0.084 & \cellcolor{red!18}0.089 & & & & & \\
WENHAI & \cellcolor{red!45}1.375 & \cellcolor{red!45}1.376 & \cellcolor{red!45}1.380 & \cellcolor{red!45}1.384 & & \cellcolor{red!45}2.003 & \cellcolor{red!45}2.003 & \cellcolor{red!45}2.011 & \cellcolor{red!45}2.024 & & \cellcolor{red!45}0.118 & \cellcolor{red!45}0.122 & \cellcolor{red!45}0.126 & \cellcolor{red!45}0.131 & & & & & \\
XIHE & \cellcolor{red!45}0.658 & \cellcolor{red!45}0.640 & \cellcolor{red!45}0.595 & \cellcolor{red!45}0.621 & & \cellcolor{red!45}0.724 & \cellcolor{red!45}0.900 & \cellcolor{red!45}0.872 & \cellcolor{red!45}0.809 & & \cellcolor{red!33}0.079 & \cellcolor{red!41}0.084 & \cellcolor{red!33}0.088 & \cellcolor{red!23}0.091 & & & & & \\
\textbf{NJORD} & \cellcolor{blue!17}0.121 & \cellcolor{blue!11}0.130 & \cellcolor{blue!7}0.140 & 0.149 & & \cellcolor{blue!9}0.118 & \cellcolor{blue!5}0.126 & \cellcolor{blue!3}0.135 & 0.143 & & \cellcolor{red!6}0.070 & \cellcolor{red!12}0.075 & \cellcolor{red!6}0.079 & 0.082 & & & & & \\
\midrule
& \multicolumn{4}{c}{\textbf{Zonal current}} && \multicolumn{4}{c}{\textbf{Meridional current}} && \multicolumn{4}{c}{\textbf{Temperature}} && \multicolumn{4}{c}{\textbf{Salinity}} \\
& \multicolumn{4}{c}{RMSE (m/s)} && \multicolumn{4}{c}{RMSE (m/s)} && \multicolumn{4}{c}{RMSE (°C)} && \multicolumn{4}{c}{RMSE (PSU)} \\
\cmidrule(lr){2-5} \cmidrule(lr){7-10} \cmidrule(lr){12-15} \cmidrule(lr){17-20}
Lead time (days) & 1 & 4 & 7 & 10 && 1 & 4 & 7 & 10 && 1 & 4 & 7 & 10 && 1 & 4 & 7 & 10 \\
\midrule \multicolumn{20}{c}{\textbf{Depth: 0.49m}} \\ \midrule
GLO12 & \cellcolor{gray!15}0.114 & \cellcolor{gray!15}0.122 & \cellcolor{gray!15}0.134 & \cellcolor{gray!15}0.145 & & \cellcolor{gray!15}0.113 & \cellcolor{gray!15}0.121 & \cellcolor{gray!15}0.132 & \cellcolor{gray!15}0.143 & & \cellcolor{gray!15}0.545 & \cellcolor{gray!15}0.559 & \cellcolor{gray!15}0.591 & \cellcolor{gray!15}0.635 & & \cellcolor{gray!15}0.729 & \cellcolor{gray!15}0.729 & \cellcolor{gray!15}0.732 & \cellcolor{gray!15}0.737 \\
GLONET & \cellcolor{red!21}0.125 & \cellcolor{red!8}0.127 & 0.135 & 0.144 & & \cellcolor{red!21}0.124 & \cellcolor{red!6}0.124 & 0.131 & \cellcolor{blue!8}0.138 & & \cellcolor{red!44}0.653 & \cellcolor{red!45}0.689 & \cellcolor{red!45}0.823 & \cellcolor{red!45}0.913 & & \cellcolor{red!17}0.784 & \cellcolor{red!17}0.785 & \cellcolor{red!21}0.801 & \cellcolor{red!17}0.794 \\
WENHAI & \cellcolor{red!45}0.175 & \cellcolor{red!45}0.183 & \cellcolor{red!45}0.191 & \cellcolor{red!45}0.201 & & \cellcolor{red!45}0.169 & \cellcolor{red!45}0.174 & \cellcolor{red!45}0.174 & \cellcolor{red!45}0.178 & & \cellcolor{red!38}0.637 & \cellcolor{red!45}0.777 & \cellcolor{red!45}0.956 & \cellcolor{red!45}1.144 & & \cellcolor{red!45}1.165 & \cellcolor{red!45}1.150 & \cellcolor{red!45}1.139 & \cellcolor{red!45}1.132 \\
XIHE & \cellcolor{red!20}0.125 & 0.123 & \cellcolor{blue!18}0.123 & \cellcolor{blue!32}0.125 & & \cellcolor{red!18}0.122 & 0.121 & \cellcolor{blue!21}0.120 & \cellcolor{blue!35}0.121 & & \cellcolor{red!44}0.651 & \cellcolor{red!45}0.679 & \cellcolor{red!37}0.690 & \cellcolor{red!45}0.792 & & \cellcolor{blue!3}0.720 & 0.734 & \cellcolor{blue!8}0.706 & \cellcolor{blue!14}0.691 \\
\textbf{NJORD} & \cellcolor{red!12}0.121 & \cellcolor{blue!5}0.119 & \cellcolor{blue!15}0.126 & \cellcolor{blue!17}0.134 & & \cellcolor{red!13}0.120 & \cellcolor{blue!5}0.118 & \cellcolor{blue!15}0.124 & \cellcolor{blue!20}0.131 & & 0.548 & \cellcolor{blue!3}0.551 & \cellcolor{blue!4}0.580 & \cellcolor{blue!6}0.618 & & \cellcolor{red!35}0.842 & \cellcolor{red!35}0.840 & \cellcolor{red!34}0.844 & \cellcolor{red!33}0.846 \\
\midrule \multicolumn{20}{c}{\textbf{Depth: 50m}} \\ \midrule
GLO12 & \cellcolor{gray!15}0.112 & \cellcolor{gray!15}0.118 & \cellcolor{gray!15}0.125 & \cellcolor{gray!15}0.132 & & \cellcolor{gray!15}0.110 & \cellcolor{gray!15}0.116 & \cellcolor{gray!15}0.124 & \cellcolor{gray!15}0.131 & & \cellcolor{gray!15}0.952 & \cellcolor{gray!15}0.960 & \cellcolor{gray!15}0.979 & \cellcolor{gray!15}1.000 & & \cellcolor{gray!15}0.324 & \cellcolor{gray!15}0.325 & \cellcolor{gray!15}0.326 & \cellcolor{gray!15}0.328 \\
GLONET & 0.111 & \cellcolor{blue!14}0.110 & \cellcolor{blue!18}0.116 & \cellcolor{blue!14}0.124 & & 0.109 & \cellcolor{blue!13}0.109 & \cellcolor{blue!16}0.115 & \cellcolor{blue!14}0.123 & & 0.951 & \cellcolor{red!8}0.996 & \cellcolor{red!29}1.105 & \cellcolor{red!45}1.261 & & \cellcolor{red!24}0.359 & \cellcolor{red!29}0.366 & \cellcolor{red!35}0.378 & \cellcolor{red!39}0.386 \\
WENHAI & \cellcolor{red!45}0.160 & \cellcolor{red!45}0.163 & \cellcolor{red!45}0.168 & \cellcolor{red!45}0.175 & & \cellcolor{red!45}0.153 & \cellcolor{red!45}0.155 & \cellcolor{red!45}0.157 & \cellcolor{red!45}0.161 & & \cellcolor{blue!10}0.909 & \cellcolor{blue!7}0.930 & 0.976 & \cellcolor{red!7}1.032 & & \cellcolor{red!45}1.119 & \cellcolor{red!45}1.117 & \cellcolor{red!45}1.116 & \cellcolor{red!45}1.116 \\
XIHE & \cellcolor{red!4}0.114 & \cellcolor{blue!16}0.110 & \cellcolor{blue!29}0.109 & \cellcolor{blue!38}0.110 & & \cellcolor{red!7}0.113 & \cellcolor{blue!12}0.109 & \cellcolor{blue!28}0.108 & \cellcolor{blue!38}0.108 & & \cellcolor{blue!33}0.813 & \cellcolor{blue!30}0.832 & \cellcolor{blue!39}0.810 & \cellcolor{blue!18}0.922 & & \cellcolor{red!30}0.367 & \cellcolor{red!36}0.377 & \cellcolor{red!35}0.377 & \cellcolor{red!45}0.403 \\
\textbf{NJORD} & \cellcolor{red!6}0.115 & \cellcolor{blue!12}0.112 & \cellcolor{blue!21}0.114 & \cellcolor{blue!24}0.118 & & \cellcolor{red!9}0.114 & \cellcolor{blue!10}0.111 & \cellcolor{blue!19}0.113 & \cellcolor{blue!24}0.117 & & \cellcolor{blue!24}0.850 & \cellcolor{blue!29}0.836 & \cellcolor{blue!33}0.835 & \cellcolor{blue!35}0.844 & & \cellcolor{red!36}0.376 & \cellcolor{red!34}0.375 & \cellcolor{red!34}0.375 & \cellcolor{red!33}0.377 \\
\midrule \multicolumn{20}{c}{\textbf{Depth: 100m}} \\ \midrule
GLO12 & \cellcolor{gray!15}0.110 & \cellcolor{gray!15}0.115 & \cellcolor{gray!15}0.121 & \cellcolor{gray!15}0.127 & & \cellcolor{gray!15}0.107 & \cellcolor{gray!15}0.112 & \cellcolor{gray!15}0.119 & \cellcolor{gray!15}0.125 & & \cellcolor{gray!15}0.932 & \cellcolor{gray!15}0.941 & \cellcolor{gray!15}0.963 & \cellcolor{gray!15}0.985 & & \cellcolor{gray!15}0.225 & \cellcolor{gray!15}0.226 & \cellcolor{gray!15}0.227 & \cellcolor{gray!15}0.229 \\
GLONET & 0.111 & \cellcolor{blue!10}0.110 & \cellcolor{blue!15}0.113 & \cellcolor{blue!15}0.118 & & 0.106 & \cellcolor{blue!14}0.105 & \cellcolor{blue!20}0.108 & \cellcolor{blue!23}0.112 & & \cellcolor{red!20}1.014 & \cellcolor{red!25}1.047 & \cellcolor{red!35}1.111 & \cellcolor{red!45}1.224 & & \cellcolor{red!21}0.247 & \cellcolor{red!25}0.251 & \cellcolor{red!30}0.257 & \cellcolor{red!35}0.264 \\
WENHAI & \cellcolor{red!45}0.141 & \cellcolor{red!45}0.142 & \cellcolor{red!45}0.146 & \cellcolor{red!41}0.150 & & \cellcolor{red!45}0.136 & \cellcolor{red!45}0.137 & \cellcolor{red!40}0.140 & \cellcolor{red!32}0.142 & & \cellcolor{red!27}1.044 & \cellcolor{red!25}1.046 & \cellcolor{red!23}1.062 & \cellcolor{red!22}1.082 & & \cellcolor{red!45}1.058 & \cellcolor{red!45}1.058 & \cellcolor{red!45}1.057 & \cellcolor{red!45}1.057 \\
XIHE & \cellcolor{red!7}0.113 & \cellcolor{blue!15}0.107 & \cellcolor{blue!29}0.106 & \cellcolor{blue!38}0.106 & & \cellcolor{red!5}0.109 & \cellcolor{blue!17}0.104 & \cellcolor{blue!32}0.102 & \cellcolor{blue!42}0.101 & & \cellcolor{red!6}0.958 & \cellcolor{red!15}1.004 & \cellcolor{red!8}0.999 & \cellcolor{red!17}1.059 & & 0.228 & 0.228 & \cellcolor{red!5}0.232 & \cellcolor{red!16}0.245 \\
\textbf{NJORD} & \cellcolor{red!11}0.115 & \cellcolor{blue!8}0.111 & \cellcolor{blue!18}0.112 & \cellcolor{blue!21}0.115 & & \cellcolor{red!10}0.112 & \cellcolor{blue!9}0.108 & \cellcolor{blue!19}0.109 & \cellcolor{blue!25}0.111 & & \cellcolor{red!20}1.014 & \cellcolor{red!16}1.008 & \cellcolor{red!13}1.018 & \cellcolor{red!11}1.033 & & \cellcolor{red!27}0.253 & \cellcolor{red!23}0.249 & \cellcolor{red!21}0.248 & \cellcolor{red!19}0.248 \\
\midrule \multicolumn{20}{c}{\textbf{Depth: 200m}} \\ \midrule
GLO12 & \cellcolor{gray!15}0.107 & \cellcolor{gray!15}0.111 & \cellcolor{gray!15}0.115 & \cellcolor{gray!15}0.120 & & \cellcolor{gray!15}0.103 & \cellcolor{gray!15}0.107 & \cellcolor{gray!15}0.111 & \cellcolor{gray!15}0.116 & & \cellcolor{gray!15}0.800 & \cellcolor{gray!15}0.811 & \cellcolor{gray!15}0.830 & \cellcolor{gray!15}0.848 & & \cellcolor{gray!15}0.149 & \cellcolor{gray!15}0.150 & \cellcolor{gray!15}0.151 & \cellcolor{gray!15}0.153 \\
GLONET & 0.108 & \cellcolor{blue!8}0.107 & \cellcolor{blue!14}0.108 & \cellcolor{blue!16}0.111 & & 0.102 & \cellcolor{blue!12}0.101 & \cellcolor{blue!19}0.102 & \cellcolor{blue!23}0.104 & & \cellcolor{red!19}0.867 & \cellcolor{red!20}0.884 & \cellcolor{red!20}0.904 & \cellcolor{red!23}0.936 & & \cellcolor{red!16}0.160 & \cellcolor{red!16}0.160 & \cellcolor{red!17}0.163 & \cellcolor{red!20}0.166 \\
WENHAI & \cellcolor{red!45}0.130 & \cellcolor{red!39}0.130 & \cellcolor{red!32}0.131 & \cellcolor{red!26}0.133 & & \cellcolor{red!44}0.123 & \cellcolor{red!33}0.122 & \cellcolor{red!24}0.123 & \cellcolor{red!17}0.125 & & \cellcolor{red!24}0.884 & \cellcolor{red!21}0.887 & \cellcolor{red!18}0.898 & \cellcolor{red!17}0.911 & & \cellcolor{red!45}0.998 & \cellcolor{red!45}0.998 & \cellcolor{red!45}0.998 & \cellcolor{red!45}0.998 \\
XIHE & \cellcolor{red!5}0.109 & \cellcolor{blue!6}0.108 & \cellcolor{blue!16}0.107 & \cellcolor{blue!27}0.105 & & \cellcolor{red!4}0.105 & \cellcolor{blue!8}0.103 & \cellcolor{blue!18}0.102 & \cellcolor{blue!30}0.100 & & \cellcolor{red!7}0.825 & \cellcolor{blue!7}0.785 & \cellcolor{blue!4}0.816 & \cellcolor{blue!6}0.826 & & \cellcolor{blue!5}0.146 & \cellcolor{blue!7}0.145 & \cellcolor{blue!12}0.143 & \cellcolor{blue!12}0.144 \\
\textbf{NJORD} & \cellcolor{red!15}0.114 & \cellcolor{blue!4}0.109 & \cellcolor{blue!17}0.106 & \cellcolor{blue!26}0.106 & & \cellcolor{red!13}0.109 & \cellcolor{blue!6}0.104 & \cellcolor{blue!20}0.101 & \cellcolor{blue!29}0.101 & & \cellcolor{red!15}0.853 & \cellcolor{red!9}0.845 & \cellcolor{red!5}0.849 & 0.855 & & \cellcolor{red!16}0.159 & \cellcolor{red!8}0.155 & \cellcolor{red!3}0.153 & 0.152 \\
\midrule \multicolumn{20}{c}{\textbf{Depth: 300m}} \\ \midrule
GLO12 & \cellcolor{gray!15}0.103 & \cellcolor{gray!15}0.106 & \cellcolor{gray!15}0.110 & \cellcolor{gray!15}0.113 & & \cellcolor{gray!15}0.100 & \cellcolor{gray!15}0.103 & \cellcolor{gray!15}0.107 & \cellcolor{gray!15}0.111 & & \cellcolor{gray!15}0.679 & \cellcolor{gray!15}0.690 & \cellcolor{gray!15}0.709 & \cellcolor{gray!15}0.727 & & \cellcolor{gray!15}0.116 & \cellcolor{gray!15}0.117 & \cellcolor{gray!15}0.119 & \cellcolor{gray!15}0.121 \\
GLONET & \cellcolor{red!3}0.104 & \cellcolor{blue!7}0.102 & \cellcolor{blue!14}0.103 & \cellcolor{blue!18}0.104 & & 0.100 & \cellcolor{blue!10}0.098 & \cellcolor{blue!18}0.098 & \cellcolor{blue!21}0.100 & & \cellcolor{red!18}0.735 & \cellcolor{red!15}0.737 & \cellcolor{red!8}0.736 & \cellcolor{red!8}0.752 & & \cellcolor{red!16}0.125 & \cellcolor{red!15}0.125 & \cellcolor{red!15}0.127 & \cellcolor{red!14}0.129 \\
WENHAI & \cellcolor{red!39}0.120 & \cellcolor{red!30}0.120 & \cellcolor{red!24}0.121 & \cellcolor{red!19}0.123 & & \cellcolor{red!35}0.115 & \cellcolor{red!25}0.114 & \cellcolor{red!17}0.115 & \cellcolor{red!11}0.116 & & \cellcolor{red!22}0.746 & \cellcolor{red!17}0.743 & \cellcolor{red!12}0.748 & \cellcolor{red!8}0.755 & & \cellcolor{red!45}0.925 & \cellcolor{red!45}0.924 & \cellcolor{red!45}0.924 & \cellcolor{red!45}0.924 \\
XIHE & \cellcolor{red!7}0.106 & \cellcolor{blue!3}0.104 & \cellcolor{blue!12}0.104 & \cellcolor{blue!24}0.102 & & \cellcolor{red!5}0.102 & \cellcolor{blue!6}0.100 & \cellcolor{blue!15}0.100 & \cellcolor{blue!27}0.097 & & \cellcolor{red!8}0.704 & \cellcolor{blue!9}0.662 & \cellcolor{blue!9}0.681 & \cellcolor{blue!11}0.693 & & \cellcolor{blue!10}0.111 & \cellcolor{blue!12}0.111 & \cellcolor{blue!20}0.109 & \cellcolor{blue!18}0.111 \\
\textbf{NJORD} & \cellcolor{red!17}0.110 & 0.105 & \cellcolor{blue!16}0.102 & \cellcolor{blue!25}0.101 & & \cellcolor{red!16}0.106 & \cellcolor{blue!4}0.101 & \cellcolor{blue!20}0.098 & \cellcolor{blue!29}0.096 & & \cellcolor{red!14}0.722 & \cellcolor{red!9}0.717 & \cellcolor{red!4}0.722 & 0.730 & & \cellcolor{red!18}0.126 & \cellcolor{red!9}0.122 & 0.120 & \cellcolor{blue!3}0.119 \\
\midrule \multicolumn{20}{c}{\textbf{Depth: 500m}} \\ \midrule
GLO12 & \cellcolor{gray!15}0.094 & \cellcolor{gray!15}0.096 & \cellcolor{gray!15}0.099 & \cellcolor{gray!15}0.102 & & \cellcolor{gray!15}0.091 & \cellcolor{gray!15}0.094 & \cellcolor{gray!15}0.097 & \cellcolor{gray!15}0.100 & & \cellcolor{gray!15}0.508 & \cellcolor{gray!15}0.517 & \cellcolor{gray!15}0.532 & \cellcolor{gray!15}0.547 & & \cellcolor{gray!15}0.085 & \cellcolor{gray!15}0.086 & \cellcolor{gray!15}0.087 & \cellcolor{gray!15}0.088 \\
GLONET & 0.095 & \cellcolor{blue!7}0.093 & \cellcolor{blue!13}0.093 & \cellcolor{blue!16}0.095 & & 0.091 & \cellcolor{blue!10}0.089 & \cellcolor{blue!17}0.089 & \cellcolor{blue!21}0.090 & & \cellcolor{red!17}0.546 & \cellcolor{red!10}0.540 & \cellcolor{red!8}0.551 & \cellcolor{red!8}0.568 & & \cellcolor{red!19}0.092 & \cellcolor{red!18}0.092 & \cellcolor{red!20}0.095 & \cellcolor{red!21}0.097 \\
WENHAI & \cellcolor{red!25}0.104 & \cellcolor{red!18}0.104 & \cellcolor{red!12}0.104 & \cellcolor{red!8}0.106 & & \cellcolor{red!29}0.103 & \cellcolor{red!20}0.102 & \cellcolor{red!13}0.102 & \cellcolor{red!8}0.103 & & \cellcolor{red!22}0.559 & \cellcolor{red!18}0.558 & \cellcolor{red!13}0.564 & \cellcolor{red!10}0.571 & & \cellcolor{red!45}0.808 & \cellcolor{red!45}0.808 & \cellcolor{red!45}0.808 & \cellcolor{red!45}0.808 \\
XIHE & \cellcolor{red!5}0.096 & 0.095 & \cellcolor{blue!10}0.094 & \cellcolor{blue!21}0.093 & & \cellcolor{red!4}0.093 & \cellcolor{blue!5}0.091 & \cellcolor{blue!14}0.091 & \cellcolor{blue!25}0.089 & & \cellcolor{red!8}0.526 & \cellcolor{blue!7}0.502 & \cellcolor{blue!9}0.512 & \cellcolor{blue!11}0.522 & & \cellcolor{blue!9}0.081 & \cellcolor{blue!9}0.082 & \cellcolor{blue!18}0.080 & \cellcolor{blue!16}0.082 \\
\textbf{NJORD} & \cellcolor{red!18}0.101 & 0.096 & \cellcolor{blue!16}0.092 & \cellcolor{blue!26}0.090 & & \cellcolor{red!16}0.098 & \cellcolor{blue!5}0.091 & \cellcolor{blue!21}0.088 & \cellcolor{blue!31}0.086 & & \cellcolor{red!23}0.559 & \cellcolor{red!17}0.555 & \cellcolor{red!10}0.557 & \cellcolor{red!6}0.562 & & \cellcolor{red!25}0.094 & \cellcolor{red!16}0.092 & \cellcolor{red!7}0.090 & 0.089 \\
\bottomrule
\end{tabular}
} 
\vspace{0.1cm}
\begin{center}
\begin{tikzpicture}
    \shade[left color=blue!45, right color=white] (0,0) rectangle (4,0.15);
    \shade[left color=white, right color=red!45] (4,0) rectangle (8,0.15);
    \draw[darkgray] (0,0) rectangle (8,0.15);
    \draw (0,0) -- (0,-0.1) node[below] {\tiny $\le -0.2$};
    \draw (4,0) -- (4,-0.1) node[below] {\tiny $0$};
    \draw (8,0) -- (8,-0.1) node[below] {\tiny $\ge +0.2$};
    \node at (4,-0.8) {\scriptsize Normalized RMSE Difference w.r.t. GLO12};
\end{tikzpicture}
\end{center}

%% file: tables/glo12_track.tex
\resizebox{\textwidth}{!}{%
\begin{tabular}{l cccc c cccc c cccc c cccc}
\toprule
& \multicolumn{4}{c}{\textbf{Zonal geo. current}} && \multicolumn{4}{c}{\textbf{Meridional geo. current}} && \multicolumn{4}{c}{\textbf{SSH}} && \multicolumn{4}{c}{} \\
& \multicolumn{4}{c}{RMSE (m/s)} && \multicolumn{4}{c}{RMSE (m/s)} && \multicolumn{4}{c}{RMSE (m)} && \multicolumn{4}{c}{} \\
\cmidrule(lr){2-5} \cmidrule(lr){7-10} \cmidrule(lr){12-15}
Lead time (days) & 1 & 4 & 7 & 10 && 1 & 4 & 7 & 10 && 1 & 4 & 7 & 10 && & & & \\
\cmidrule{1-15}
GLO12 & \cellcolor{gray!15}0.045 & \cellcolor{gray!15}0.071 & \cellcolor{gray!15}0.104 & \cellcolor{gray!15}0.129 & & \cellcolor{gray!15}0.043 & \cellcolor{gray!15}0.067 & \cellcolor{gray!15}0.099 & \cellcolor{gray!15}0.122 & & \cellcolor{gray!15}0.009 & \cellcolor{gray!15}0.019 & \cellcolor{gray!15}0.036 & \cellcolor{gray!15}0.049 & & & & & \\
GLONET & \cellcolor{red!45}0.119 & \cellcolor{red!45}0.169 & \cellcolor{red!45}0.225 & \cellcolor{red!45}0.273 & & \cellcolor{red!45}0.148 & \cellcolor{red!45}0.200 & \cellcolor{red!45}0.279 & \cellcolor{red!45}0.319 & & \cellcolor{red!45}0.026 & \cellcolor{red!45}0.040 & \cellcolor{red!22}0.054 & \cellcolor{red!15}0.066 & & & & & \\
WENHAI & \cellcolor{red!45}0.108 & \cellcolor{red!45}0.192 & \cellcolor{red!45}0.250 & \cellcolor{red!45}0.301 & & \cellcolor{red!45}0.149 & \cellcolor{red!45}0.403 & \cellcolor{red!45}0.680 & \cellcolor{red!45}0.962 & & \cellcolor{red!45}0.025 & \cellcolor{red!45}0.047 & \cellcolor{red!33}0.063 & \cellcolor{red!23}0.075 & & & & & \\
XIHE & \cellcolor{red!45}0.528 & \cellcolor{red!45}0.525 & \cellcolor{red!45}0.486 & \cellcolor{red!45}0.515 & & \cellcolor{red!45}0.518 & \cellcolor{red!45}0.577 & \cellcolor{red!45}0.518 & \cellcolor{red!45}0.558 & & \cellcolor{red!45}0.037 & \cellcolor{red!45}0.049 & \cellcolor{red!33}0.063 & \cellcolor{red!13}0.064 & & & & & \\
\textbf{NJORD} & \cellcolor{red!31}0.076 & \cellcolor{red!26}0.112 & \cellcolor{red!13}0.135 & \cellcolor{red!7}0.149 & & \cellcolor{red!26}0.067 & \cellcolor{red!25}0.105 & \cellcolor{red!13}0.128 & \cellcolor{red!7}0.141 & & \cellcolor{red!45}0.021 & \cellcolor{red!45}0.040 & \cellcolor{red!19}0.052 & \cellcolor{red!9}0.060 & & & & & \\
\midrule
& \multicolumn{4}{c}{\textbf{Zonal current}} && \multicolumn{4}{c}{\textbf{Meridional current}} && \multicolumn{4}{c}{\textbf{Temperature}} && \multicolumn{4}{c}{\textbf{Salinity}} \\
& \multicolumn{4}{c}{RMSE (m/s)} && \multicolumn{4}{c}{RMSE (m/s)} && \multicolumn{4}{c}{RMSE (°C)} && \multicolumn{4}{c}{RMSE (PSU)} \\
\cmidrule(lr){2-5} \cmidrule(lr){7-10} \cmidrule(lr){12-15} \cmidrule(lr){17-20}
Lead time (days) & 1 & 4 & 7 & 10 && 1 & 4 & 7 & 10 && 1 & 4 & 7 & 10 && 1 & 4 & 7 & 10 \\
\midrule \multicolumn{20}{c}{\textbf{Depth: 0.49m}} \\ \midrule
GLO12 & \cellcolor{gray!15}0.027 & \cellcolor{gray!15}0.056 & \cellcolor{gray!15}0.093 & \cellcolor{gray!15}0.120 & & \cellcolor{gray!15}0.028 & \cellcolor{gray!15}0.055 & \cellcolor{gray!15}0.092 & \cellcolor{gray!15}0.118 & & \cellcolor{gray!15}0.107 & \cellcolor{gray!15}0.209 & \cellcolor{gray!15}0.352 & \cellcolor{gray!15}0.472 & & \cellcolor{gray!15}0.054 & \cellcolor{gray!15}0.100 & \cellcolor{gray!15}0.168 & \cellcolor{gray!15}0.223 \\
GLONET & \cellcolor{red!45}0.065 & \cellcolor{red!20}0.081 & \cellcolor{red!8}0.109 & \cellcolor{red!5}0.132 & & \cellcolor{red!45}0.068 & \cellcolor{red!21}0.080 & \cellcolor{red!8}0.107 & \cellcolor{red!4}0.127 & & \cellcolor{red!45}0.389 & \cellcolor{red!45}0.501 & \cellcolor{red!45}0.711 & \cellcolor{red!34}0.833 & & \cellcolor{red!45}0.141 & \cellcolor{red!45}0.213 & \cellcolor{red!35}0.300 & \cellcolor{red!26}0.349 \\
WENHAI & \cellcolor{red!45}0.084 & \cellcolor{red!45}0.121 & \cellcolor{red!26}0.148 & \cellcolor{red!19}0.170 & & \cellcolor{red!45}0.075 & \cellcolor{red!42}0.107 & \cellcolor{red!16}0.123 & \cellcolor{red!8}0.138 & & \cellcolor{red!45}0.218 & \cellcolor{red!45}0.553 & \cellcolor{red!45}0.841 & \cellcolor{red!45}1.096 & & \cellcolor{red!45}0.157 & \cellcolor{red!45}0.283 & \cellcolor{red!45}0.362 & \cellcolor{red!41}0.423 \\
XIHE & \cellcolor{red!45}0.055 & \cellcolor{red!20}0.081 & \cellcolor{red!3}0.100 & \cellcolor{blue!4}0.110 & & \cellcolor{red!41}0.053 & \cellcolor{red!20}0.080 & \cellcolor{red!3}0.098 & \cellcolor{blue!4}0.108 & & \cellcolor{red!45}0.440 & \cellcolor{red!45}0.496 & \cellcolor{red!25}0.551 & \cellcolor{red!16}0.643 & & \cellcolor{red!45}0.250 & \cellcolor{red!45}0.295 & \cellcolor{red!45}0.344 & \cellcolor{red!35}0.397 \\
\textbf{NJORD} & \cellcolor{red!45}0.055 & \cellcolor{red!33}0.098 & \cellcolor{red!13}0.120 & \cellcolor{red!6}0.136 & & \cellcolor{red!45}0.057 & \cellcolor{red!35}0.098 & \cellcolor{red!13}0.119 & \cellcolor{red!5}0.132 & & \cellcolor{red!33}0.185 & \cellcolor{red!31}0.352 & \cellcolor{red!12}0.444 & \cellcolor{red!4}0.512 & & \cellcolor{red!44}0.107 & \cellcolor{red!45}0.209 & \cellcolor{red!25}0.262 & \cellcolor{red!16}0.300 \\
\midrule \multicolumn{20}{c}{\textbf{Depth: 50m}} \\ \midrule
GLO12 & \cellcolor{gray!15}0.037 & \cellcolor{gray!15}0.053 & \cellcolor{gray!15}0.081 & \cellcolor{gray!15}0.103 & & \cellcolor{gray!15}0.035 & \cellcolor{gray!15}0.052 & \cellcolor{gray!15}0.081 & \cellcolor{gray!15}0.103 & & \cellcolor{gray!15}0.545 & \cellcolor{gray!15}0.579 & \cellcolor{gray!15}0.667 & \cellcolor{gray!15}0.744 & & \cellcolor{gray!15}0.122 & \cellcolor{gray!15}0.130 & \cellcolor{gray!15}0.147 & \cellcolor{gray!15}0.164 \\
GLONET & \cellcolor{red!12}0.047 & \cellcolor{red!6}0.060 & 0.083 & 0.106 & & \cellcolor{red!16}0.048 & \cellcolor{red!6}0.060 & 0.082 & 0.104 & & \cellcolor{red!13}0.702 & \cellcolor{red!15}0.775 & \cellcolor{red!19}0.946 & \cellcolor{red!23}1.131 & & \cellcolor{blue!10}0.096 & \cellcolor{red!3}0.137 & \cellcolor{red!11}0.184 & \cellcolor{red!11}0.205 \\
WENHAI & \cellcolor{red!9}0.044 & \cellcolor{red!19}0.075 & \cellcolor{red!13}0.104 & \cellcolor{red!10}0.127 & & \cellcolor{red!16}0.048 & \cellcolor{red!20}0.075 & \cellcolor{red!9}0.098 & \cellcolor{red!5}0.115 & & \cellcolor{blue!27}0.218 & \cellcolor{blue!12}0.431 & \cellcolor{blue!4}0.607 & 0.741 & & \cellcolor{blue!25}0.055 & \cellcolor{blue!9}0.103 & \cellcolor{blue!3}0.137 & 0.162 \\
XIHE & \cellcolor{red!3}0.039 & \cellcolor{red!8}0.063 & 0.083 & \cellcolor{blue!4}0.095 & & \cellcolor{red!5}0.039 & \cellcolor{red!9}0.062 & 0.082 & \cellcolor{blue!4}0.093 & & \cellcolor{blue!4}0.494 & 0.561 & \cellcolor{blue!4}0.610 & 0.713 & & \cellcolor{blue!3}0.114 & 0.130 & 0.142 & \cellcolor{blue!5}0.147 \\
\textbf{NJORD} & \cellcolor{red!12}0.047 & \cellcolor{red!28}0.086 & \cellcolor{red!14}0.106 & \cellcolor{red!7}0.119 & & \cellcolor{red!17}0.048 & \cellcolor{red!31}0.087 & \cellcolor{red!14}0.106 & \cellcolor{red!6}0.117 & & \cellcolor{blue!22}0.274 & \cellcolor{blue!6}0.497 & \cellcolor{blue!3}0.616 & \cellcolor{blue!3}0.687 & & \cellcolor{blue!24}0.058 & \cellcolor{blue!7}0.108 & \cellcolor{blue!4}0.133 & \cellcolor{blue!5}0.147 \\
\midrule \multicolumn{20}{c}{\textbf{Depth: 100m}} \\ \midrule
GLO12 & \cellcolor{gray!15}0.023 & \cellcolor{gray!15}0.042 & \cellcolor{gray!15}0.070 & \cellcolor{gray!15}0.092 & & \cellcolor{gray!15}0.023 & \cellcolor{gray!15}0.042 & \cellcolor{gray!15}0.071 & \cellcolor{gray!15}0.092 & & \cellcolor{gray!15}0.176 & \cellcolor{gray!15}0.286 & \cellcolor{gray!15}0.461 & \cellcolor{gray!15}0.583 & & \cellcolor{gray!15}0.029 & \cellcolor{gray!15}0.047 & \cellcolor{gray!15}0.075 & \cellcolor{gray!15}0.096 \\
GLONET & \cellcolor{red!36}0.041 & \cellcolor{red!12}0.053 & 0.073 & 0.093 & & \cellcolor{red!36}0.042 & \cellcolor{red!11}0.052 & 0.072 & 0.090 & & \cellcolor{red!45}0.523 & \cellcolor{red!45}0.623 & \cellcolor{red!35}0.820 & \cellcolor{red!32}1.004 & & \cellcolor{red!45}0.076 & \cellcolor{red!45}0.102 & \cellcolor{red!37}0.137 & \cellcolor{red!25}0.151 \\
WENHAI & \cellcolor{red!21}0.033 & \cellcolor{red!21}0.061 & \cellcolor{red!11}0.088 & \cellcolor{red!7}0.107 & & \cellcolor{red!20}0.034 & \cellcolor{red!18}0.060 & \cellcolor{red!8}0.084 & \cellcolor{red!4}0.101 & & \cellcolor{red!13}0.229 & \cellcolor{red!22}0.426 & \cellcolor{red!13}0.597 & \cellcolor{red!11}0.725 & & \cellcolor{red!20}0.041 & \cellcolor{red!27}0.075 & \cellcolor{red!15}0.101 & \cellcolor{red!11}0.119 \\
XIHE & \cellcolor{red!24}0.035 & \cellcolor{red!18}0.059 & \cellcolor{red!4}0.077 & 0.088 & & \cellcolor{red!21}0.035 & \cellcolor{red!16}0.057 & \cellcolor{red!3}0.076 & \cellcolor{blue!3}0.087 & & \cellcolor{red!45}0.501 & \cellcolor{red!44}0.564 & \cellcolor{red!17}0.634 & \cellcolor{red!10}0.716 & & \cellcolor{red!45}0.093 & \cellcolor{red!45}0.112 & \cellcolor{red!23}0.114 & \cellcolor{red!10}0.118 \\
\textbf{NJORD} & \cellcolor{red!39}0.042 & \cellcolor{red!41}0.080 & \cellcolor{red!19}0.100 & \cellcolor{red!10}0.112 & & \cellcolor{red!39}0.044 & \cellcolor{red!42}0.082 & \cellcolor{red!18}0.100 & \cellcolor{red!9}0.110 & & \cellcolor{red!30}0.293 & \cellcolor{red!39}0.533 & \cellcolor{red!20}0.669 & \cellcolor{red!13}0.754 & & \cellcolor{red!29}0.048 & \cellcolor{red!41}0.091 & \cellcolor{red!22}0.112 & \cellcolor{red!13}0.123 \\
\midrule \multicolumn{20}{c}{\textbf{Depth: 200m}} \\ \midrule
GLO12 & \cellcolor{gray!15}0.018 & \cellcolor{gray!15}0.034 & \cellcolor{gray!15}0.058 & \cellcolor{gray!15}0.076 & & \cellcolor{gray!15}0.019 & \cellcolor{gray!15}0.035 & \cellcolor{gray!15}0.060 & \cellcolor{gray!15}0.078 & & \cellcolor{gray!15}0.130 & \cellcolor{gray!15}0.222 & \cellcolor{gray!15}0.369 & \cellcolor{gray!15}0.473 & & \cellcolor{gray!15}0.022 & \cellcolor{gray!15}0.037 & \cellcolor{gray!15}0.059 & \cellcolor{gray!15}0.076 \\
GLONET & \cellcolor{red!34}0.031 & \cellcolor{red!9}0.040 & 0.059 & 0.076 & & \cellcolor{red!31}0.033 & \cellcolor{red!8}0.041 & 0.059 & 0.076 & & \cellcolor{red!45}0.392 & \cellcolor{red!45}0.465 & \cellcolor{red!24}0.567 & \cellcolor{red!18}0.662 & & \cellcolor{red!45}0.050 & \cellcolor{red!41}0.070 & \cellcolor{red!28}0.095 & \cellcolor{red!19}0.107 \\
WENHAI & \cellcolor{red!20}0.026 & \cellcolor{red!20}0.049 & \cellcolor{red!9}0.070 & \cellcolor{red!5}0.085 & & \cellcolor{red!18}0.027 & \cellcolor{red!19}0.049 & \cellcolor{red!8}0.070 & \cellcolor{red!4}0.086 & & \cellcolor{red!14}0.170 & \cellcolor{red!19}0.317 & \cellcolor{red!10}0.447 & \cellcolor{red!7}0.542 & & \cellcolor{red!22}0.032 & \cellcolor{red!29}0.060 & \cellcolor{red!16}0.080 & \cellcolor{red!11}0.095 \\
XIHE & \cellcolor{red!45}0.052 & \cellcolor{red!16}0.046 & \cellcolor{red!3}0.062 & \cellcolor{blue!3}0.072 & & \cellcolor{red!45}0.056 & \cellcolor{red!15}0.046 & \cellcolor{red!3}0.063 & \cellcolor{blue!3}0.073 & & \cellcolor{red!45}0.389 & \cellcolor{red!37}0.404 & \cellcolor{red!12}0.470 & \cellcolor{red!3}0.503 & & \cellcolor{red!45}0.063 & \cellcolor{red!40}0.069 & \cellcolor{red!15}0.078 & \cellcolor{red!4}0.083 \\
\textbf{NJORD} & \cellcolor{red!38}0.033 & \cellcolor{red!45}0.068 & \cellcolor{red!22}0.086 & \cellcolor{red!12}0.096 & & \cellcolor{red!38}0.035 & \cellcolor{red!45}0.071 & \cellcolor{red!22}0.089 & \cellcolor{red!12}0.098 & & \cellcolor{red!29}0.212 & \cellcolor{red!39}0.415 & \cellcolor{red!20}0.531 & \cellcolor{red!12}0.600 & & \cellcolor{red!29}0.036 & \cellcolor{red!44}0.072 & \cellcolor{red!24}0.090 & \cellcolor{red!15}0.100 \\
\midrule \multicolumn{20}{c}{\textbf{Depth: 300m}} \\ \midrule
GLO12 & \cellcolor{gray!15}0.016 & \cellcolor{gray!15}0.031 & \cellcolor{gray!15}0.053 & \cellcolor{gray!15}0.070 & & \cellcolor{gray!15}0.018 & \cellcolor{gray!15}0.032 & \cellcolor{gray!15}0.055 & \cellcolor{gray!15}0.072 & & \cellcolor{gray!15}0.108 & \cellcolor{gray!15}0.191 & \cellcolor{gray!15}0.321 & \cellcolor{gray!15}0.415 & & \cellcolor{gray!15}0.018 & \cellcolor{gray!15}0.031 & \cellcolor{gray!15}0.050 & \cellcolor{gray!15}0.065 \\
GLONET & \cellcolor{red!33}0.028 & \cellcolor{red!9}0.037 & 0.054 & 0.070 & & \cellcolor{red!30}0.030 & \cellcolor{red!8}0.038 & 0.055 & 0.071 & & \cellcolor{red!45}0.336 & \cellcolor{red!45}0.400 & \cellcolor{red!21}0.474 & \cellcolor{red!15}0.554 & & \cellcolor{red!45}0.048 & \cellcolor{red!45}0.063 & \cellcolor{red!28}0.081 & \cellcolor{red!19}0.091 \\
WENHAI & \cellcolor{red!20}0.024 & \cellcolor{red!21}0.046 & \cellcolor{red!10}0.065 & \cellcolor{red!6}0.080 & & \cellcolor{red!18}0.025 & \cellcolor{red!20}0.047 & \cellcolor{red!10}0.067 & \cellcolor{red!6}0.081 & & \cellcolor{red!15}0.144 & \cellcolor{red!20}0.274 & \cellcolor{red!9}0.387 & \cellcolor{red!6}0.469 & & \cellcolor{red!25}0.028 & \cellcolor{red!31}0.052 & \cellcolor{red!17}0.069 & \cellcolor{red!12}0.082 \\
XIHE & \cellcolor{red!45}0.048 & \cellcolor{red!17}0.043 & \cellcolor{red!4}0.058 & 0.068 & & \cellcolor{red!45}0.052 & \cellcolor{red!16}0.044 & \cellcolor{red!4}0.060 & 0.069 & & \cellcolor{red!45}0.346 & \cellcolor{red!36}0.344 & \cellcolor{red!11}0.402 & 0.436 & & \cellcolor{red!45}0.054 & \cellcolor{red!40}0.058 & \cellcolor{red!15}0.066 & \cellcolor{red!4}0.071 \\
\textbf{NJORD} & \cellcolor{red!38}0.030 & \cellcolor{red!45}0.064 & \cellcolor{red!24}0.081 & \cellcolor{red!13}0.091 & & \cellcolor{red!38}0.033 & \cellcolor{red!45}0.067 & \cellcolor{red!24}0.084 & \cellcolor{red!14}0.094 & & \cellcolor{red!29}0.177 & \cellcolor{red!39}0.358 & \cellcolor{red!20}0.464 & \cellcolor{red!12}0.527 & & \cellcolor{red!30}0.030 & \cellcolor{red!45}0.061 & \cellcolor{red!24}0.077 & \cellcolor{red!14}0.086 \\
\midrule \multicolumn{20}{c}{\textbf{Depth: 500m}} \\ \midrule
GLO12 & \cellcolor{gray!15}0.014 & \cellcolor{gray!15}0.026 & \cellcolor{gray!15}0.044 & \cellcolor{gray!15}0.059 & & \cellcolor{gray!15}0.015 & \cellcolor{gray!15}0.027 & \cellcolor{gray!15}0.046 & \cellcolor{gray!15}0.060 & & \cellcolor{gray!15}0.082 & \cellcolor{gray!15}0.147 & \cellcolor{gray!15}0.251 & \cellcolor{gray!15}0.327 & & \cellcolor{gray!15}0.013 & \cellcolor{gray!15}0.022 & \cellcolor{gray!15}0.037 & \cellcolor{gray!15}0.049 \\
GLONET & \cellcolor{red!34}0.024 & \cellcolor{red!12}0.032 & \cellcolor{red!4}0.048 & 0.062 & & \cellcolor{red!31}0.025 & \cellcolor{red!11}0.033 & \cellcolor{red!3}0.049 & 0.063 & & \cellcolor{red!45}0.195 & \cellcolor{red!33}0.254 & \cellcolor{red!18}0.350 & \cellcolor{red!11}0.409 & & \cellcolor{red!45}0.035 & \cellcolor{red!45}0.047 & \cellcolor{red!32}0.063 & \cellcolor{red!20}0.071 \\
WENHAI & \cellcolor{red!25}0.021 & \cellcolor{red!29}0.042 & \cellcolor{red!16}0.059 & \cellcolor{red!10}0.072 & & \cellcolor{red!23}0.023 & \cellcolor{red!28}0.043 & \cellcolor{red!14}0.060 & \cellcolor{red!9}0.073 & & \cellcolor{red!17}0.113 & \cellcolor{red!24}0.225 & \cellcolor{red!11}0.314 & \cellcolor{red!7}0.378 & & \cellcolor{red!35}0.023 & \cellcolor{red!40}0.042 & \cellcolor{red!22}0.055 & \cellcolor{red!15}0.065 \\
XIHE & \cellcolor{red!45}0.040 & \cellcolor{red!24}0.039 & \cellcolor{red!9}0.052 & 0.060 & & \cellcolor{red!45}0.044 & \cellcolor{red!22}0.040 & \cellcolor{red!8}0.053 & 0.062 & & \cellcolor{red!45}0.268 & \cellcolor{red!39}0.275 & \cellcolor{red!10}0.309 & 0.332 & & \cellcolor{red!45}0.044 & \cellcolor{red!45}0.047 & \cellcolor{red!17}0.051 & \cellcolor{red!5}0.054 \\
\textbf{NJORD} & \cellcolor{red!41}0.026 & \cellcolor{red!45}0.057 & \cellcolor{red!29}0.073 & \cellcolor{red!18}0.082 & & \cellcolor{red!40}0.028 & \cellcolor{red!45}0.059 & \cellcolor{red!29}0.076 & \cellcolor{red!18}0.084 & & \cellcolor{red!29}0.135 & \cellcolor{red!41}0.279 & \cellcolor{red!20}0.365 & \cellcolor{red!12}0.415 & & \cellcolor{red!30}0.022 & \cellcolor{red!45}0.045 & \cellcolor{red!24}0.057 & \cellcolor{red!14}0.064 \\
\bottomrule
\end{tabular}
} 
\vspace{0.1cm}
\begin{center}
\begin{tikzpicture}
    \shade[left color=blue!45, right color=white] (0,0) rectangle (4,0.15);
    \shade[left color=white, right color=red!45] (4,0) rectangle (8,0.15);
    \draw[darkgray] (0,0) rectangle (8,0.15);
    \draw (0,0) -- (0,-0.1) node[below] {\tiny $\le -1.0$};
    \draw (4,0) -- (4,-0.1) node[below] {\tiny $0$};
    \draw (8,0) -- (8,-0.1) node[below] {\tiny $\ge +1.0$};
    \node at (4,-0.8) {\scriptsize Normalized RMSE Difference w.r.t. GLO12};
\end{tikzpicture}
\end{center}

%% file: tables/observation_track.tex
\resizebox{0.9\textwidth}{!}{%
\begin{tabular}{l ccccc c ccccc}
\toprule
& \multicolumn{5}{c}{\textbf{Temperature [0-5m]}} && \multicolumn{5}{c}{\textbf{Salinity [0-5m]}} \\
& \multicolumn{5}{c}{RMSE (°C)} && \multicolumn{5}{c}{RMSE (psu)} \\
\cmidrule(lr){2-6} \cmidrule(lr){8-12}
Lead time (days) & 1 & 3 & 5 & 7 & 10 && 1 & 3 & 5 & 7 & 10 \\
\midrule
GLO12 & \cellcolor{gray!15}0.491 & \cellcolor{gray!15}0.570 & \cellcolor{gray!15}0.865 & \cellcolor{gray!15}0.881 & \cellcolor{gray!15}0.961 & & \cellcolor{gray!15}0.251 & \cellcolor{gray!15}0.333 & \cellcolor{gray!15}0.305 & \cellcolor{gray!15}0.275 & \cellcolor{gray!15}0.320 \\
GLONET & \cellcolor{red!31}0.592 & \cellcolor{red!25}0.665 & \cellcolor{red!18}0.971 & \cellcolor{red!25}1.028 & \cellcolor{red!30}1.154 & & \cellcolor{blue!11}0.234 & \cellcolor{red!13}0.361 & \cellcolor{blue!14}0.277 & \cellcolor{blue!9}0.259 & \cellcolor{blue!3}0.313 \\
WENHAI & \cellcolor{red!11}0.526 & \cellcolor{red!44}0.738 & \cellcolor{red!41}1.104 & \cellcolor{red!45}1.260 & \cellcolor{red!45}1.564 & & \cellcolor{red!9}0.266 & \cellcolor{red!6}0.346 & 0.308 & \cellcolor{red!13}0.299 & \cellcolor{red!9}0.340 \\
XIHE & \cellcolor{red!18}0.548 & \cellcolor{red!9}0.605 & \cellcolor{red!10}0.921 & 0.894 & \cellcolor{red!9}1.016 & & \cellcolor{blue!5}0.243 & \cellcolor{red!11}0.358 & \cellcolor{blue!4}0.297 & 0.275 & \cellcolor{blue!4}0.311 \\
\textbf{NJORD} & \cellcolor{blue!6}0.470 & \cellcolor{blue!13}0.519 & \cellcolor{blue!4}0.844 & \cellcolor{blue!8}0.836 & \cellcolor{blue!8}0.912 & & \cellcolor{blue!4}0.245 & \cellcolor{red!3}0.340 & \cellcolor{blue!18}0.268 & \cellcolor{blue!7}0.261 & \cellcolor{blue!4}0.312 \\
\midrule
& \multicolumn{5}{c}{\textbf{Temperature [5-100m]}} && \multicolumn{5}{c}{\textbf{Salinity [5-100m]}} \\
& \multicolumn{5}{c}{RMSE (°C)} && \multicolumn{5}{c}{RMSE (psu)} \\
\midrule
GLO12 & \cellcolor{gray!15}1.054 & \cellcolor{gray!15}1.036 & \cellcolor{gray!15}1.067 & \cellcolor{gray!15}1.107 & \cellcolor{gray!15}1.151 & & \cellcolor{gray!15}0.205 & \cellcolor{gray!15}0.177 & \cellcolor{gray!15}0.231 & \cellcolor{gray!15}0.230 & \cellcolor{gray!15}0.239 \\
GLONET & \cellcolor{red!8}1.108 & 1.035 & \cellcolor{red!11}1.147 & \cellcolor{red!16}1.226 & \cellcolor{red!25}1.345 & & \cellcolor{blue!9}0.192 & \cellcolor{red!17}0.197 & \cellcolor{blue!5}0.224 & \cellcolor{blue!8}0.217 & \cellcolor{blue!4}0.232 \\
WENHAI & \cellcolor{blue!23}0.891 & \cellcolor{blue!13}0.948 & \cellcolor{blue!6}1.024 & 1.097 & \cellcolor{red!12}1.240 & & \cellcolor{red!27}0.242 & \cellcolor{red!45}0.246 & \cellcolor{red!45}0.310 & \cellcolor{red!39}0.289 & \cellcolor{red!32}0.290 \\
XIHE & \cellcolor{blue!21}0.908 & \cellcolor{blue!17}0.921 & \cellcolor{blue!12}0.982 & \cellcolor{blue!15}0.994 & \cellcolor{blue!10}1.076 & & \cellcolor{red!26}0.241 & \cellcolor{red!45}0.237 & \cellcolor{red!15}0.254 & \cellcolor{red!17}0.255 & \cellcolor{red!7}0.250 \\
\textbf{NJORD} & \cellcolor{blue!3}1.036 & \cellcolor{blue!3}1.016 & \cellcolor{blue!4}1.036 & \cellcolor{blue!6}1.065 & \cellcolor{blue!4}1.121 & & \cellcolor{blue!5}0.198 & \cellcolor{red!6}0.184 & \cellcolor{blue!14}0.209 & \cellcolor{blue!11}0.213 & \cellcolor{blue!9}0.225 \\
\midrule
& \multicolumn{5}{c}{\textbf{Temperature [100-300m]}} && \multicolumn{5}{c}{\textbf{Salinity [100-300m]}} \\
& \multicolumn{5}{c}{RMSE (°C)} && \multicolumn{5}{c}{RMSE (psu)} \\
\midrule
GLO12 & \cellcolor{gray!15}1.004 & \cellcolor{gray!15}0.922 & \cellcolor{gray!15}0.905 & \cellcolor{gray!15}0.936 & \cellcolor{gray!15}0.997 & & \cellcolor{gray!15}0.161 & \cellcolor{gray!15}0.145 & \cellcolor{gray!15}0.147 & \cellcolor{gray!15}0.152 & \cellcolor{gray!15}0.159 \\
GLONET & \cellcolor{red!4}1.034 & 0.925 & 0.897 & 0.948 & \cellcolor{red!8}1.048 & & 0.161 & \cellcolor{blue!3}0.143 & \cellcolor{blue!13}0.134 & \cellcolor{blue!13}0.139 & \cellcolor{blue!9}0.149 \\
WENHAI & \cellcolor{blue!12}0.923 & \cellcolor{blue!5}0.891 & \cellcolor{blue!3}0.884 & \cellcolor{blue!3}0.918 & 0.986 & & \cellcolor{red!45}0.229 & 0.146 & \cellcolor{red!45}0.355 & \cellcolor{red!45}0.359 & \cellcolor{red!23}0.183 \\
XIHE & \cellcolor{blue!17}0.892 & \cellcolor{blue!17}0.818 & \cellcolor{blue!12}0.833 & \cellcolor{blue!16}0.835 & \cellcolor{blue!17}0.885 & & \cellcolor{blue!5}0.156 & \cellcolor{blue!16}0.130 & \cellcolor{blue!9}0.138 & \cellcolor{blue!14}0.138 & \cellcolor{blue!19}0.138 \\
\textbf{NJORD} & \cellcolor{red!29}1.196 & \cellcolor{red!37}1.151 & \cellcolor{red!22}1.035 & \cellcolor{red!20}1.061 & \cellcolor{red!18}1.116 & & \cellcolor{red!10}0.172 & \cellcolor{red!19}0.164 & \cellcolor{red!5}0.152 & \cellcolor{red!4}0.156 & \cellcolor{red!4}0.163 \\
\midrule
& \multicolumn{5}{c}{\textbf{Temperature [300-600m]}} && \multicolumn{5}{c}{\textbf{Salinity [300-600m]}} \\
& \multicolumn{5}{c}{RMSE (°C)} && \multicolumn{5}{c}{RMSE (psu)} \\
\midrule
GLO12 & \cellcolor{gray!15}0.616 & \cellcolor{gray!15}0.631 & \cellcolor{gray!15}0.616 & \cellcolor{gray!15}0.641 & \cellcolor{gray!15}0.690 & & \cellcolor{gray!15}0.108 & \cellcolor{gray!15}0.106 & \cellcolor{gray!15}0.095 & \cellcolor{gray!15}0.100 & \cellcolor{gray!15}0.106 \\
GLONET & \cellcolor{red!4}0.632 & \cellcolor{blue!18}0.553 & \cellcolor{blue!14}0.560 & \cellcolor{blue!13}0.586 & \cellcolor{blue!10}0.646 & & \cellcolor{blue!7}0.103 & \cellcolor{blue!30}0.085 & \cellcolor{blue!16}0.085 & \cellcolor{blue!18}0.088 & \cellcolor{blue!12}0.097 \\
WENHAI & \cellcolor{blue!5}0.597 & 0.632 & \cellcolor{blue!5}0.596 & \cellcolor{blue!6}0.617 & \cellcolor{blue!3}0.675 & & \cellcolor{red!45}0.143 & \cellcolor{red!9}0.113 & \cellcolor{red!45}0.231 & \cellcolor{red!45}0.292 & \cellcolor{red!45}0.212 \\
XIHE & \cellcolor{blue!21}0.528 & \cellcolor{blue!30}0.505 & \cellcolor{blue!14}0.557 & \cellcolor{blue!18}0.562 & \cellcolor{blue!19}0.602 & & \cellcolor{blue!25}0.090 & \cellcolor{blue!33}0.083 & \cellcolor{blue!22}0.081 & \cellcolor{blue!25}0.083 & \cellcolor{blue!27}0.086 \\
\textbf{NJORD} & 0.616 & \cellcolor{blue!13}0.577 & \cellcolor{red!4}0.634 & \cellcolor{red!3}0.653 & \cellcolor{red!7}0.720 & & \cellcolor{red!6}0.113 & \cellcolor{blue!8}0.100 & \cellcolor{red!4}0.098 & 0.101 & \cellcolor{red!4}0.108 \\
\midrule
& \multicolumn{5}{c}{\textbf{SST}} && \multicolumn{5}{c}{\textbf{SLA}} \\
& \multicolumn{5}{c}{RMSE (°C)} && \multicolumn{5}{c}{RMSE (m)} \\
\midrule
GLO12 & \cellcolor{gray!15}0.949 & \cellcolor{gray!15}0.845 & \cellcolor{gray!15}0.888 & \cellcolor{gray!15}0.907 & \cellcolor{gray!15}0.931 & & \cellcolor{gray!15}0.112 & \cellcolor{gray!15}0.114 & \cellcolor{gray!15}0.120 & \cellcolor{gray!15}0.122 & \cellcolor{gray!15}0.126 \\
GLONET & \cellcolor{blue!13}0.868 & 0.839 & \cellcolor{red!7}0.928 & \cellcolor{red!12}0.981 & \cellcolor{red!15}1.024 & & \cellcolor{blue!7}0.107 & \cellcolor{blue!9}0.107 & \cellcolor{blue!9}0.113 & \cellcolor{blue!8}0.116 & \cellcolor{blue!3}0.124 \\
WENHAI & \cellcolor{red!10}1.009 & \cellcolor{red!21}0.961 & \cellcolor{red!40}1.123 & \cellcolor{red!45}1.262 & \cellcolor{red!45}1.453 & & 0.113 & \cellcolor{red!3}0.116 & \cellcolor{red!5}0.124 & \cellcolor{red!8}0.128 & \cellcolor{red!10}0.135 \\
XIHE & \cellcolor{blue!6}0.912 & 0.834 & \cellcolor{red!6}0.925 & \cellcolor{red!4}0.933 & \cellcolor{red!6}0.968 & & \cellcolor{blue!10}0.104 & \cellcolor{blue!11}0.106 & \cellcolor{blue!4}0.116 & \cellcolor{blue!3}0.120 & \cellcolor{blue!9}0.119 \\
\textbf{NJORD} & \cellcolor{blue!10}0.887 & \cellcolor{blue!15}0.758 & \cellcolor{blue!7}0.846 & \cellcolor{blue!8}0.855 & \cellcolor{blue!10}0.869 & & 0.113 & 0.113 & \cellcolor{blue!4}0.117 & \cellcolor{blue!4}0.119 & \cellcolor{blue!5}0.122 \\
\midrule
& \multicolumn{5}{c}{\textbf{Zonal current [15m]}} && \multicolumn{5}{c}{\textbf{Meridional current [15m]}} \\
& \multicolumn{5}{c}{RMSE (m/s)} && \multicolumn{5}{c}{RMSE (m/s)} \\
\midrule
GLO12 & \cellcolor{gray!15}0.206 & \cellcolor{gray!15}0.219 & \cellcolor{gray!15}0.221 & \cellcolor{gray!15}0.229 & \cellcolor{gray!15}0.236 & & \cellcolor{gray!15}0.191 & \cellcolor{gray!15}0.211 & \cellcolor{gray!15}0.210 & \cellcolor{gray!15}0.216 & \cellcolor{gray!15}0.223 \\
GLONET & \cellcolor{blue!8}0.194 & \cellcolor{blue!7}0.208 & \cellcolor{blue!8}0.210 & \cellcolor{blue!9}0.216 & \cellcolor{blue!11}0.219 & & \cellcolor{blue!11}0.178 & \cellcolor{blue!11}0.195 & \cellcolor{blue!10}0.196 & \cellcolor{blue!12}0.199 & \cellcolor{blue!13}0.203 \\
WENHAI & \cellcolor{red!6}0.215 & \cellcolor{red!7}0.228 & 0.224 & 0.231 & 0.234 & & \cellcolor{red!3}0.195 & 0.213 & 0.207 & \cellcolor{blue!3}0.211 & \cellcolor{blue!6}0.215 \\
XIHE & \cellcolor{red!3}0.210 & 0.220 & 0.218 & \cellcolor{blue!6}0.220 & \cellcolor{blue!10}0.221 & & 0.194 & \cellcolor{blue!3}0.206 & \cellcolor{blue!6}0.202 & \cellcolor{blue!10}0.202 & \cellcolor{blue!14}0.203 \\
\textbf{NJORD} & \cellcolor{blue!3}0.202 & \cellcolor{blue!5}0.211 & \cellcolor{blue!7}0.211 & \cellcolor{blue!9}0.216 & \cellcolor{blue!10}0.220 & & \cellcolor{blue!3}0.188 & \cellcolor{blue!5}0.203 & \cellcolor{blue!8}0.199 & \cellcolor{blue!9}0.203 & \cellcolor{blue!11}0.207 \\
\bottomrule
\end{tabular}
} 
\vspace{0.1cm}
\begin{center}
\begin{tikzpicture}
    \shade[left color=blue!45, right color=white] (0,0) rectangle (3.5,0.15);
    \shade[left color=white, right color=red!45] (3.5,0) rectangle (7,0.15);
    \draw[darkgray] (0,0) rectangle (7,0.15);
    \draw (0,0) -- (0,-0.1) node[below] {\tiny $\le -0.3$};
    \draw (3.5,0) -- (3.5,-0.1) node[below] {\tiny $0$};
    \draw (7,0) -- (7,-0.1) node[below] {\tiny $\ge +0.3$};
    \node at (3.5,-0.8) {\scriptsize Normalized RMSE Difference w.r.t. GLO12};
\end{tikzpicture}
\end{center}

%% file: checklist.tex
\section*{NeurIPS Paper Checklist}

\begin{enumerate}

\item {\bf Claims}
    \item[] Question: Do the main claims made in the abstract and introduction accurately reflect the paper's contributions and scope?
    \item[] Answer: \answerYes{}
    \item[] Justification: We outline the scope and contributions of our work in the introduction, with a summary of the main contributions provided in the final paragraph.
    \item[] Guidelines:
    \begin{itemize}
        \item The answer \answerNA{} means that the abstract and introduction do not include the claims made in the paper.
        \item The abstract and/or introduction should clearly state the claims made, including the contributions made in the paper and important assumptions and limitations. A \answerNo{} or \answerNA{} answer to this question will not be perceived well by the reviewers. 
        \item The claims made should match theoretical and experimental results, and reflect how much the results can be expected to generalize to other settings. 
        \item It is fine to include aspirational goals as motivation as long as it is clear that these goals are not attained by the paper. 
    \end{itemize}

\item {\bf Limitations}
    \item[] Question: Does the paper discuss the limitations of the work performed by the authors?
    \item[] Answer: \answerYes{}
    \item[] Justification: Limitations are discussed in \cref{sec:discussion}.
    \item[] Guidelines:
    \begin{itemize}
        \item The answer \answerNA{} means that the paper has no limitation while the answer \answerNo{} means that the paper has limitations, but those are not discussed in the paper. 
        \item The authors are encouraged to create a separate ``Limitations'' section in their paper.
        \item The paper should point out any strong assumptions and how robust the results are to violations of these assumptions (e.g., independence assumptions, noiseless settings, model well-specification, asymptotic approximations only holding locally). The authors should reflect on how these assumptions might be violated in practice and what the implications would be.
        \item The authors should reflect on the scope of the claims made, e.g., if the approach was only tested on a few datasets or with a few runs. In general, empirical results often depend on implicit assumptions, which should be articulated.
        \item The authors should reflect on the factors that influence the performance of the approach. For example, a facial recognition algorithm may perform poorly when image resolution is low or images are taken in low lighting. Or a speech-to-text system might not be used reliably to provide closed captions for online lectures because it fails to handle technical jargon.
        \item The authors should discuss the computational efficiency of the proposed algorithms and how they scale with dataset size.
        \item If applicable, the authors should discuss possible limitations of their approach to address problems of privacy and fairness.
        \item While the authors might fear that complete honesty about limitations might be used by reviewers as grounds for rejection, a worse outcome might be that reviewers discover limitations that aren't acknowledged in the paper. The authors should use their best judgment and recognize that individual actions in favor of transparency play an important role in developing norms that preserve the integrity of the community. Reviewers will be specifically instructed to not penalize honesty concerning limitations.
    \end{itemize}

\item {\bf Theory assumptions and proofs}
    \item[] Question: For each theoretical result, does the paper provide the full set of assumptions and a complete (and correct) proof?
    \item[] Answer: \answerNA{}
    \item[] Justification: The paper does not include theoretical results.
    \item[] Guidelines:
    \begin{itemize}
        \item The answer \answerNA{} means that the paper does not include theoretical results. 
        \item All the theorems, formulas, and proofs in the paper should be numbered and cross-referenced.
        \item All assumptions should be clearly stated or referenced in the statement of any theorems.
        \item The proofs can either appear in the main paper or the supplemental material, but if they appear in the supplemental material, the authors are encouraged to provide a short proof sketch to provide intuition. 
        \item Inversely, any informal proof provided in the core of the paper should be complemented by formal proofs provided in appendix or supplemental material.
        \item Theorems and Lemmas that the proof relies upon should be properly referenced. 
    \end{itemize}

    \item {\bf Experimental result reproducibility}
    \item[] Question: Does the paper fully disclose all the information needed to reproduce the main experimental results of the paper to the extent that it affects the main claims and/or conclusions of the paper (regardless of whether the code and data are provided or not)?
    \item[] Answer: \answerYes{}
    \item[] Justification:  We provide sufficient details on the model architecture and experimental setup to enable reproduction of our results. Apart from the descriptions in the main paper, model details are given in \cref{app:model-details}, data details in \cref{app:data-details}, graph construction details in \cref{app:graph-details}, and training details in \cref{app:training-details}.
        \item[] Guidelines:
    \begin{itemize}
        \item The answer \answerNA{} means that the paper does not include experiments.
        \item If the paper includes experiments, a \answerNo{} answer to this question will not be perceived well by the reviewers: Making the paper reproducible is important, regardless of whether the code and data are provided or not.
        \item If the contribution is a dataset and\slash or model, the authors should describe the steps taken to make their results reproducible or verifiable. 
        \item Depending on the contribution, reproducibility can be accomplished in various ways. For example, if the contribution is a novel architecture, describing the architecture fully might suffice, or if the contribution is a specific model and empirical evaluation, it may be necessary to either make it possible for others to replicate the model with the same dataset, or provide access to the model. In general. releasing code and data is often one good way to accomplish this, but reproducibility can also be provided via detailed instructions for how to replicate the results, access to a hosted model (e.g., in the case of a large language model), releasing of a model checkpoint, or other means that are appropriate to the research performed.
        \item While NeurIPS does not require releasing code, the conference does require all submissions to provide some reasonable avenue for reproducibility, which may depend on the nature of the contribution. For example
        \begin{enumerate}
            \item If the contribution is primarily a new algorithm, the paper should make it clear how to reproduce that algorithm.
            \item If the contribution is primarily a new model architecture, the paper should describe the architecture clearly and fully.
            \item If the contribution is a new model (e.g., a large language model), then there should either be a way to access this model for reproducing the results or a way to reproduce the model (e.g., with an open-source dataset or instructions for how to construct the dataset).
            \item We recognize that reproducibility may be tricky in some cases, in which case authors are welcome to describe the particular way they provide for reproducibility. In the case of closed-source models, it may be that access to the model is limited in some way (e.g., to registered users), but it should be possible for other researchers to have some path to reproducing or verifying the results.
        \end{enumerate}
    \end{itemize}

\item {\bf Open access to data and code}
    \item[] Question: Does the paper provide open access to the data and code, with sufficient instructions to faithfully reproduce the main experimental results, as described in supplemental material?
    \item[] Answer: \answerYes{}
    \item[] Justification: Data sources are described in the paper, and are all openly available. We provide anonymized code and instructions in the supplementary material to reproduce the main experimental results. The archive includes code for training and evaluation, along with environment specifications and example commands. The code will be open sourced upon publication of the paper.
        \item[] Guidelines:
    \begin{itemize}
        \item The answer \answerNA{} means that paper does not include experiments requiring code.
        \item Please see the NeurIPS code and data submission guidelines (\url{https://neurips.cc/public/guides/CodeSubmissionPolicy}) for more details.
        \item While we encourage the release of code and data, we understand that this might not be possible, so \answerNo{} is an acceptable answer. Papers cannot be rejected simply for not including code, unless this is central to the contribution (e.g., for a new open-source benchmark).
        \item The instructions should contain the exact command and environment needed to run to reproduce the results. See the NeurIPS code and data submission guidelines (\url{https://neurips.cc/public/guides/CodeSubmissionPolicy}) for more details.
        \item The authors should provide instructions on data access and preparation, including how to access the raw data, preprocessed data, intermediate data, and generated data, etc.
        \item The authors should provide scripts to reproduce all experimental results for the new proposed method and baselines. If only a subset of experiments are reproducible, they should state which ones are omitted from the script and why.
        \item At submission time, to preserve anonymity, the authors should release anonymized versions (if applicable).
        \item Providing as much information as possible in supplemental material (appended to the paper) is recommended, but including URLs to data and code is permitted.
    \end{itemize}

\item {\bf Experimental setting/details}
    \item[] Question: Does the paper specify all the training and test details (e.g., data splits, hyperparameters, how they were chosen, type of optimizer) necessary to understand the results?
    \item[] Answer: \answerYes{}
    \item[] Justification: Experimental details are discussed in the main paper and \cref{app:training-details}
        \item[] Guidelines:
    \begin{itemize}
        \item The answer \answerNA{} means that the paper does not include experiments.
        \item The experimental setting should be presented in the core of the paper to a level of detail that is necessary to appreciate the results and make sense of them.
        \item The full details can be provided either with the code, in appendix, or as supplemental material.
    \end{itemize}

\item {\bf Experiment statistical significance}
    \item[] Question: Does the paper report error bars suitably and correctly defined or other appropriate information about the statistical significance of the experiments?
    \item[] Answer: \answerNo{}
    \item[] Justification: The main sources of randomness in our experiments relate to the random seed used for parameter initialization and sampling. A proper statistical analysis would require training and evaluating multiple models. However, the computational cost associated makes it unfeasible to train enough models to draw any well founded conclusions.
     \item[] Guidelines:
    \begin{itemize}
        \item The answer \answerNA{} means that the paper does not include experiments.
        \item The authors should answer \answerYes{} if the results are accompanied by error bars, confidence intervals, or statistical significance tests, at least for the experiments that support the main claims of the paper.
        \item The factors of variability that the error bars are capturing should be clearly stated (for example, train/test split, initialization, random drawing of some parameter, or overall run with given experimental conditions).
        \item The method for calculating the error bars should be explained (closed form formula, call to a library function, bootstrap, etc.)
        \item The assumptions made should be given (e.g., Normally distributed errors).
        \item It should be clear whether the error bar is the standard deviation or the standard error of the mean.
        \item It is OK to report 1-sigma error bars, but one should state it. The authors should preferably report a 2-sigma error bar than state that they have a 96\% CI, if the hypothesis of Normality of errors is not verified.
        \item For asymmetric distributions, the authors should be careful not to show in tables or figures symmetric error bars that would yield results that are out of range (e.g., negative error rates).
        \item If error bars are reported in tables or plots, the authors should explain in the text how they were calculated and reference the corresponding figures or tables in the text.
    \end{itemize}

\item {\bf Experiments compute resources}
    \item[] Question: For each experiment, does the paper provide sufficient information on the computer resources (type of compute workers, memory, time of execution) needed to reproduce the experiments?
    \item[] Answer: \answerYes{}
    \item[] Justification: See \cref{app:training-details} for computational details. The amount of compute required to train the models is provided in \crefrange{tab:global_training_schedule}{tab:baltic_training_schedule}.
    \item[] Guidelines:
     \begin{itemize}
        \item The answer \answerNA{} means that the paper does not include experiments.
        \item The paper should indicate the type of compute workers CPU or GPU, internal cluster, or cloud provider, including relevant memory and storage.
        \item The paper should provide the amount of compute required for each of the individual experimental runs as well as estimate the total compute. 
        \item The paper should disclose whether the full research project required more compute than the experiments reported in the paper (e.g., preliminary or failed experiments that didn't make it into the paper). 
    \end{itemize}
    
\item {\bf Code of ethics}
    \item[] Question: Does the research conducted in the paper conform, in every respect, with the NeurIPS Code of Ethics \url{https://neurips.cc/public/EthicsGuidelines}?
    \item[] Answer: \answerYes{} 
    \item[] Justification:
    \item[] Guidelines:
    \begin{itemize}
        \item The answer \answerNA{} means that the authors have not reviewed the NeurIPS Code of Ethics.
        \item If the authors answer \answerNo, they should explain the special circumstances that require a deviation from the Code of Ethics.
        \item The authors should make sure to preserve anonymity (e.g., if there is a special consideration due to laws or regulations in their jurisdiction).
    \end{itemize}

\item {\bf Broader impacts}
    \item[] Question: Does the paper discuss both potential positive societal impacts and negative societal impacts of the work performed?
    \item[] Answer: \answerYes{}
    \item[] Justification: Societal impacts of improved ocean forecasting is discussed in the introduction and conclusion. It is beneficial for e.g. maritime navigation, managing fisheries and environmental monitoring.
    While we don't see direct negative impact of this research, we acknowledge that further validation of forecasting models is required before taking them into deployment, in order to avoid downstream risks.
        \item[] Guidelines:
    \begin{itemize}
        \item The answer \answerNA{} means that there is no societal impact of the work performed.
        \item If the authors answer \answerNA{} or \answerNo, they should explain why their work has no societal impact or why the paper does not address societal impact.
        \item Examples of negative societal impacts include potential malicious or unintended uses (e.g., disinformation, generating fake profiles, surveillance), fairness considerations (e.g., deployment of technologies that could make decisions that unfairly impact specific groups), privacy considerations, and security considerations.
        \item The conference expects that many papers will be foundational research and not tied to particular applications, let alone deployments. However, if there is a direct path to any negative applications, the authors should point it out. For example, it is legitimate to point out that an improvement in the quality of generative models could be used to generate Deepfakes for disinformation. On the other hand, it is not needed to point out that a generic algorithm for optimizing neural networks could enable people to train models that generate Deepfakes faster.
        \item The authors should consider possible harms that could arise when the technology is being used as intended and functioning correctly, harms that could arise when the technology is being used as intended but gives incorrect results, and harms following from (intentional or unintentional) misuse of the technology.
        \item If there are negative societal impacts, the authors could also discuss possible mitigation strategies (e.g., gated release of models, providing defenses in addition to attacks, mechanisms for monitoring misuse, mechanisms to monitor how a system learns from feedback over time, improving the efficiency and accessibility of ML).
    \end{itemize}
    
\item {\bf Safeguards}
    \item[] Question: Does the paper describe safeguards that have been put in place for responsible release of data or models that have a high risk for misuse (e.g., pre-trained language models, image generators, or scraped datasets)?
    \item[] Answer: \answerNA{}
    \item[] Justification:
    \item[] Guidelines:
        \begin{itemize}
        \item The answer \answerNA{} means that the paper poses no such risks.
        \item Released models that have a high risk for misuse or dual-use should be released with necessary safeguards to allow for controlled use of the model, for example by requiring that users adhere to usage guidelines or restrictions to access the model or implementing safety filters. 
        \item Datasets that have been scraped from the Internet could pose safety risks. The authors should describe how they avoided releasing unsafe images.
        \item We recognize that providing effective safeguards is challenging, and many papers do not require this, but we encourage authors to take this into account and make a best faith effort.
    \end{itemize}

\item {\bf Licenses for existing assets}
    \item[] Question: Are the creators or original owners of assets (e.g., code, data, models), used in the paper, properly credited and are the license and terms of use explicitly mentioned and properly respected?
    \item[] Answer: \answerYes{}
    \item[] Justification: See \cref{app:data-details} where we refer to the public licenses of OceanBench, ERA5, and Copernicus Marine Service data. The data sources are also cited in the text.
        \item[] Guidelines:
    \begin{itemize}
        \item The answer \answerNA{} means that the paper does not use existing assets.
        \item The authors should cite the original paper that produced the code package or dataset.
        \item The authors should state which version of the asset is used and, if possible, include a URL.
        \item The name of the license (e.g., CC-BY 4.0) should be included for each asset.
        \item For scraped data from a particular source (e.g., website), the copyright and terms of service of that source should be provided.
        \item If assets are released, the license, copyright information, and terms of use in the package should be provided. For popular datasets, \url{paperswithcode.com/datasets} has curated licenses for some datasets. Their licensing guide can help determine the license of a dataset.
        \item For existing datasets that are re-packaged, both the original license and the license of the derived asset (if it has changed) should be provided.
        \item If this information is not available online, the authors are encouraged to reach out to the asset's creators.
    \end{itemize}

\item {\bf New assets}
    \item[] Question: Are new assets introduced in the paper well documented and is the documentation provided alongside the assets?
    \item[] Answer: \answerYes{}
    \item[] Justification: We provide anonymized code as a supplement to the paper. The code comes with documentation to the degree that other researchers can reproduce and extend our work.
        \item[] Guidelines:
    \begin{itemize}
        \item The answer \answerNA{} means that the paper does not release new assets.
        \item Researchers should communicate the details of the dataset\slash code\slash model as part of their submissions via structured templates. This includes details about training, license, limitations, etc. 
        \item The paper should discuss whether and how consent was obtained from people whose asset is used.
        \item At submission time, remember to anonymize your assets (if applicable). You can either create an anonymized URL or include an anonymized zip file.
    \end{itemize}

\item {\bf Crowdsourcing and research with human subjects}
    \item[] Question: For crowdsourcing experiments and research with human subjects, does the paper include the full text of instructions given to participants and screenshots, if applicable, as well as details about compensation (if any)? 
    \item[] Answer: \answerNA{}
    \item[] Justification:
        \item[] Guidelines:
    \begin{itemize}
        \item The answer \answerNA{} means that the paper does not involve crowdsourcing nor research with human subjects.
        \item Including this information in the supplemental material is fine, but if the main contribution of the paper involves human subjects, then as much detail as possible should be included in the main paper. 
        \item According to the NeurIPS Code of Ethics, workers involved in data collection, curation, or other labor should be paid at least the minimum wage in the country of the data collector. 
    \end{itemize}

\item {\bf Institutional review board (IRB) approvals or equivalent for research with human subjects}
    \item[] Question: Does the paper describe potential risks incurred by study participants, whether such risks were disclosed to the subjects, and whether Institutional Review Board (IRB) approvals (or an equivalent approval/review based on the requirements of your country or institution) were obtained?
    \item[] Answer: \answerNA{}
    \item[] Justification:
       \item[] Guidelines:
    \begin{itemize}
        \item The answer \answerNA{} means that the paper does not involve crowdsourcing nor research with human subjects.
        \item Depending on the country in which research is conducted, IRB approval (or equivalent) may be required for any human subjects research. If you obtained IRB approval, you should clearly state this in the paper. 
        \item We recognize that the procedures for this may vary significantly between institutions and locations, and we expect authors to adhere to the NeurIPS Code of Ethics and the guidelines for their institution. 
        \item For initial submissions, do not include any information that would break anonymity (if applicable), such as the institution conducting the review.
    \end{itemize}

\item {\bf Declaration of LLM usage}
    \item[] Question: Does the paper describe the usage of LLMs if it is an important, original, or non-standard component of the core methods in this research? Note that if the LLM is used only for writing, editing, or formatting purposes and does \emph{not} impact the core methodology, scientific rigor, or originality of the research, declaration is not required.
    \item[] Answer: \answerNA{}
    \item[] Justification:
        \item[] Guidelines:
    \begin{itemize}
        \item The answer \answerNA{} means that the core method development in this research does not involve LLMs as any important, original, or non-standard components.
        \item Please refer to our LLM policy in the NeurIPS handbook for what should or should not be described.
    \end{itemize}

\end{enumerate}